%% file: main.tex
\documentclass[10pt]{article}

\usepackage[utf8]{inputenc} 
\usepackage[T1]{fontenc}    
\usepackage{lmodern}
\usepackage{hyperref}       
\usepackage{url}            
\usepackage{booktabs}       
\usepackage{amsthm}         
\usepackage{amsfonts}       
\usepackage{amssymb}        
\usepackage{mathtools}      
\usepackage{nicefrac}       
\usepackage{microtype}      
\usepackage{subfiles}       
\usepackage{bbold}          
\usepackage{xcolor}         
\usepackage{subcaption}     
\usepackage{soul}

\usepackage[numbers, compress]{natbib}
\usepackage[verbose=true,letterpaper]{geometry}
\usepackage[parfill]{parskip}

\AtBeginDocument{
  \newgeometry{
    textheight=9in,
    textwidth=5.5in,
    top=1in,
    headheight=12pt,
    headsep=25pt,
    footskip=30pt
  }
}

\definecolor{DarkGreen}{rgb}{0,0.4,0}

\usepackage{mymacros}

\input{varun_macros}

\newcommand{\ceil}[1]{\left\lceil #1 \right\rceil}
\renewcommand{\matrix}[1]{\ensuremath{\mathbf{#1}}}
\newcommand{\matrixid}{\ensuremath{\matrix{I}}}
\renewcommand{\vec}[1]{\ensuremath{\mathbf{#1}}}
\newcommand{\X}{\ensuremath{\matrix{X}}}
\newcommand{\Xt}{\ensuremath{\X^{\mathsf{T}}}}
\newcommand{\XtX}{\ensuremath{\X^{\mathsf{T}}\X}}
\newcommand{\id}[1]{\ensuremath{\vec{1}_{#1}}}
\newcommand{\maxnoise}{
  \ensuremath{\norm{\frac{1}{n} \Xt \vec{\xi}}_{\infty}}}
\newcommand{\inlinemaxnoise}{
  \ensuremath{\inlinenorm{\Xt \vec{\xi}}_{\infty} / n}}
\newcommand{\noisevec}{
  \ensuremath{\frac{1}{n} \Xt \vec{\xi}}}
\newcommand{\abs}[1]{\left\lvert #1 \right\rvert}
\newcommand{\inlineabs}[1]{\lvert #1 \rvert}
\newcommand{\wstar}{\ensuremath{\vec{w}^{\star}}}
\newcommand{\wmax}{\ensuremath{w^{\star}_{\max}}}
\newcommand{\wmin}{\ensuremath{w^{\star}_{\min}}}
\newcommand{\vone}{\id{}}
\newcommand{\vb}{\vec{b}}
\newcommand{\ve}{\vec{e}}
\newcommand{\vp}{\vec{p}}

\newcommand{\vs}{\vec{s}}
\newcommand{\vu}{\vec{u}}

\newcommand{\vw}{\vec{w}}

\newcommand{\vxi}{\vec{\xi}}
\newcommand{\LS}{\textsc{LS}}
\newcommand{\sign}{\mathrm{sign}}

\begingroup
\makeatletter
\@for\theoremstyle:=definition,remark,plain\do{%
  \expandafter\g@addto@macro\csname
  th@\theoremstyle\endcsname{%
    \addtolength\thm@preskip\parskip
  }%
}
\endgroup

\theoremstyle{plain}

\newtheorem{lemma}{Lemma}
\newtheorem{proposition}{Proposition}
\newtheorem{remark}{Remark}
\newtheorem{theorem}{Theorem}
\newtheorem{corollary}{Corollary}
\newtheorem{definition}{Definition}
\newtheorem{algorithm}{Algorithm}

\title{Implicit Regularization for Optimal Sparse Recovery}

\author{
  Tomas Va\v{s}kevi\v{c}ius\textsuperscript{1},
  Varun Kanade\textsuperscript{2},
  Patrick Rebeschini\textsuperscript{1} \\
  \textsuperscript{1} Department of Statistics,
  \textsuperscript{2} Department of Computer Science \\
  University of Oxford \\
  \texttt{\{tomas.vaskevicius, patrick.rebeschini\}@stats.ox.ac.uk} \\
  \texttt{varunk@cs.ox.ac.uk}
}

\begin{document}

\maketitle

\begin{abstract}
  We investigate implicit regularization schemes for gradient descent methods
  applied to unpenalized least squares regression to solve the problem of
  reconstructing a sparse signal from an underdetermined
  system of linear measurements under the restricted isometry assumption.
  For a given parametrization yielding a non-convex optimization problem,
  we show that prescribed choices of initialization, step size and stopping
  time yield a statistically and computationally optimal algorithm that
  achieves the minimax rate with the same cost required to read the data
  up to poly-logarithmic factors.
  Beyond minimax optimality, we show that our algorithm adapts to instance
  difficulty and yields a dimension-independent rate when the
  signal-to-noise ratio is high enough.
  Key to the computational efficiency of our method is an increasing step size
  scheme that adapts to refined estimates of the true solution.
  We validate our findings with
  numerical experiments and compare our algorithm against explicit
    $\ell_{1}$ penalization.
  Going from hard instances to easy ones,
  our algorithm is seen to undergo a phase
  transition, eventually matching least squares with an
  oracle knowledge of the true support.
\end{abstract}


\subfile{files/introduction.tex}

\subfile{files/preliminaries.tex}

\subfile{files/main_results.tex}

\subfile{files/proof_sketch.tex}

\subfile{files/simulations.tex}

\subfile{files/further_improvements.tex}

\bibliographystyle{plainnat}
\bibliography{references}

\clearpage

\subfile{files/appendix.tex}

\end{document}

%% file: varun_macros.tex
\renewcommand{\th}{^\mathit{th}}

%% file: files/introduction.tex
\section{Introduction}

Many problems in machine learning, science and engineering involve
high-dimensional datasets where the dimensionality of the data $d$ is greater
than the number of data points $n$. Linear regression with sparsity constraints
is an archetypal problem in this setting. The goal is to estimate a
$d$-dimensional vector $\vec{w}^\star\in\mathbb{R}^d$ with $k$ non-zero
components from $n$ data points $(\vec{x}_i,
y_i)\in\mathbb{R}^d\times\mathbb{R}$, $i\in\{1,\ldots,n\}$,  linked by
the linear relationship $y_i = \langle \vec{x}_i, \vec{w}^\star \rangle +
\xi_i$, where $\xi_i$ is a possible perturbation to the $i\th$ observation. In
matrix-vector form the model reads $\vec{y}=\X \vec{w}^\star + \vec{\xi}$,
where $\vec{x}_i$ corresponds to the $i\th$ row of the $n\times d$ design matrix $\X$. Over
the past couple of decades, sparse linear regression has been
extensively investigated from the point of view of both statistics and optimization.

In statistics, sparsity has been enforced by designing estimators
with \emph{explicit} regularization schemes based on the $\ell_1$ norm, such as
the lasso \cite{tibshirani1996regression} and the closely related
basis pursuit \cite{chen1998atomic}
and Dantzig selector \cite{candes2007dantzig}. In the noiseless setting
($\vec{\xi} = 0$), exact recovery is possible if and only if the design matrix
satisfies the restricted nullspace property
\cite{cohen2009compressed,donoho2001uncertainty,feuer2003sparse}. In the noisy
setting ($\vec{\xi} \neq 0$), exact recovery is not feasible and a natural criterion
involves designing estimators $\widehat{\vec{w}}$ that can recover the
minimax-optimal rate $k\sigma^2\log (d/k) /n$ for the squared $\ell_2$ error $\|
\widehat{\vec{w}} - \vec{w}^\star \|_2^2$ in the case of i.i.d.\ sub-Gaussian
noise with variance proxy $\sigma^2$ when the design matrix satisfies restricted eigenvalue conditions
\cite{bickel2009simultaneous,van2007deterministic}.
The lasso estimator, defined as any vector
$\vec{w}$ that minimizes the objective $\| \X \vec{w} - \vec{y} \|_2^2 + \lambda
\| \vec{w} \|_1$, achieves the minimax-optimal rate upon proper tuning of the
regularization parameter $\lambda$.
The restricted isometry property (RIP)
\cite{candes2005decoding} has been largely considered in the literature, as it implies both the restricted
nullspace and eigenvalue conditions
\cite{cohen2009compressed,van2009conditions}, and as it is satisfied when the
entries of $\X$ are i.i.d.\ sub-Gaussian and subexponential
with sample size $n=\Omega(k\log(d/k))$ and $n=\Omega(k\log^2(d/k))$
respectively \cite{mendelson2008uniform, adamczak2011restricted}, or when the
columns are unitary, e.g.
\cite{guedon2007subspaces,guedon2008majorizing,romberg2009compressive,rudelson2008sparse}.

In optimization, computationally efficient iterative algorithms
have been designed to solve convex problems based on $\ell_1$ constraints and
penalties, such as composite/proximal methods
\cite{bach2012optimization,parikh2014proximal}. Under restricted eigenvalue
conditions, such as restricted strong convexity and restricted smoothness, various
iterative methods have been shown to yield exponential convergence to the problem
solution globally up to the statistical precision of the model \cite{agarwal2010fast},
or locally once the iterates are close enough to the optimum and the support of the solution is identified
\cite{Bredies2008,hale2008fixed,tao2016local}. In some regimes, for a prescribed choice of the
regularization parameter, these algorithms are computationally efficient. They
require $\widetilde{O}(1)$ iterations, where the notation
$\widetilde{O}$ hides poly-logarithmic terms, and each iteration costs
$O(nd)$.
Hence the total running cost is $\widetilde{O}(nd)$, which is the
cost to store/read the data in/from memory.

These results attest that there are regimes where optimal methods for sparse
linear regression exist. However, these results reply upon tuning the hyperparameters for optimization, such as the step size, carefully, which in turn depends on identifying the correct hyperparameters, such as $\lambda$, for regularization. In practice, one has to resort to
cross-validation techniques to tune the regularization parameter.
Cross-validation adds an additional burden from a computational point of view,
as the optimization algorithms need to be run for different choices of the
regularization terms. In the context of linear regression with $\ell_2$
penalty, a.k.a.\ ridge regression, potential computational savings have motivated
research on the design of \emph{implicit} regularization schemes where model
complexity is directly controlled by tuning the hyper-parameters of solvers applied to unpenalized/unconstrained programs, such as choice of initialization, step-size, iteration/training time. There has been increasing interest in understanding the effects of implicit regularization (sometimes referred to as implicit bias) of machine learning algorithms. It is widely
acknowledged that the choice of algorithm, parametrization, and
parameter-tuning, all affect the learning performance of models derived from
training data. While implicit regularization
has been extensively investigated
in connection to the $\ell_2$ norm, there seem to be no results for
sparse regression, which is surprising considering the importance of the
problem.
\subsection{Our Contributions}
In this work, we merge statistics with optimization, and propose the first
statistically and computationally optimal algorithm based on implicit
regularization
(initialization/step-size tuning and early stopping)
for sparse linear regression under the RIP.

The algorithm that we propose is based on gradient descent applied to the unregularized, underdetermined objective function $\| \X \vec{w} - \vec{y} \|_2^2$ where $\vec{w}$ is
parametrized as $\vec{w}= \vec{u} \odot \vec{u} - \vec{v} \odot \vec{v}$, with
$\vec{u},\vec{v}\in\mathbb{R}^d$ and $\odot$ denotes the coordinate-wise
multiplication operator for vectors. This parametrization yields a non-convex
problem in $\vec{u}$ and $\vec{v}$. We treat this optimization problem as a proxy to design a sequence of statistical estimators that correspond to the iterates of gradient descent applied to solve the sparse regression problem, and hence are cheap to compute iteratively. The matrix formulation of the same type of
parametrization that we adopt has been recently considered in the setting of
low-rank matrix recovery
where it leads to exact recovery via implicit regularization in
the \emph{noiseless} setting under the RIP~\cite{gunasekar2017implicit, li2018algorithmic}. In our case, this choice of
parametrization yields an iterative algorithm that performs multiplicative
updates on the coordinates of $\vec{u}$ and $\vec{v}$, in contrast to the
additive updates obtained when gradient descent is run directly on the
parameter $\vec{w}$, as in proximal methods.
This feature allows us to reduce the convergence analysis to one-dimensional iterates and
to differentiate the convergence on the support set $S=\{i\in\{1,\ldots,d\}:w^\star_i \neq 0\}$ from the
convergence on its complement $S^c = \{1,\ldots,d\}\setminus S$.

We consider gradient descent initialized with $\vec{u}_0=\vec{v}_0=\alpha\vec{1}$, where $\vec{1}$ is the all-one vector. We show that with a sufficiently small initialization size $\alpha >0$ and early
stopping, our method achieves exact reconstruction with precision controlled by
$\alpha$ in the noiseless setting, and minimax-optimal rates in the noisy
setting. To the best of our knowledge, our results are the first to establish
non-$\ell_2$ implicit regularization for a gradient descent method in a general
\emph{noisy} setting.%
\footnote{However, see Remark~\ref{rem:concurrent_work} for work concurrent to ours that achieves similar goals.}
These results rely on a constant choice of step size
$\eta$ that satisfies a bound related to the unknown parameter $\wmax = \| \vec{w}^\star \|_\infty$. We show
how this choice of $\eta$ can be derived from the data itself, i.e.\ only based
on \emph{known} quantities.  If the noise vector $\vec{\xi}$ is made up of i.i.d.\
sub-Gaussian components with variance proxy $\sigma^2$, this
choice of $\eta$ yields $O((\wmax\sqrt{n})/(\sigma \sqrt{\log d})
\log \alpha^{-1})$ iteration complexity to achieve minimax rates. In order to achieve \emph{computational optimality}, we design a preconditioned version of gradient descent (on the parameters $\vec{u}$ and $\vec{v}$) that uses increasing step-sizes and has running time $\widetilde{O}(nd)$. The
iteration-dependent preconditioner relates to the statistical nature of the
problem. It is made by a sequence of diagonal matrices that implement a
coordinate-wise increasing step-size scheme that allows different coordinates to
accelerate convergence by taking larger steps based on refined estimates of the
corresponding coordinates of $\vec{w^{\star}}$. This algorithm yields
$O(\log ((\wmax\sqrt{n})/(\sigma \sqrt{\log d})) \log \alpha^{-1})$
iteration complexity to achieve minimax rates in the noisy
setting. Since each iteration costs $O(nd)$, the total computation
complexity is, up to poly-logarithmic factors, the same as simply
storing/reading the data. This algorithm is
minimax-optimal and, up to logarithmic factors, computationally optimal.
In contrast, we are not aware of any work on implicit $\ell_{2}$
regularization that exploits an increasing step sizes scheme in order to
attain computational optimality.

To support our theoretical results we present a simulation study of our methods and comparisons with the lasso
estimator and with the gold standard oracle least squares estimator, which performs least squares regression on $S$ assuming oracle knowledge of it.  We show that the number of iterations $t$ in our method plays a
role similar to the lasso regularization parameter $\lambda$. Despite both
algorithms being minimax-optimal with the right choice of $t$ and $\lambda$ respectively, the gradient descent optimization path---which is cheaper to compute as each iteration of gradient descent yields a new model---exhibits qualitative and
quantitative differences from the lasso regularization path---which is more expensive to compute as each model requires solving a new lasso optimization program. In particular, the simulations emphasize how the multiplicative updates allow gradient descent to fit one coordinate of $\vec{w}^\star$ at a time, as opposed to the lasso estimator that tends to fit all coordinates at once. Beyond minimax results, we prove that our methods adapt to instance difficulty:
for ``easy'' problems where the signal is greater than the noise, i.e.\ $\wmin
\gtrsim \inlinemaxnoise$ with $\wmin = \min_{i \in S} \abs{w^{\star}_{i}}$, our
estimators achieve the statistical rate $k \sigma^{2} \log (k)/n$, which does
\emph{not} depend on $d$. The experiments confirm this behavior and further
attest that our estimators undergo a phase transition that is not observed for
the lasso. Going from hard instances to easy ones, the learning capacity
of implicitly-regularized gradient descent exhibits a qualitative transition
and eventually matches the performance of oracle least squares.

\subfile{files/related_work.tex}

%% file: files/related_work.tex
\subsection{Related Work}
\label{section:related-work}

\paragraph{Sparse Recovery.}
The statistical properties of explicit $\ell_{1}$ penalization
techniques are well studied
\cite{
van2007deterministic,
candes2007dantzig,
bickel2009simultaneous,
meinshausen2009lasso,
negahban2012unified}.
Minimax rates for regression under sparsity constraints are derived in
\cite{raskutti2011minimax}.
Computing the whole lasso regularization path can be done via the
lars algorithm \cite{efron2004least}.
Another widely used approach is the glmnet
which uses cyclic coordinate-descent with warm starts
to compute regularization paths for generalized linear models with convex
penalties on a pre-specified grid of regularization parameters
\cite{friedman2010regularization}.
\cite{bach2012optimization} reviews various optimization techniques used in solving empirical risk
minimization problems with sparsity inducing penalties.
Using recent advances in mixed integer optimization,
\cite{bertsimas2016best} shows that the best subset selection problem can
be tackled for problems of moderate size.
For such problem sizes, comparisons between the lasso and best subset selection
problem ($\ell_{0}$ regularization)
were recently made, suggesting that the best subset selection performs
better in high signal-to-noise ratio regimes whereas the lasso performs
better when the signal-to-noise ratio is low~\cite{hastie2017extended}.
In this sense, our empirical study in Section~\ref{section:simulations}
suggests that implicitly-regularized gradient descent
is more similar to $\ell_{0}$ regularization than $\ell_{1}$ regularization.
Several other techniques related to $\ell_{1}$ regularization and extensions to the
lasso exist. We refer the interested reader to the books
\cite{
buhlmann2011statistics,
tibshirani2015statistical}.

\paragraph{Implicit Regularization/Bias.}
Connections between $\ell_{2}$ regularization and gradient descent
optimization paths have been known for a long time and are well studied
\cite{
buhlmann2003boosting,
friedman2004gradient,
yao2007early,
bauer2007regularization,
raskutti2014early,
wei2017early,
neu2018iterate,
suggala2018connecting,
ali2018continuous}.
In contrast, the literature on implicit regularization inducing sparsity is
scarce.
Coordinate-descent optimization paths have been shown to be
related to $\ell_{1}$ regularization paths in some regimes
\cite{
friedman2001elements,
efron2004least,
rosset2004boosting,
zhang2005boosting}.
Understanding such connections can potentially allow
transferring the now well-understood theory developed for penalized forms of
regularization to early-stopping-based regularization which can result in lower
computational complexity.
Recently, \cite{zhang2016understanding} have shown that neural networks
generalize well even without explicit regularization despite the capacity
to fit unstructured noise.
This suggests that some implicit regularization effect is limiting
the capacity of the obtained models along the optimization path and thus
explaining generalization on structured data.
Understanding such effects has recently drawn a lot of attention in the machine
learning community.
In particular, it is now well understood that the optimization algorithm itself
can be biased towards a particular set of solutions
for underdetermined problems with many global minima
where, in contrast to the work cited above, the bias of optimization algorithm
is investigated at or near convergence, usually in a noiseless setting
\cite{soudry2018implicit, gunasekar2018implicit,
gunasekar2018characterizing, gunasekar2017implicit, li2018algorithmic}.
We compare our assumptions with the ones made in \cite{li2018algorithmic}
in Appendix~\ref{appendix:comparing-with-colt-paper}.

\begin{remark}[Concurrent Work] \label{rem:concurrent_work} After completing this work we became aware of independent concurrent work
\cite{zhao2019implicit}
which considers Hadamard product reparametrization $\vec{w}_{t} = \vec{u}_{t}
\odot \vec{v}_{t}$ in order to implicitly induce sparsity for linear
regression under the RIP assumption. Our work is significantly different in many aspects
discussed in Appendix~\ref{appendix:comparing-with-hadamard-product-paper}.
In particular, we obtain computational optimality and can properly handle
the general noisy setting.
\end{remark}

%% file: files/preliminaries.tex
\section{Model and Algorithms}
\label{section:preliminaries}

We consider the model defined in the introduction.
We denote vectors with boldface letters and real numbers with normal font;
thus, $\vec{w}$ denotes a vector and
$w_{i}$ denotes the $i\th$ coordinate of $\vec{w}$.
For any index set $A$
we let $\id{A}$ denote a vector that has a $1$ entry in all coordinates $i \in A$
and a $0$ entry elsewhere.
We denote coordinate-wise inequalities by
$\preccurlyeq$. With a slight abuse of notation we write
$\vec{w}^{2}$ to mean
the vector obtained by squaring each component of $\vec{w}$.
Finally, we denote inequalities up to multiplicative absolute constants,
meaning that they do not depend on any parameters of the problem, by
$\lesssim$.
A table of notation can be found in
Appendix~\ref{appendix:table-of-notation}.

We now define the restricted isometry property which is the key assumption
in our main theorems.

\begin{definition}[Restricted Isometry Property (RIP)]
  \label{assumption:rip}
  A $n \times d$ matrix $\X / \sqrt{n}$
  satisfies the ($\delta, k$)-(RIP)
  if for any $k$-sparse vector
  $\vec{w} \in \mathbb{R}^{d}$ we have
  $
    (1 - \delta) \norm{\vec{w}}_{2}^{2}
    \leq
    \norm{\X \vec{w} / \sqrt{n}}_{2}^{2}
    \leq
    (1 + \delta) \norm{\vec{w}}_{2}^{2}.
  $
\end{definition}

The RIP assumption was introduced in \cite{candes2005decoding} and is standard
in the compressed sensing literature.
It requires that all $n \times k$
sub-matrices of $\X/\sqrt{n}$ are approximately orthonormal
where $\delta$ controls extent to which this approximation holds.
Checking if a given matrix satisfies the RIP is NP-hard~\cite{bandeira2013certifying}.
In compressed sensing applications the matrix $\X / \sqrt{n}$ corresponds
to how we measure signals and it can be chosen by the
designer of a sparse-measurement device.
Random matrices are known to satisfy the RIP with high probability,
with $\delta$ decreasing to $0$ as $n$ increases for a fixed $k$
\cite{baraniuk2008simple}.

We consider the following problem setting.
Let $\vec{u}, \vec{v} \in \mathbb{R}^{d}$ and define the mean squared loss as
\begin{equation*}
  \label{eq:reparameterized-loss}
  \mathcal{L}(\vec{u}, \vec{v})
  = \frac{1}{n} \norm{\X\left(
  \vec{u} \odot \vec{u} - \vec{v} \odot \vec{v}\right) - \vec{y}}^{2}_{2}.
\end{equation*}
Letting
$\vec{w} = \vec{u} \odot \vec{u} - \vec{v} \odot \vec{v}$
and performing gradient descent updates on $\vec{w}$, we recover
the original parametrization of mean squared error loss which does not
implicitly induce sparsity.
Instead,
we perform gradient descent updates on $(\vec{u}, \vec{v})$ treating it
as a vector in $\mathbb{R}^{2d}$
and we show that the corresponding optimization path contains sparse
solutions.

Let $\eta > 0$ be the learning rate,
$(\vec{m}_{t})_{t\geq 0}$ be a sequence of vectors in $\mathbb{R}^{d}$
and $\operatorname{diag}(\vec{m}_{t})$ be a $d \times d$ diagonal matrix
with $\vec{m}_{t}$ on its diagonal.
We consider the following general form of gradient descent:
\begin{equation}
  \label{equation:gradient-descent}
  (\vec{u}_{t + 1}, \vec{v}_{t + 1})
  =
  (\vec{u}_{t}, \vec{v}_{t})
  - \eta \operatorname{diag}(\vec{m}_{t}, \vec{m}_{t})
  \frac{\partial \mathcal{L}(\vec{u}_{t}, \vec{v}_{t})}
  {\partial (\vec{u}_{t}, \vec{v}_{t})}.
\end{equation}

We analyze two different choices of
sequences $(\vec{m}_{t})_{t \geq 0}$ yielding two separate algorithms.

\begin{algorithm}
  \label{alg:gd}
  Let $\alpha, \eta > 0$ be two given parameters.
  Let $\vec{u}_{0} = \vec{v}_{0} = \alpha$ and for all $t \geq 0$
  we let $\vec{m}_{t} = \id{}$.
  Perform the updates given in \eqref{equation:gradient-descent}.
\end{algorithm}

\begin{algorithm}
  \label{alg:gd-increasing-steps}
  Let $\alpha$, $\tau \in \mathbb{N}$ and
  $\wmax \leq \hat{z} \leq 2\wmax$
  be three given parameters.
  Set $\eta = \frac{1}{20 \hat{z}}$ and $\vec{u}_{0} = \vec{v}_{0} = \alpha$.
  Perform the updates in \eqref{equation:gradient-descent}
  with $\vec{m}_{0} = \id{}$ and
  $\vec{m}_{t}$ adaptively defined as follows:
  \begin{enumerate}
    \item Set $\vec{m}_{t} = \vec{m}_{t-1}$.
    \item If $t = m \tau \ceil{ \log \alpha^{-1}}$ for some
      natural number $m \geq 2$ then
      let $m_{t,j} = 2 m_{t-1,j}$
      for all $j$ such that
      $u_{t,j}^{2} \vee v_{t,j}^{2} \leq 2^{-m-1}\hat{z}$.
  \end{enumerate}
\end{algorithm}

Algorithm~\ref{alg:gd} corresponds to gradient descent with a constant
step size, whereas Algorithm~\ref{alg:gd-increasing-steps}
doubles the step-sizes for small enough coordinates
after every $\tau \ceil{ \log \alpha^{-1}}$ iterations.

Before stating the main results we define some key quantities.
First, our results are sensitive to the condition number
$\kappa = \kappa(\wstar) = \wmax / \wmin$
of the true parameter vector $\wstar$.
Since we are not able to recover coordinates below the maximum noise term
$\inlinemaxnoise$, for a desired precision $\varepsilon$
we can treat
all coordinates of $\wstar$ below
$\varepsilon \vee (\inlinemaxnoise)$ as $0$. This motivates the following definition
of an effective condition number for given $\wstar, \X, \xi$ and $\varepsilon$:
\begin{align*}
   \kappa^{\text{eff}}
   =
   \kappa^{\text{eff}}(\wstar, \X, \xi, \varepsilon)
   = \wmax / (\wmin \vee \varepsilon \vee (\inlinemaxnoise)).
\end{align*}
We remark that $\kappa^{\text{eff}}(\wstar, \X, \xi, \varepsilon) \leq \kappa(\wstar)$.
Second, we need to put restrictions on the RIP constant
$\delta$ and initialization size $\alpha$.
These restrictions are given by the following:
\begin{align*}
  \delta ( k, \wstar, \X, \xi, \varepsilon )
  =
  1 / (\sqrt{k}
  ( 1 \vee \log \kappa^{\text{eff}}(\wstar) )
  ),
  \quad
  \alpha(\wstar, \varepsilon, d) \coloneqq
  \frac{\varepsilon^{2} \wedge \varepsilon \wedge 1}
  {(2d + 1)^{2} \vee (\wmax)^{2}} \wedge \frac{\sqrt{\wmin}}{2}.
\end{align*}

%% file: files/main_results.tex
\section{Main Results}
\label{section:main-results}

The following result is the backbone of our contributions.
It establishes rates for Algorithm~\ref{alg:gd} in the $\ell_\infty$ norm
as opposed to the typical rates
for the lasso that are often only derived for the $\ell_2$ norm.

\begin{theorem}
\label{thm:constant-step-sizes}
Fix any $\varepsilon > 0$.
Suppose that $\X / \sqrt{n}$ satisfies the $(k+1, \delta)$-RIP
with
$\delta \lesssim \delta \left( k, \wstar, \X, \xi, \varepsilon \right)$
and let the initialization $\alpha$ satisfy
$\alpha \leq \alpha(\wstar, \varepsilon, d)$.
Then, Algorithm~\ref{alg:gd} with $\eta \leq 1/(20 \wmax)$ and
$
  t = O (
    (\kappa^{\text{eff}}(\wstar))/(\eta \wmax) \log \alpha^{-1}
  )
$
iterations satisfies
\begin{align*}
    \abs{w_{t,i} - w^{\star}_{i}}
    &\lesssim
    \begin{cases}
      \maxnoise \vee \varepsilon
       & \text{if } i \in S \text{ and }
        \wmin \lesssim \maxnoise \vee \varepsilon, \\
      \abs{\frac{1}{n} \left(\Xt \vec{\xi}\right)_{i}}
      \vee \delta \sqrt{k} \norm{\noisevec \odot \id{S}}_{\infty}
      \vee \varepsilon
       & \text{if } i \in S \text{ and }
        \wmin \gtrsim \maxnoise \vee \varepsilon, \\
      \sqrt{\alpha}
       & \text{if } i \notin S.
    \end{cases}
\end{align*}
\end{theorem}
This result shows how the parameters
$\alpha, \eta$ and $t$ affect the learning
performance of gradient descent.
The size of $\alpha$ controls
the size of the coordinates outside the true support $S$
at the stopping time.
We discuss the role  and also the necessity of small initialization size to
achieve the desired statistical performance in Section~\ref{section:simulations}.
A different role is played by the step size $\eta$
whose size affects the optimal stopping time $t$.
In particular, $(\eta t)/\log \alpha^{-1}$
can be seen as a regularization parameter closely related to $\lambda^{-1}$
for the lasso.
To see this, suppose that the noise $\xi$ is $\sigma^{2}$-sub-Gaussian with independent
components. Then with high probability $\inlinemaxnoise \lesssim (\sigma
\sqrt{\log d})/\sqrt{n})$.
In such a setting an optimal choice of $\lambda$ for the lasso
is $\Theta((\sigma \sqrt{\log d})/\sqrt{n})$.
On the other hand,
letting $t^{\star}$ be the optimal stopping time given in
Theorem~\ref{thm:constant-step-sizes},
we have
$(\eta t^{\star}) / \log \alpha^{-1}
= O(1/\wmin(\X, \xi, \varepsilon))
= O(\sqrt{n}/(\sigma \sqrt{\log d}))$.

The condition $\eta \leq 1/(20\wmax)$ is also necessary up to
constant factors in order to prevent explosion.
If we can set
$1/\wmax \lesssim \eta \leq 1/(20\wmax)$
then the iteration complexity of Theorem~\ref{thm:constant-step-sizes}
reduces to $O(\kappa^{\text{eff}}(\wstar) \log \alpha^{-1})$.
The magnitude of $\wmax$ is, however, an unknown quantity.
Similarly, setting the proper initialization size $\alpha$ depends on
$\wmax$, $\wmin$, $d$ and the desired precision $\varepsilon$.
The requirement that $\alpha \leq \sqrt{\wmin}/2$ is an artifact
of our proof technique and tighter analysis could replace this condition
by simply $\alpha \leq \varepsilon$. Hence the only unknown
quantity for selecting a proper initialization size is $\wmax$.

The next theorem shows how $\wmax$ can be estimated from the data up to a
multiplicative factor $2$ at the cost of one gradient descent iteration.
Once this estimate is computed, we can properly set the initialization size
and the learning rate $\eta \asymp \frac{1}{\wmax}$
which satisfies our theory and is tight up to constant multiplicative factors.
We remark that $\tilde{\eta}$ used in Theorem~\ref{thm:selecting-the-step-size}
can be set arbitrarily small (e.g., set $\tilde{\eta} = 10^{-10}$)
and is only used for one gradient descent step in order to estimate $\wmax$.

\begin{theorem}[Estimating $\wmax$]
  \label{thm:selecting-the-step-size}
  Set $\alpha = 1$ and suppose that $\X/\sqrt{n}$
  satisfies the
  $(k+1, \delta)$-RIP with
  $
    \delta
    \leq
    1/(20\sqrt{k}).
  $
  Let the step size $\tilde{\eta}$ be any number satisfying
  $0 < \tilde{\eta} \leq 1/(5 \wmax)$
  and suppose that $\wmax \geq 5\inlinemaxnoise$.
  Perform one step of gradient descent and for each $i \in \{1, \dots, d\}$
  compute the update factors defined as
  $f^{+}_{i} = (u_{1})_{i}$ and
  $f^{-}_{i} = (v_{1})_{i}$.
  Let $f_{\max} = \norm{\vec{f}^{+}}_{\infty} \vee
  \norm{\vec{f}^{-}}_{\infty}$. Then
  $
    \wmax
    \leq
    (f_{\max} - 1)/(3\tilde{\eta})
    <
    2\wmax.
  $
\end{theorem}

We present three main corollaries of
Theorem~\ref{thm:constant-step-sizes}.
The first one shows that in the noiseless setting
exact recovery is possible
and is
controlled by the desired precision
$\varepsilon$ and hence by
the initialization size $\alpha$.

\begin{corollary}[Noiseless Recovery]
  \label{corollary:noiseless}
  Let $\xi = 0$. Under the assumptions of
  Theorem~\ref{thm:constant-step-sizes},
  the choice of $\eta$ given by Theorem~\ref{thm:selecting-the-step-size}
  and $t = O(\kappa^{\text{eff}}(\wstar) \log \alpha^{-1})$,
  Algorithm~\ref{alg:gd} yields
  $\norm{\vec{w}_{t} - \wstar}_{2}^{2} \lesssim k \varepsilon^{2}$.
\end{corollary}

In the general noisy setting exact reconstruction of $\wstar$ is not possible.
In fact, the bounds in Theorem~\ref{thm:constant-step-sizes}
do not improve with $\varepsilon$ chosen below
the maximum noise term $\inlinemaxnoise$.
In the following corollary we show that with a small enough $\varepsilon$
if the design matrix $\X$ is fixed and the noise vector $\xi$
is sub-Gaussian,
we recover minimax-optimal rates for $\ell_{2}$ error.
Our error bound is minimax-optimal
in the setting of sub-linear sparsity, meaning that there
exists a constant $\gamma > 1$ such that $k^{\gamma} \leq d$.

\begin{corollary}[Minimax Rates in the Noisy Setting]
  \label{corollary:minimax}
  Let the noise vector $\xi$ be made of independent
  $\sigma^{2}$-sub-Gaussian entries.
  Let $\varepsilon = 4\sqrt{\sigma^{2} \log (2d)} / \sqrt{n}$.
  Under the assumptions of Theorem~\ref{thm:constant-step-sizes},
  the choice of $\eta$ given by Theorem~\ref{thm:selecting-the-step-size}
  and
  $t = O(\kappa^{\text{eff}}(\wstar) \log \alpha^{-1})
     = O((\wmax \sqrt{n})/(\sigma \sqrt{\log d}) \log \alpha^{-1})$,
  Algorithm~\ref{alg:gd} yields
  $\norm{\vec{w}_{t} - \wstar}_{2}^{2} \lesssim
  (k \sigma^{2} \log d)/n$
  with probability at least
  $1 - 1/(8d^{3})$.
\end{corollary}

The next corollary states that gradient descent automatically adapts to the
difficulty of the problem.
The statement of Theorem~\ref{thm:constant-step-sizes}
suggests that our bounds undergo a phase-transition
when $\wmin \gtrsim \inlinemaxnoise$ which is also supported by our
empirical findings in Section~\ref{section:simulations}. In the
$\sigma^{2}$-sub-Gaussian noise setting the transition occurs as soon as
$n \gtrsim (\sigma^{2} \log d) / (\wmin)^{2}$.
As a result, the statistical bounds achieved by our algorithm are independent
of $d$ in such a setting.
To see that, note that while the term $\inlinemaxnoise$ grows as $O(\log d)$,
the term $\inlinenorm{\Xt\vec{\xi} \odot \id{S}}_{\infty}/n$ grows only as $O(\log k)$.
In contrast, performance of the lasso deteriorates with $d$ regardless of the
difficulty of the problem. We illustrate this graphically and give
a theoretical explanation in Section~\ref{section:simulations}.
We remark that the following result does not contradict minimax optimality
because we now treat the true parameter $\wstar$ as fixed.

\begin{corollary}[Instance Adaptivity]
  \label{corollary:adaptivity}
  Let the noise vector $\xi$ be made of independent
  $\sigma^{2}$-sub-Gaussian entries.
  Let $\varepsilon = 4\sqrt{\sigma^{2} \log (2k)} / \sqrt{n}$.
  Under the assumptions of Theorem~\ref{thm:constant-step-sizes},
  the choice of $\eta$ given by Theorem~\ref{thm:selecting-the-step-size}
  and
  $t = O(\kappa^{\text{eff}}(\wstar) \log \alpha^{-1})
     = O((\wmax \sqrt{n})/(\sigma \sqrt{\log k}) \log \alpha^{-1})$,
  Algorithm~\ref{alg:gd} yields
  $\norm{\vec{w}_{t} - \wstar}_{2}^{2} \lesssim (k \sigma^{2} \log k)/n$.
  with probability at least
  $1 - 1/(8k^{3})$.
\end{corollary}

The final theorem we present shows that the same statistical bounds
achieved by Algorithm~\ref{alg:gd}
are also attained by
Algorithm~\ref{alg:gd-increasing-steps}.
This algorithm is not only optimal in a statistical sense, but it is also
optimal computationally up to poly-logarithmic factors.

\begin{theorem}
  \label{thm:main-theorem-noisy-minimax-rates-exponential-convergence}
  Compute $\hat{z}$ using Theorem~\ref{thm:selecting-the-step-size}.
  Under the setting of Theorem~\ref{thm:constant-step-sizes}
  there exists a large enough absolute constant $\tau$
  so that Algorithm~\ref{alg:gd-increasing-steps} parameterized
  with $\alpha, \tau$ and $\hat{z}$
  satisfies the result of Theorem~\ref{thm:constant-step-sizes}
  and
  $
    t = O (
      \log \kappa^{\text{eff}}
      \log \alpha^{-1}
    )
  $
  iterations.
\end{theorem}

Corollaries~\ref{corollary:noiseless},
\ref{corollary:minimax} and \ref{corollary:adaptivity}
also hold for
Algorithm~\ref{alg:gd-increasing-steps}
with stopping time equal to
$O(\log \kappa^{\text{eff}} \log \alpha^{-1})$.
We emphasize that both Theorem~\ref{thm:constant-step-sizes}
and \ref{thm:main-theorem-noisy-minimax-rates-exponential-convergence}
use gradient-based updates to obtain a sequence of models with
optimal statistical properties instead of optimizing the objective
function $\mathcal{L}$.
In fact, if we let $t \to \infty$ for Algorithm~\ref{alg:gd-increasing-steps}
the iterates would explode.

%% file: files/proof_sketch.tex
\section{Proof Sketch}
\label{section:proof-sketch}

In this section we prove a simplified version of
Theorem~\ref{thm:constant-step-sizes} under the assumption $\XtX/n = \matrixid$.
We further highlight the intricacies involved in the general setting
and present the intuition behind the key ideas there.
The gradient descent updates
on $\vec{u}_{t}$ and $\vec{v}_{t}$ as given in
\eqref{equation:gradient-descent} can be written as
\begin{align}
  \begin{split}
   \label{eqn:uv-updates}
	   \vec{u}_{t+1}
     = \vec{u}_{t} \odot \left(\id{} - (4\eta/n)\Xt(\X\vec{w}_{t} - \vec{y})\right),
	   &\quad
     \vec{v}_{t+1}
     = \vec{v}_{t} \odot \left(\id{} + ({4 \eta}/{n})\Xt(\X\vec{w}_{t} -
     \vec{y}) \right).
  \end{split}
\end{align}
The updates can be succinctly represented as $\vec{u}_{t+1} = \vec{u}_t \odot
(\id{} - \vec{r})$ and $\vec{v}_{t+1} = \vec{v}_t \odot (\id{} + \vec{r})$,
where by our choice of $\eta$, $\norm{\vec{r}}_\infty \leq 1$. Thus, $(\id{} - \vec{r}) \odot (\id{} +
\vec{r}) \preccurlyeq \id{}$ and we have $\vec{u}_t \odot \vec{v}_t \preccurlyeq
\vec{u}_0 \odot \vec{v}_0 = \alpha^2 \id{}$.
Hence for any $i$, only one of $|u_{t,
i}|$ and $|v_{t, i}|$ can be larger then the initialization size
while the other is effectively equal to $0$. Intuitively,
$u_{t, i}$ is used if $w^\star_i > 0$, $v_{t, i}$ if $w^\star_i < 0$
and hence one of these terms can be merged into an error term
$b_{t,i}$ as defined below.
The details appear in Appendix~\ref{appendix:multiplicative-updates:negative-targets}.
To avoid getting lost in cumbersome notation,
in this section we will assume that
$\wstar \succcurlyeq 0$ and
$\vec{w} = \vec{u} \odot \vec{u}$.

\begin{theorem}
  \label{thm:simplified-version}
	Assume that
  $\wstar \succcurlyeq \vec{0}$,
  $\frac{1}{n}\XtX = \matrixid$,
  and that there is no noise ($\vxi = 0$).
  Parameterize $\vw = \vu \odot \vu$ with
  $\vu_0 = \alpha \vone$ for some $0 < \alpha < \sqrt{\wmin}$.
  Letting $\eta \leq 1/(10\wmax)$ and
  $t = O(\log(\wmax/\alpha^{2})/(\eta \wmin))$,
  Algorithm~\ref{alg:gd} yields
  $\inlinenorm{\vec{w}_{t} - \wstar}_{\infty} \leq \alpha^{2}$.
\end{theorem}

\begin{proof}
  As $\XtX/n = \matrixid, \vec y = \X \vec w^\star$, and $\vec{v}_{t} =
  \vec{0}$,
  the updates given in equation~\eqref{eqn:uv-updates}
  reduce component-wise to updates on $\vec{w}_{t}$ given by
  $
	  w_{t + 1, i} = w_{t, i} \cdot (1 - 4 \eta (w_{t, i} - w^\star_i))^2.
  $
  For $i$ such that $w^\star_i = 0$, $w_{t, i}$ is non-increasing and hence
  stays below $\alpha^2$. For $i$ such that $w^\star_i > \alpha^2$, the update
  rule given above ensures that as long as
  $w_{t, i} < w^\star_i / 2$, $w_{t, i}$ increases at an exponential rate with
  base at least $(1 + 2 \eta w^\star_i)^2$. As $w_{0, i} = \alpha^2$, in
  $O(\log(w^\star_i/\alpha^{2})/(\eta w^\star_i))$ steps, it holds that $w_{t, i}
  \geq w^\star_i/2$.  Subsequently, the gap $(w^\star_i - w_{t, i})$ halves
  every $O(1/\eta w^\star_i)$ steps; thus, in $O(\log(\wmax/\alpha^{2})/(\eta
  \wmin))$ steps we have $\inlinenorm{\vec{w}_{t} - \wstar}_{\infty} \leq
  \alpha^{2}$. The exact details are an exercise in calculus, albeit a
  rather tedious one, and appear in
  Appendix~\ref{appendix:multiplicative-updates:basic-lemmas}.
\end{proof}

The proof of Theorem~\ref{thm:simplified-version} contains the key ideas of the
proof of Theorem~\ref{thm:constant-step-sizes}.
However, the presence of noise
($\vec{\xi} \neq 0$) and only having \emph{restricted isometry} of $\XtX$
rather than \emph{isometry} requires a subtle and involved analysis.
We remark that we can prove tighter bounds in Theorem~\ref{thm:simplified-version} 
than the ones in
Theorem~\ref{thm:constant-step-sizes}
because we are working in a simplified setting.

\paragraph{Error Decompositions.} We decompose
$\vec{w}_{t}$ into
$\vec{s}_t \coloneqq \vec{w}_t \odot \id{S}$
and
$\vec{e}_t \coloneqq \vec{w}_t \odot \id{S^c}$ so that
$\vec{w}_{t} = \vec{s}_{t} + \vec{e}_{t}$.
We define the following error sequences:
\begin{align*}
  \vb_{t} = {\XtX}\ve_{t}/n + {\Xt\vxi}/{n},
  &\qquad\vp_{t} = \left({\XtX}/{n} - \matrixid \right) (\vs_{t} - \wstar),
\end{align*}
which allows us to write updates on $\vec{s}_{t}$ and $\vec{e}_{t}$
as
$$
	\vs_{t+1} = \vs_{t} \odot (1 - 4\eta(\vs_{t} - \vw^{*} + \vp_{t} +
  \vb_{t}))^{2},
  \qquad
	\ve_{t+1} = \ve_{t} \odot (1 - 4\eta(\vp_{t} + \vb_{t}))^{2}.
$$
\paragraph{Error Sequence $\vec{b}_{t}$.}
Since our theorems require stopping before $\norm{\vec{e}_{t}}_{\infty}$
exceeds $\sqrt{\alpha}$, the term $\XtX\vec{e}_{t}/n$ can be controlled
entirely by the initialization size.
Hence $\vec{b}_{t} \approx \Xt\vec{\xi}/n$ and it represents an
\emph{irreducible} error arising due to the noise on the labels.
For any $i \in S$ at stopping time $t$ we cannot expect
the error on the $i\th$ coordinate
$\inlineabs{w_{t,i} - w^\star_{i}}$
to be smaller than
$\inlineabs{(\Xt \vec{\xi})/n)_{i}}$.
If we assume $\vec{p}_{t} = 0$ and $\xi \neq 0$ then in light of our simplified
Theorem~\ref{thm:simplified-version} we see that
the terms in $\vec{e}_{t}$ grow exponentially with
base at most $(1 + 4\eta\inlinemaxnoise)$.
We can fit all the terms in $\vec{s}_{t}$ such that
$\inlineabs{w^{\star}_{i}} \gtrsim \inlinemaxnoise$ which leads to
minimax-optimal rates.
Moreover, if $\wmin \gtrsim \inlinemaxnoise$
then all the elements in $\vec{s}_{t}$ grow exponentially at a faster rate
than all of the error terms. This corresponds to the easy setting where
the resulting error depends only on $\inlinenorm{\Xt\vec{\xi}/n \odot
  \id{S}}_{\infty}$ yielding dimension-independent error bounds.
For more details see Appendix~\ref{appendix:multiplicative-updates:bounded-errors}.

\paragraph{Error Sequence $\vec{p}_{t}$.}
Since $\vec{s}_{t} - \wstar$ is a $k$-sparse vector
using the RIP we can upper-bound
$\norm{\vec{p}_{t}}_{\infty} \leq \sqrt{k}\delta\norm{\vec{s}_{t} -
\wstar}_{\infty}$.
Note that for small $t$ we have
$\norm{\vec{s}_{0}}_{\infty} \approx \alpha^{2} \approx 0$
and hence, ignoring the logarithmic factor in the definition
of $\delta$ in the worst case we have
$C\wmax \leq \norm{\vec{p}_{t}}_{\infty} < \wmax$
for some absolute constant $0 < C < 1$.
If $\wmax \gg \inlinemaxnoise$
then the error terms can grow exponentially with base
$(1 + 4\eta \cdot C\wmax)$
whereas the signal terms such that $\inlineabs{w^\star_{i}} \ll \wmax$
can shrink exponentially at rate
$(1 - 4\eta \cdot C\wmax)$.
On the other hand, in the light of Theorem~\ref{thm:simplified-version}
the signal elements converge exponentially fast to the true parameters
$\wmax$ and hence the error sequence $\vec{p}_{t}$
should be exponentially decreasing.
For small enough $C$ and a careful choice of initialization size
$\alpha$ we can ensure that elements of $\vec{p}_{t}$
decrease before the error components in $\vec{e}_{t}$ get too large or the signal
components in $\vec{s}_{t}$ get too small.
For more details see
Appendix~\ref{appendix:main-proofs:key-propositions}
and \ref{appendix:multiplicative-updates:dealing-with-rip-errors}.

\paragraph{Tuning Learning Rates.}
The proof of Theorem~\ref{thm:selecting-the-step-size} is given in
Appendix~\ref{appendix:selecting-the-step-size}.
If we choose $1/\wmax \lesssim \eta \leq 1/(10\wmax)$ in
Theorem~\ref{thm:simplified-version}
then all coordinates converge in
$O(\kappa \log(\wmax/\alpha^{2}))$ iterations.
The reason the factor $\kappa$ appears is the need to
ensure that the convergence of the component $w^\star_i = \wmax$ is stable.
However, this conservative setting of the learning rate unnecessarily
slows down the convergence for components with $w^\star_i \ll \wmax$.
In Theorem~\ref{thm:simplified-version}, oracle knowledge of $\wstar$
would allow to set an individual step size for each coordinate $i \in S$
equal to $\eta_{i} = 1/(10w^\star_i)$
yielding the total number of iterations equal to
$O(\log(\wmax/\alpha^{2}))$.
In the setting where $\XtX/n \neq \matrixid$ this would not be possible
even with the knowledge of $\wstar$, since the error sequence $\vec{p}_{t}$
can be initially too large which would result in explosion of the
coordinates $i$ with $\inlineabs{w^\star_{i}} \ll \wmax$.
Instead, we need to wait for $\vec{p}_{t}$ to get small enough before we
increase the step size for some of the coordinates as described
in Algorithm~\ref{alg:gd-increasing-steps}.
The analysis is considerably involved
and 
the full proof can be found
in Appendix~\ref{appendix:theorem-minimax-rates-exponential-convergence-proof}.
We illustrate effects of increasing step sizes in
Section~\ref{section:simulations}.

%% file: files/simulations.tex
\section{Simulations}
\label{section:simulations}

Unless otherwise specified, the default simulation set up is as follows.
We let $\wstar = \gamma\id{S}$ for some constant $\gamma$.
For each run the entries of $\X$ are sampled as i.i.d.\ Rademacher random
variables and the noise vector $\vec{\xi}$ follows i.i.d.\ $N(0, \sigma^{2})$
distribution.
For $\ell_{2}$ plots each simulation is
repeated a total of $30$ times and the median $\ell_{2}$ error is depicted.
The error bars in all the plots denote the
$25\th$ and $75\th$ percentiles.
Unless otherwise specified, the default values for simulation parameters are
$n = 500, d = 10^4, k = 25, \alpha = 10^{-12}, \gamma = 1, \sigma = 1$
and for Algorithm~\ref{alg:gd-increasing-steps} we set $\tau = 10$.

\begin{figure}[h]
  \centering
  \begin{subfigure}{.33\textwidth}
    \centering
    \includegraphics[width=\linewidth]{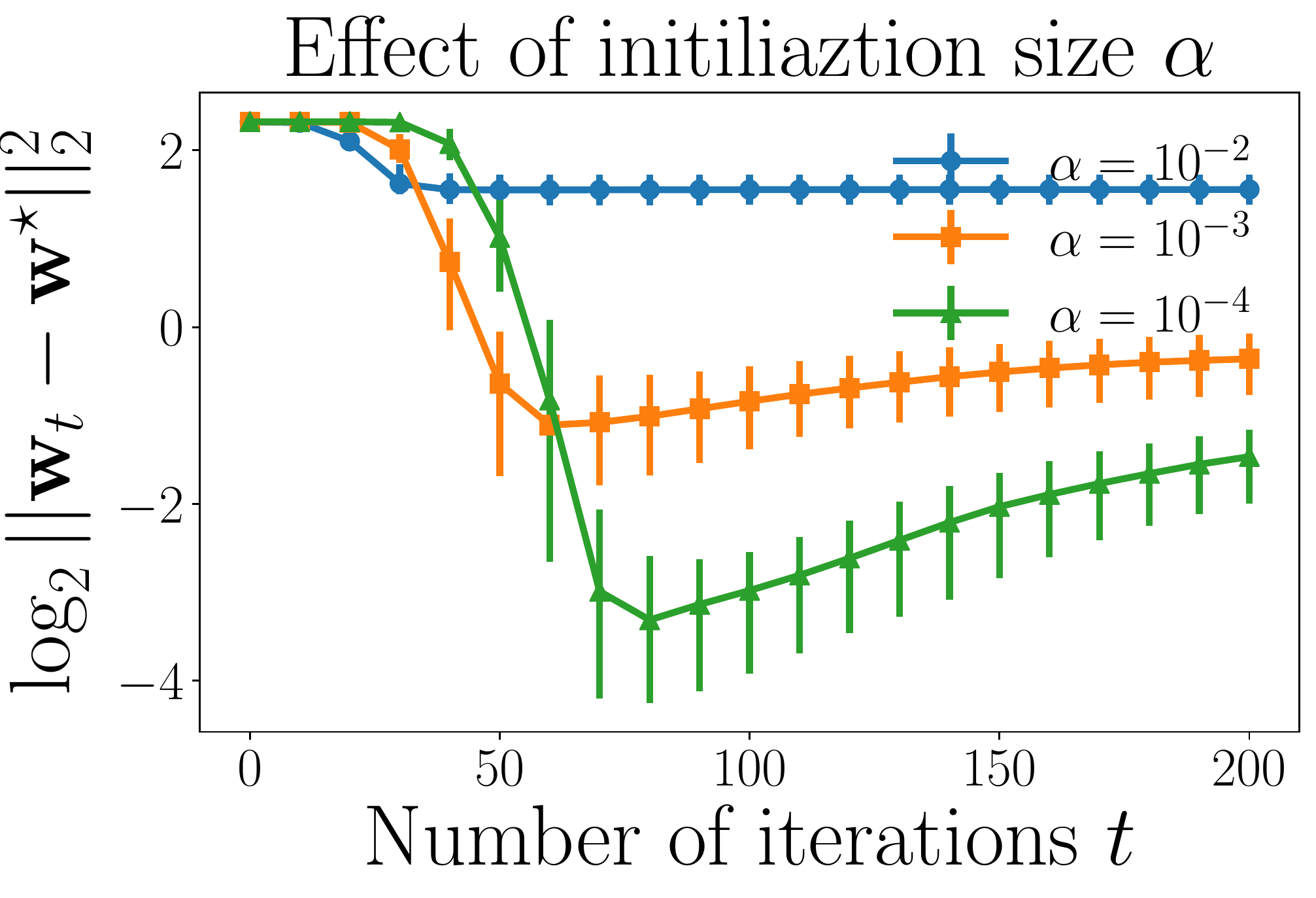}
  \end{subfigure}%
  \begin{subfigure}{.33\textwidth}
    \centering
    \includegraphics[width=\linewidth]{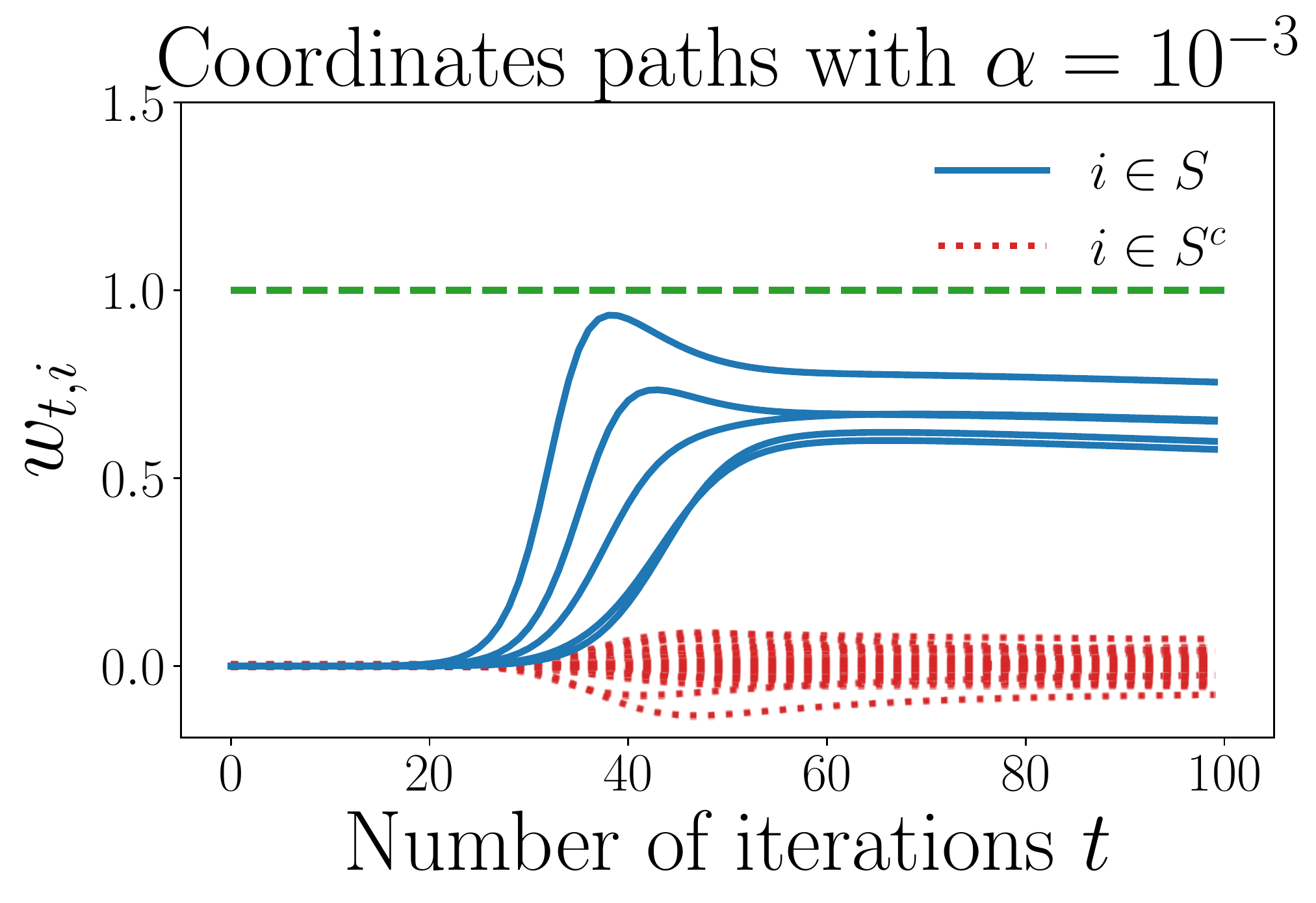}
  \end{subfigure}%
  \begin{subfigure}{.33\textwidth}
    \centering
    \includegraphics[width=\linewidth]{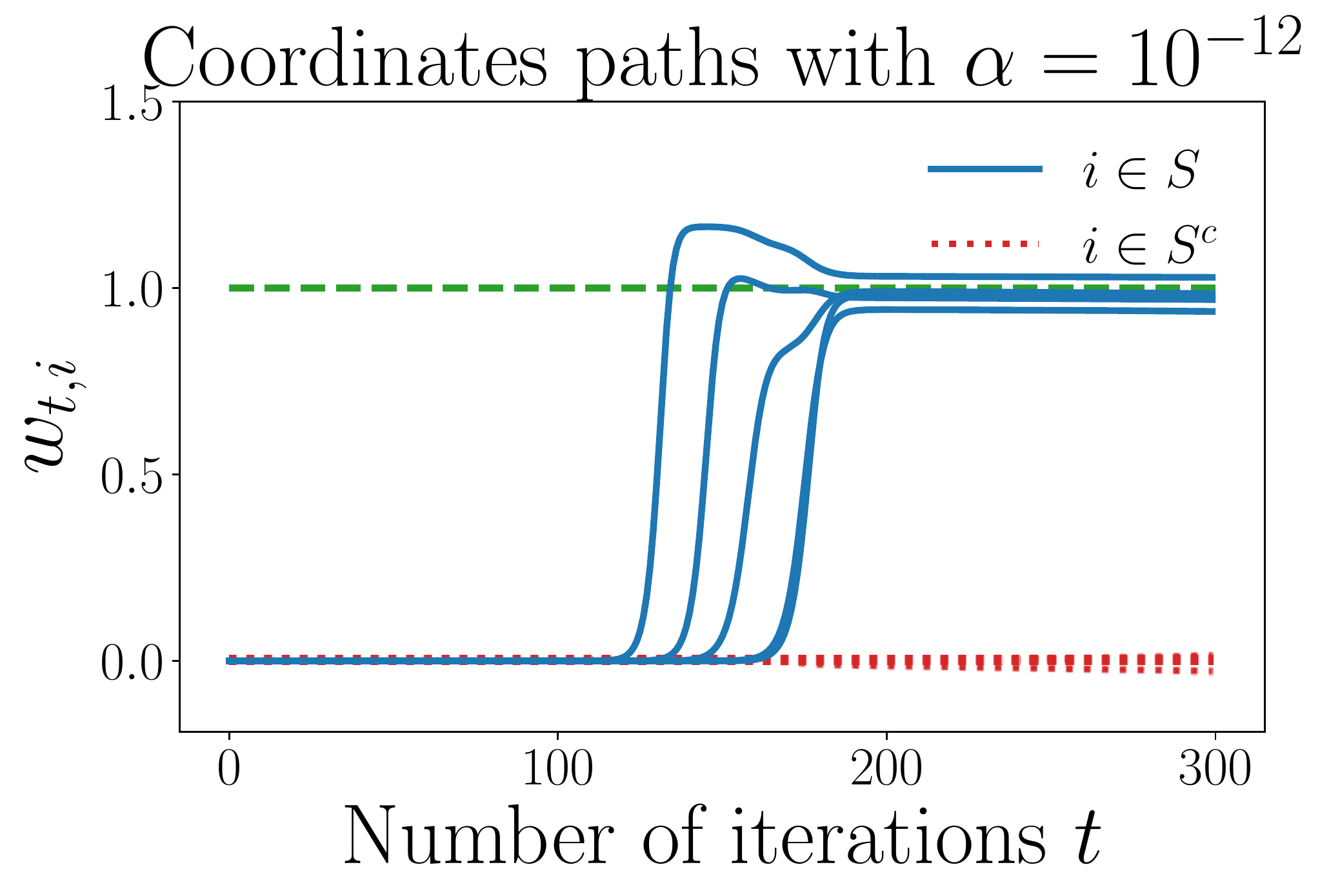}
  \end{subfigure}%
  \caption{Effects of initialization size. We set $k = 5, n = 100, \eta = 0.05,
    \sigma = 0.5$
    and run Algorithm~\ref{alg:gd}.
    We remark that the $X$ axes in the two figures on the right
    differ due to different choices of $\alpha$.}
  \label{fig:effects-of-initialization-size}
\end{figure}

\paragraph{Effects of Initialization Size.}
As discussed in Section~\ref{section:proof-sketch} each coordinate grows
exponentially at a different rate.
In Figure~\ref{fig:effects-of-initialization-size} we illustrate
the necessity of small initialization for bringing out the exponential
nature of coordinate paths allowing to effectively fit them one at a time.
For more intuition, suppose that coordinates outside the true support grow
at most as fast as $(1 + \varepsilon)^{t}$ while the coordinates on the true support
grow at least as fast as $(1 + 2\varepsilon)^{t}$. Since exponential function
is very sensitive to its base, for large enough $t$ we have
$(1 + \varepsilon)^{t} \ll (1 + 2\varepsilon)^{t}$.
The role of the initialization size $\alpha$ is then finding a small enough
$\alpha$ such that for large enough $t$ we have
$\alpha^{2}(1 + \varepsilon)^{t} \approx 0$ while $\alpha^{2}(1 + 2\varepsilon)^{t}$
is large enough to ensure convergence of the coordinates on the true support.

\begin{figure}[h]
  \centering
  \begin{subfigure}{.33\textwidth}
    \centering
    \includegraphics[width=\linewidth]{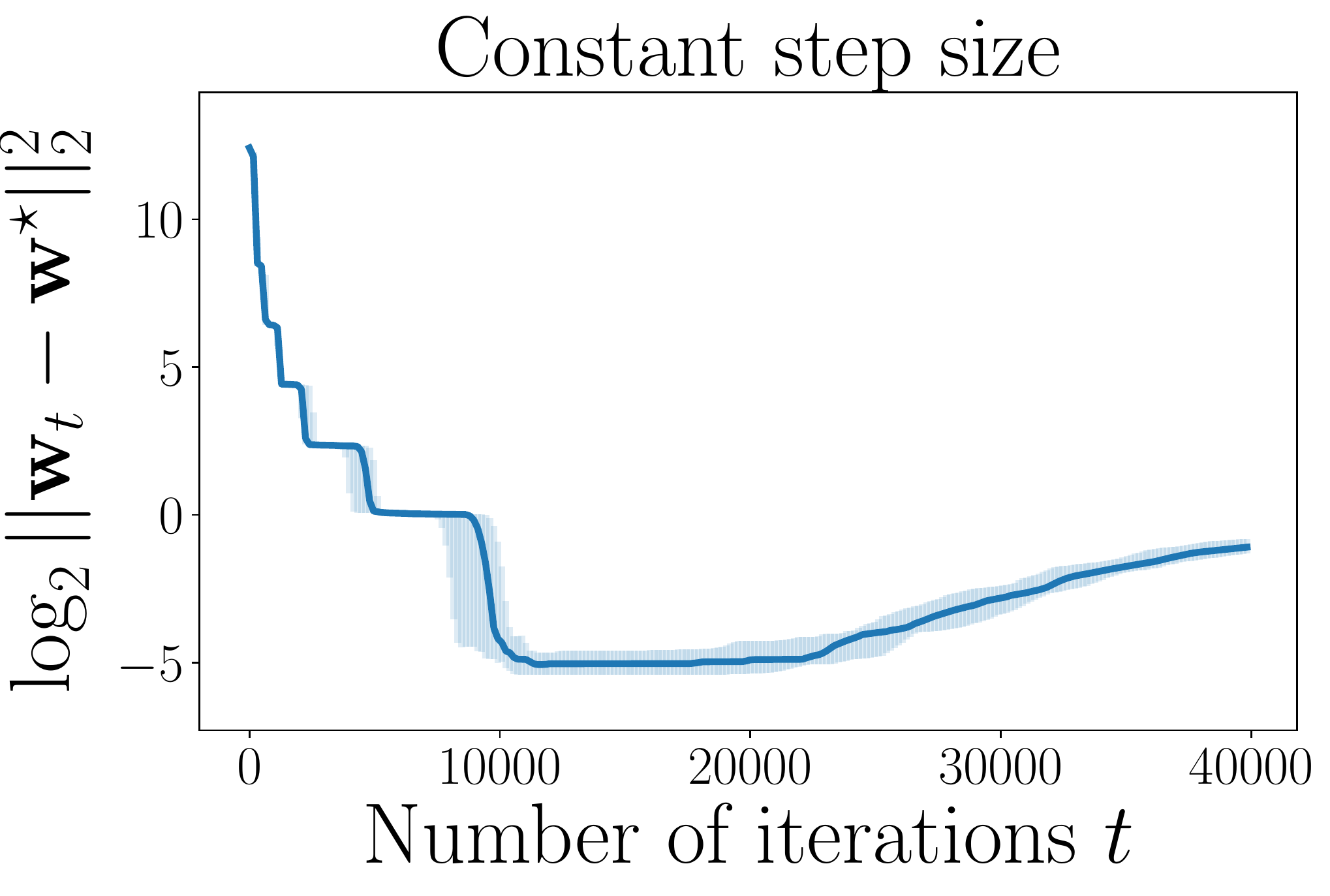}
  \end{subfigure}%
  \begin{subfigure}{.33\textwidth}
    \centering
    \includegraphics[width=\linewidth]{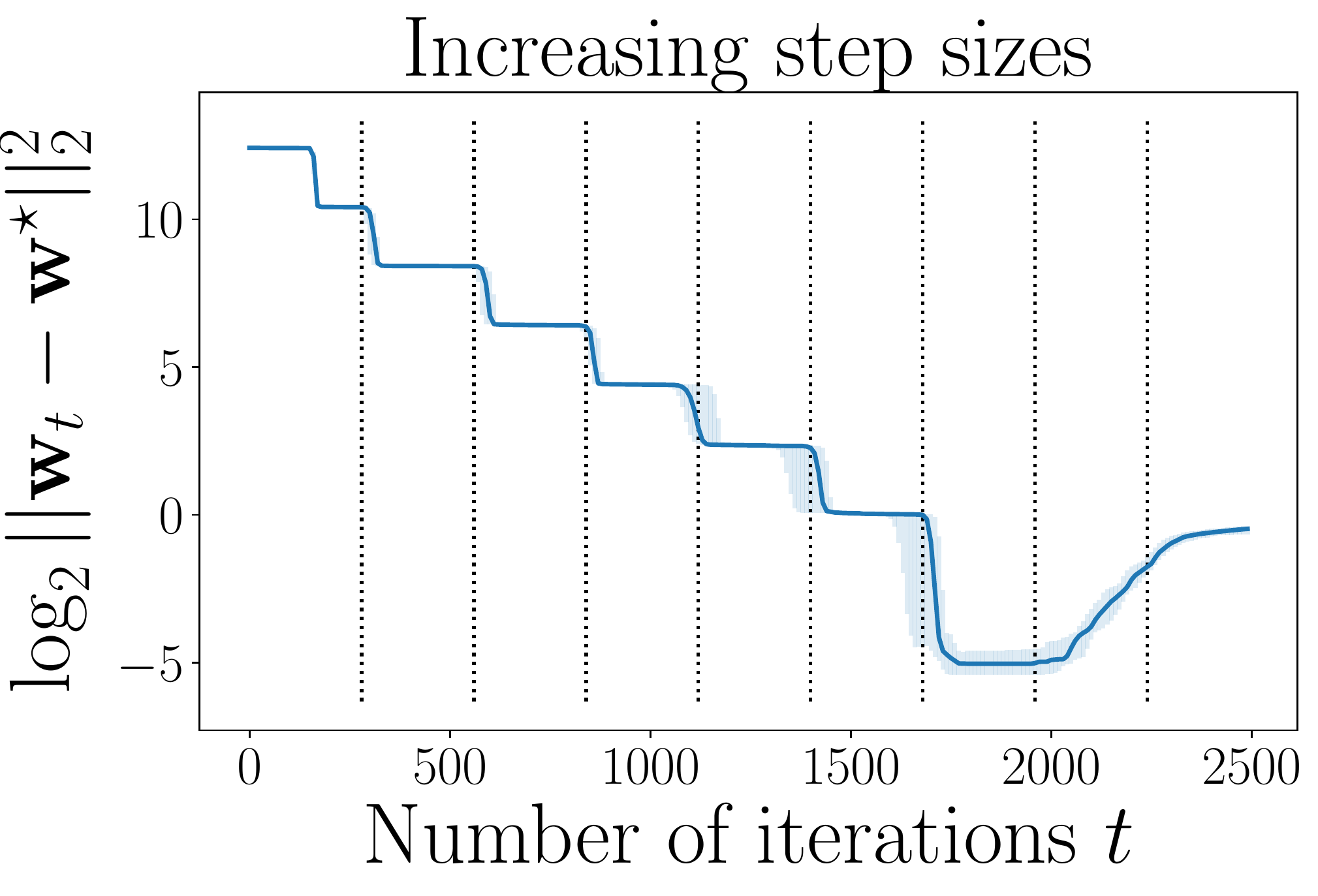}
  \end{subfigure}%
  \caption{Comparison of Algorithms~\ref{alg:gd} and \ref{alg:gd-increasing-steps}.
    Let $n = 250, k = 7$ and the non-zero coordinates of $\wstar$
    be $\{2^{i} : i = 0, \dots, 6\}$. For both algorithms let
    $\eta = 1/(20\cdot2^6)$.
    The vertical lines in the figure on the right
    are equally spaced at $\tau \log \alpha^{-1}$ iterations.
    The scale of x axes differ by a factor of $16$.
    The shaded region corresponds to $25\th$ and $75\th$ percentiles over $30$ runs.}
  \label{fig:step-size-schemes-comparison}
\end{figure}

\paragraph{Exponential Convergence with Increasing Step Sizes.}
We illustrate the effects of Algorithm~\ref{alg:gd-increasing-steps}
on an ill-conditioned target with $\kappa = 64$.
Algorithm~\ref{alg:gd} spends approximately twice the time to fit each
coordinate that the previous one, which is expected, since the coordinate sizes
decrease by half. On the other hand,
as soon as we increase the corresponding step size,
Algorithm~\ref{alg:gd-increasing-steps} fits each coordinate at approximately
the same number of iterations, resulting in $O(\log \kappa \log \alpha^{-1})$
total iterations. Figure~\ref{fig:step-size-schemes-comparison} confirms this behavior in simulations.

\begin{figure}[h]
  \begin{subfigure}{.33\textwidth}
    \centering
    \includegraphics[width=\linewidth]{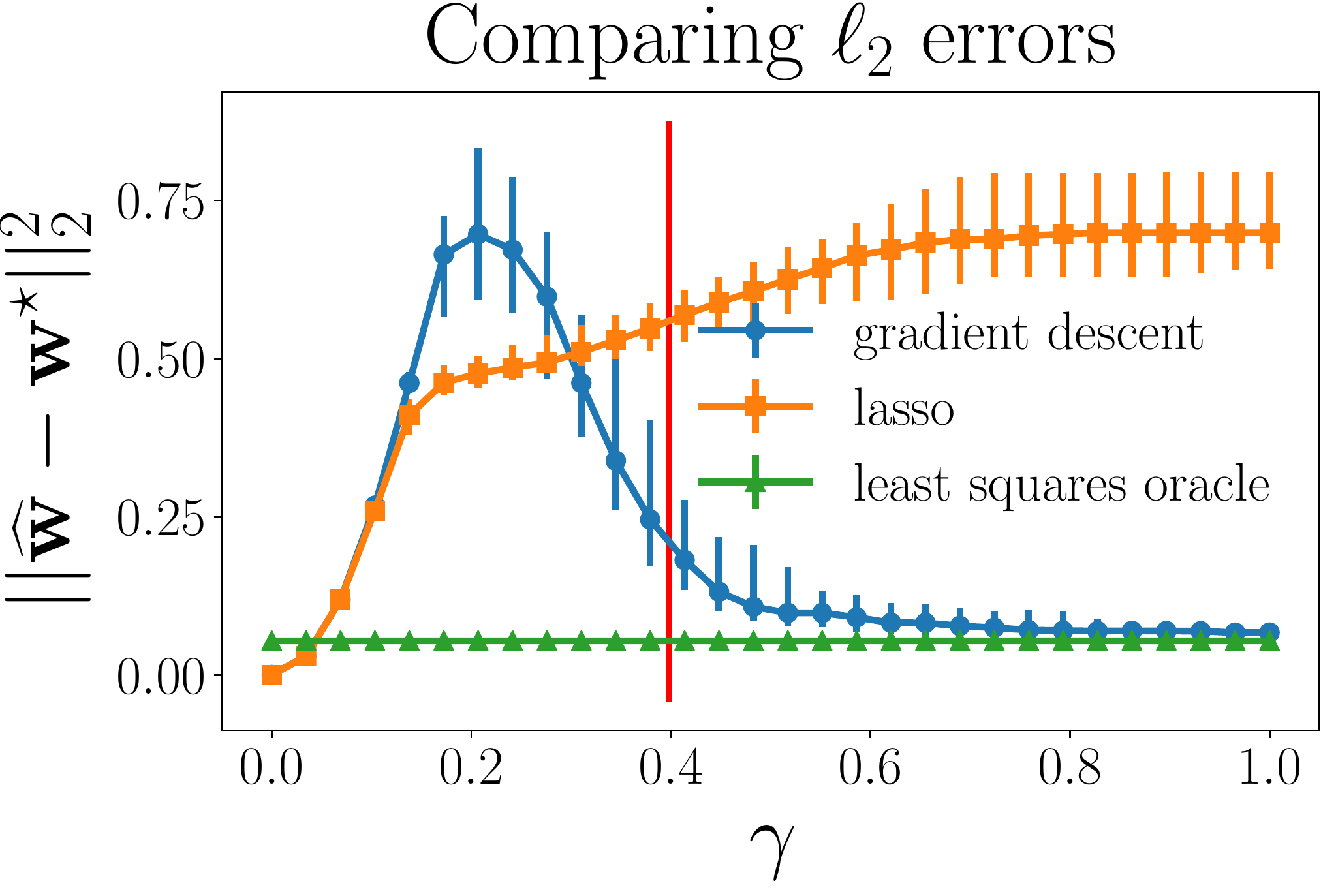}
    \label{fig:phase_transition_gamma}
  \end{subfigure}%
  \begin{subfigure}{.33\textwidth}
    \centering
    \includegraphics[width=\linewidth]{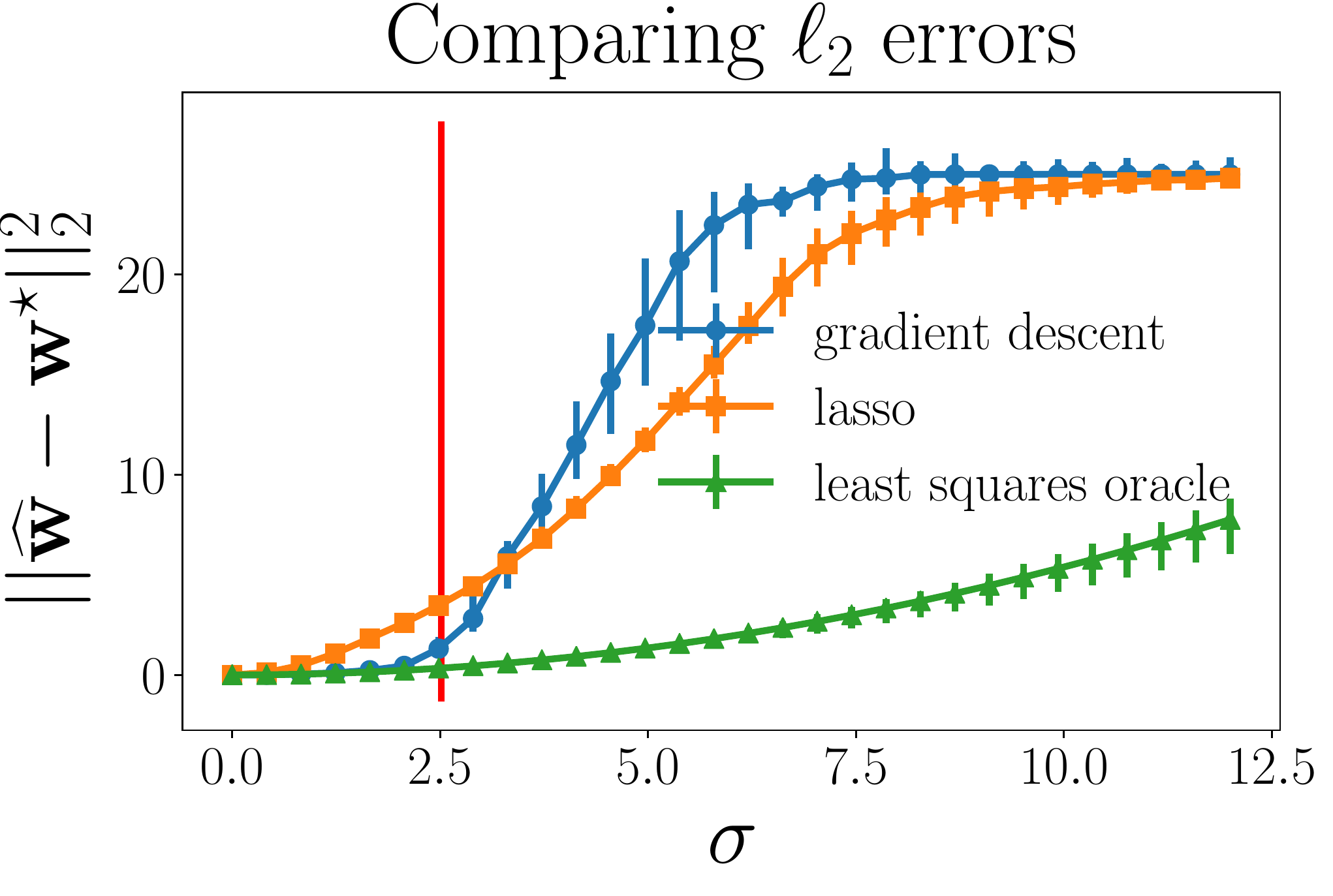}
    \label{fig:phase_transition_sigma}
  \end{subfigure}%
  \begin{subfigure}{.33\textwidth}
    \centering
    \includegraphics[width=\linewidth]{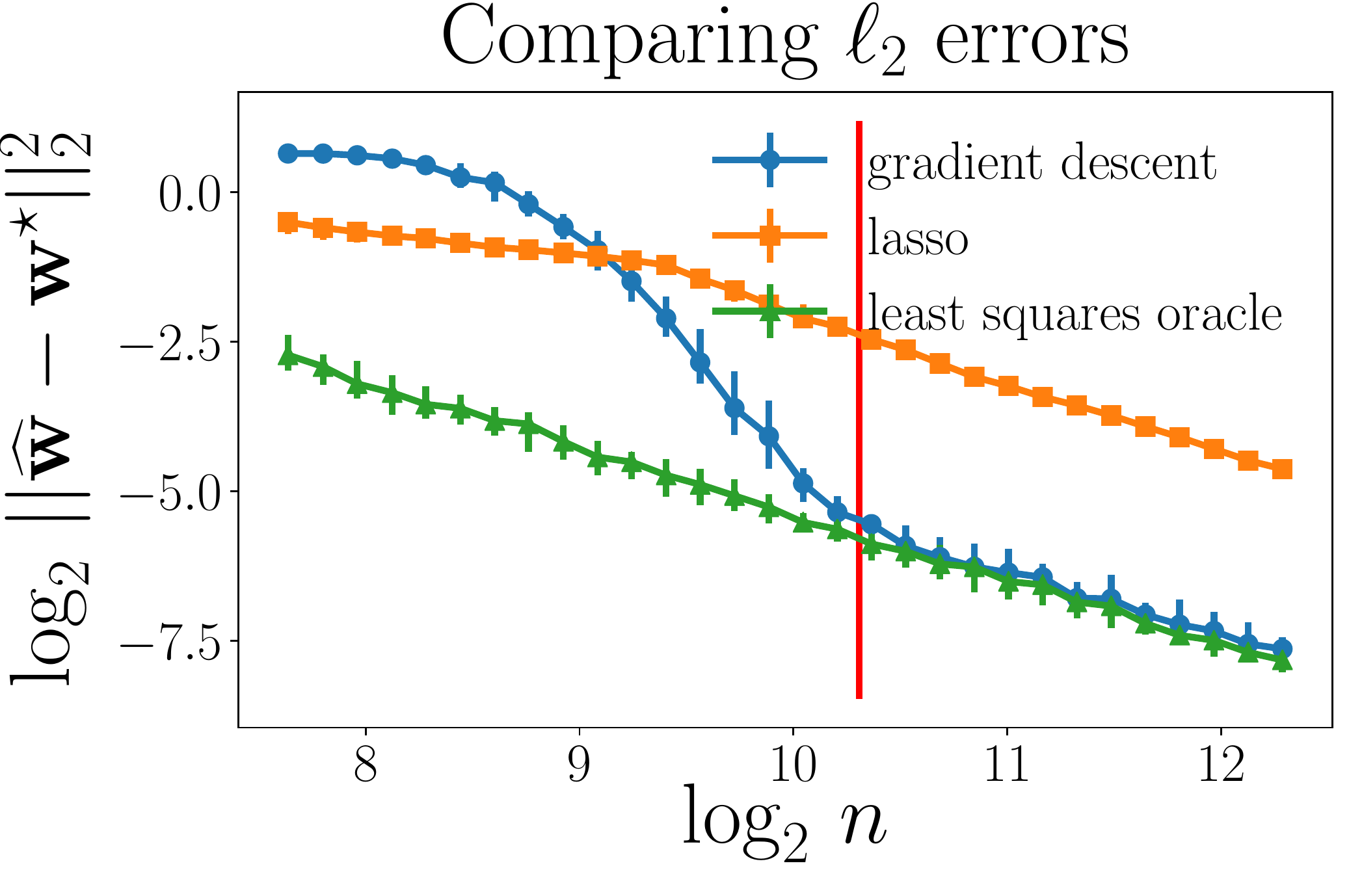}
    \label{fig:phase_transition_n}
  \end{subfigure}%
  \caption{Phase transitions. The figure on the right uses $\gamma = 1/4$.
  The red vertical lines show solutions of the equation
  $\wmin = \gamma = 2 \cdot \ex{\inlinemaxnoise}
   \leq 2 \cdot \sigma \sqrt{2 \log (2d)} / \sqrt{n}.$ }
  \label{fig:phase_transitions}
\end{figure}

\paragraph{Phase Transitions.}
As suggested by our main results, we present empirical evidence that
when $\wmin \gtrsim \inlinemaxnoise$ our algorithms undergo a phase transition
with dimension-independent error bounds.
We plot results for three different estimators.
First we run
Algorithm~\ref{alg:gd-increasing-steps} for $2000$ iterations and save
every $10\th$ model. Among the $200$ obtained models we choose the one with
the smallest error on a \emph{validation} dataset of size $n/4$.
We run the lasso for $200$ choices of $\lambda$ equally spaced on a
logarithmic scale and for each run we select a model with the
smallest $\ell_{2}$ parameter estimation error using an \emph{oracle knowledge}
of $\wstar$.
Finally, we perform a least squares fit using an \emph{oracle knowledge} of the
true support $S$.
Figure~\ref{fig:phase_transitions}
illustrates, that with varying $\gamma, \sigma$ and $n$
we can satisfy the condition $\wmin \gtrsim \inlinemaxnoise$ at which point our method
approaches an oracle-like performance.
Given exponential nature of the coordinate-wise convergence, all coordinates
of the true support grow at a strictly larger exponential rate than all of
the coordinates on $S^{c}$ as soon as $\wmin - \inlinemaxnoise >
\inlinemaxnoise$.
An approximate solution of this equation is shown in Figure~\ref{fig:phase_transitions}
using vertical red lines.

\begin{figure}[h]
  \centering
  \begin{subfigure}{.33\textwidth}
    \centering
    \includegraphics[width=\linewidth]{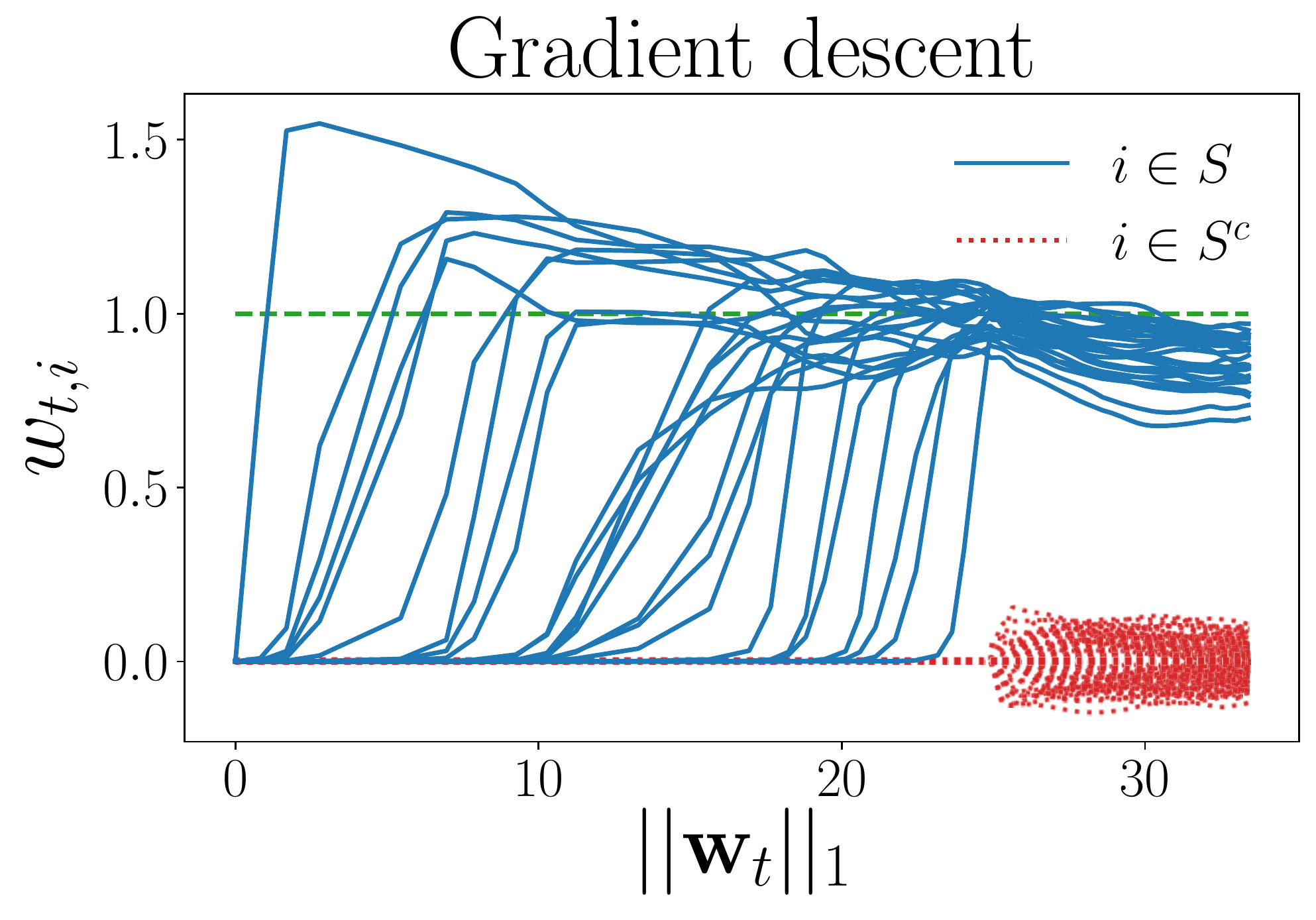}
  \end{subfigure}%
  \begin{subfigure}{.33\textwidth}
    \centering
    \includegraphics[width=\linewidth]{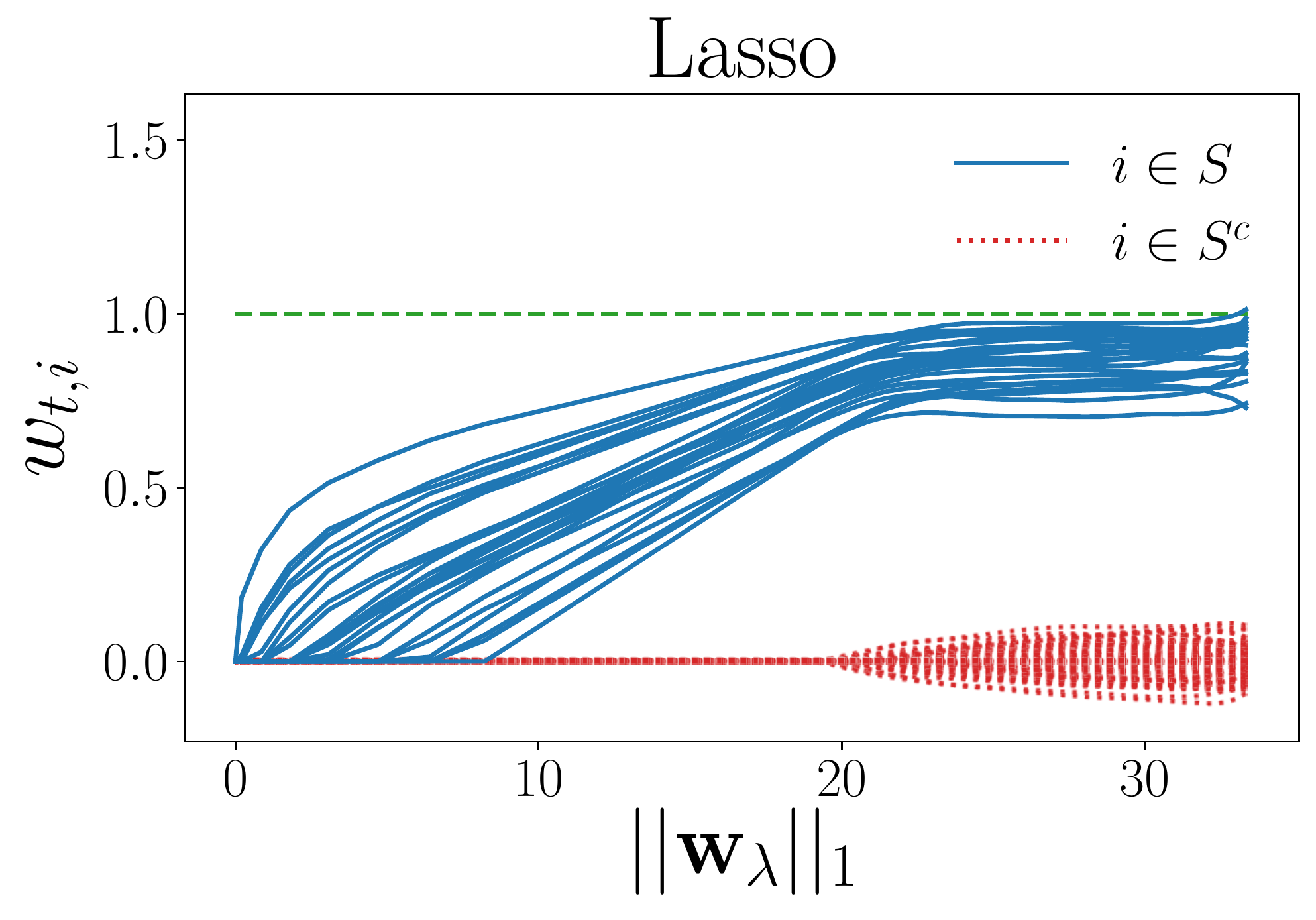}
  \end{subfigure}%
  \begin{subfigure}{.33\textwidth}
    \centering
    \includegraphics[width=\linewidth]{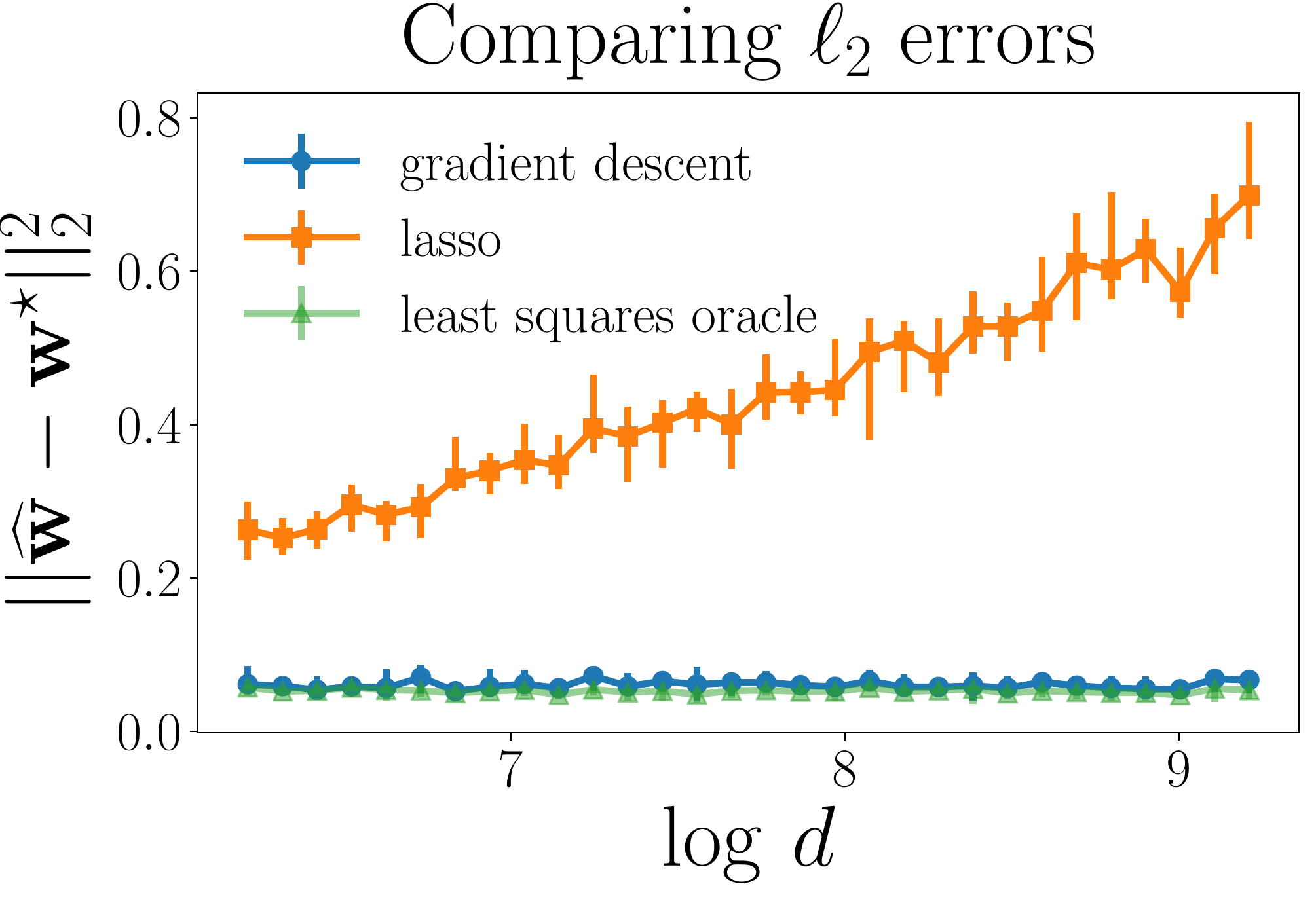}
  \end{subfigure}%
  \caption{Dimension dependent bias for the lasso. In contrast, in a
    high signal-to-noise ratio setting gradient descent is able to recover
    coordinates on  $S$ without a visible bias.}
  \label{fig:comparison_with_lasso}
\end{figure}

\paragraph{Dimension Free Bounds in the Easy Setting.}
Figure~\ref{fig:comparison_with_lasso} shows that
when $\wmin \gtrsim \inlinemaxnoise$
our algorithm matches the performance of oracle least squares which is
independent of $d$. In contrast, the performance of the lasso deteriorates as $d$
increases. To see why this is the case, in the setting where
$\XtX/n = \matrixid$, the lasso solution with parameter $\lambda$ has a closed
form solution $w^\lambda_i = \sign(w^\LS_i)(|w^\LS_i| - \lambda)_+$, where
$\vw^\LS$ is the least squares solution.
In the sub-Gaussian noise setting, the minimax rates are achieved by the choice
$\lambda = \Theta(\sqrt{\sigma^2 \log(d)/n})$ introducing
a bias which depends on $\log d$.
Such a bias is illustrated in Figure~\ref{fig:comparison_with_lasso} and is
not present at the optimal stopping time of our algorithm.
\newline 

%% file: files/further_improvements.tex
\section{Further Improvements}
\label{section:further-improvements}

While we show that Algorithms~\ref{alg:gd} and \ref{alg:gd-increasing-steps}
yield optimal statistical rates and in addition
Algorithm~\ref{alg:gd-increasing-steps} is optimal in terms of computational
requirements, our results can be improved in two different aspects.
First, our constraints on the RIP parameter $\delta$ result in sub-optimal
sample complexity.
Second, the RIP condition could potentially be replaced by the
restricted eigenvalue (RE) condition which allows correlated designs.
We expand on both of the points below and provide empirical evidence
suggesting that both inefficiencies are artifacts of our analysis and
not inherent limitations of our algorithms.

\textbf{Sub-Optimal Sample Complexity.}
Our RIP parameter $\delta$ scales as $\widetilde{O}(1/\sqrt{k})$.
We remark that such scaling on $\delta$ is less restrictive than
in \cite{li2018algorithmic, zhao2019implicit} (see
Appendix~\ref{appendix:comparing-with-colt-paper} and
\ref{appendix:comparing-with-hadamard-product-paper}).
If we consider, for
example, sub-Gaussian isotropic designs, then satisfying such an assumption
requires $n \gtrsim k^{2} \log(ed/k)$ samples.
To see that, consider an $n \times k$ i.i.d.\ standard normal ensemble which
we denote by $\X$.
By standard results in random-matrix theory
\cite[Chapter 6]{wainwright2019high},
$\inlinenorm{\XtX/n - \matrixid} \lesssim \sqrt{k/n} + k/n$ where
$\norm{\cdot}$ denotes the operator norm.
Hence, we need $n \gtrsim k^{2}$ to satisfy
$\inlinenorm{\XtX/n - \matrixid} \lesssim 1/\sqrt{k}$.

\begin{figure}[h]
  \centering
  \begin{minipage}[c]{0.54\textwidth}
    \begin{subfigure}{.5\textwidth}
      \centering
      \includegraphics[width=\linewidth]{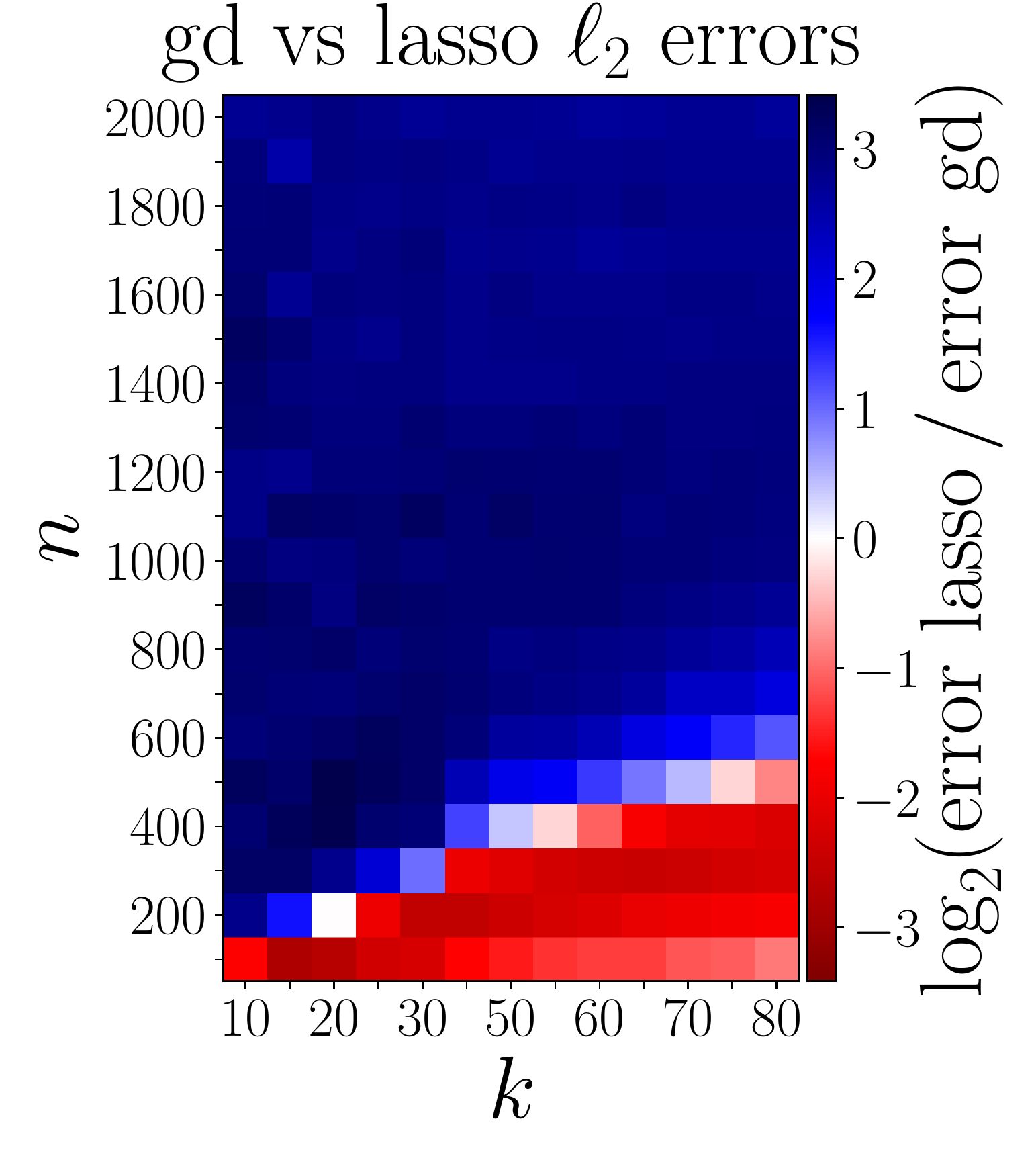}%
    \end{subfigure}%
    \begin{subfigure}{.5\textwidth}
      \centering
      \includegraphics[width=\linewidth]{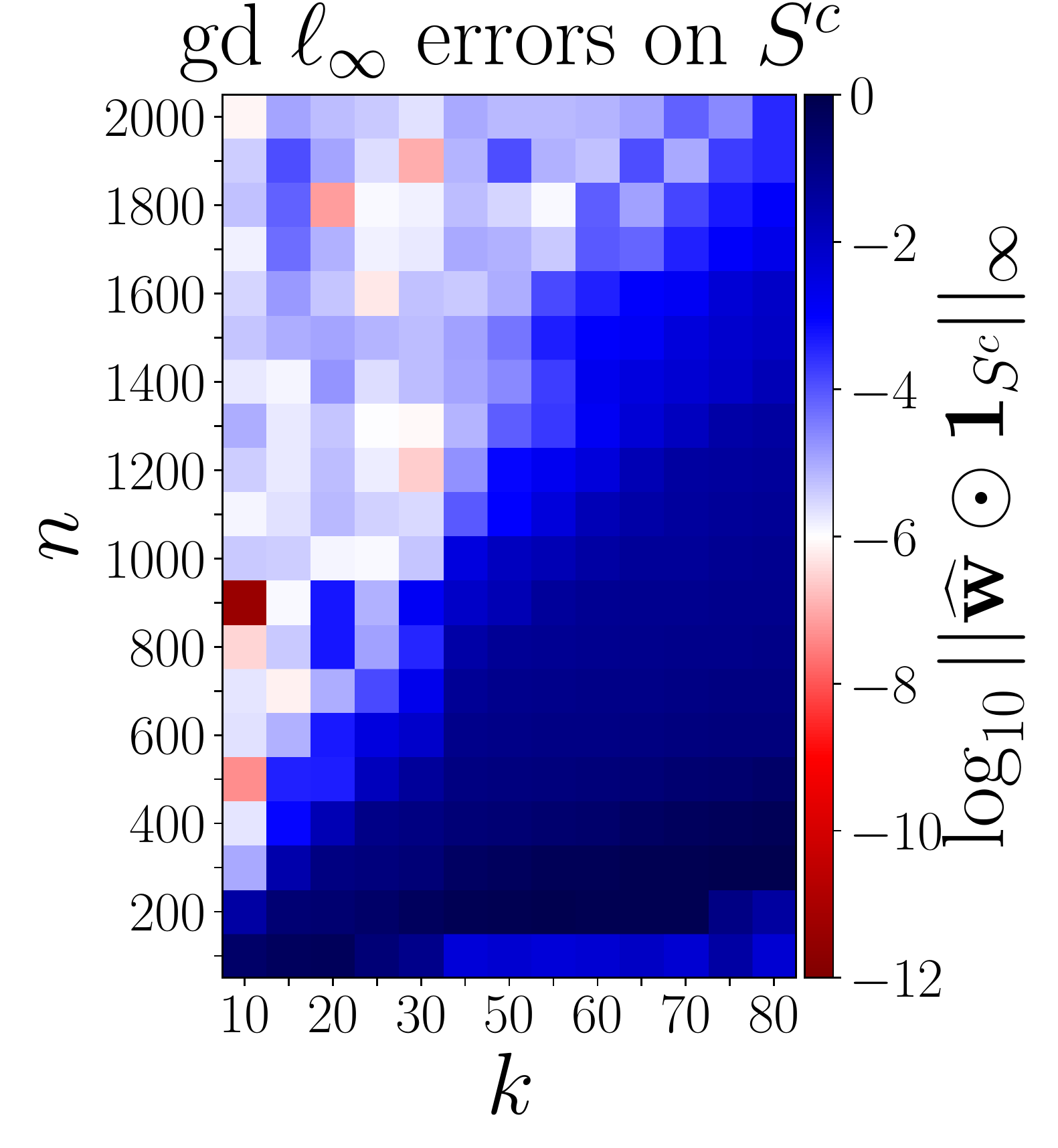}%
    \end{subfigure}%
  \end{minipage}
  \begin{minipage}[c]{0.45\textwidth}
    \caption{Sample complexity requirements. We let $d = 5000, \sigma = 1$
     and $\wstar_{S} = \id{S}$. The plot on the left computes the $\log_{2}$
     error ratio for our method (stopping time chosen by cross-validation) and
     the lasso ($\lambda$ chosen optimally using knowledge of $\wstar$).
     The plot on the right computes
     $\inlinenorm{\vec{w}_{t} \odot \id{S^{c}}}_{\infty}$
     for optimally chosen $t$.
    }
    \label{fig:sample-complexity}
  \end{minipage}
\end{figure}

Note that
Theorems~\ref{thm:constant-step-sizes} and
\ref{thm:main-theorem-noisy-minimax-rates-exponential-convergence}
provide coordinate-wise bounds which is in general harder than providing $\ell_{2}$ error bounds directly.
In particular, under the condition that $\delta = \widetilde{O}(1/\sqrt{k})$, our main theorems imply minimax-optimal $\ell_{2}$ bounds; this requirement on $\delta$ implies that $n$ needs to be at least quadratic in $k$. 
Hence we need to answer two questions. First, do we need sample complexity
quadratic in $k$ to obtain minimax-rates?
The left plot in Figure~\ref{fig:sample-complexity} suggests that
linear sample complexity in $k$ is enough for our method
to match and eventually exceed performance of the lasso in terms of
$\ell_{2}$ error.
Second, is it necessary to change our $\ell_{\infty}$ based analysis to an
$\ell_{2}$ based analysis in order to obtain optimal sample complexity?
The right plot in Figure~\ref{fig:sample-complexity} once again suggests that
sample complexity linear in $k$ is enough for our main theorems to hold.

\textbf{Relaxation to the Restricted Eigenvalue (RE) Assumption.}
The RIP assumption is crucial for our analysis. However, the lasso satisfies
minimax optimal rates under less restrictive assumptions, namely, the
RE assumption introduced in \cite{bickel2009simultaneous}.
The RE assumption with parameter $\gamma$ requires
that $\inlinenorm{\X\vec{w}}_{2}^{2}/n \geq \gamma\inlinenorm{\vec{w}}_{2}^{2}$
for vectors $\vec{w}$ satisfying the cone condition
$\inlinenorm{\vec{w}_{S^{c}}}_{1} \leq c \inlinenorm{\vec{w}_{S}}_{1}$
for a suitable choice of constant $c \geq 1$.
In contrast to RIP, RE only imposes constraints on the \emph{lower}
eigenvalue of $\XtX/n$ for approximately sparse vectors and can be satisfied
by random \emph{correlated} designs \cite{raskutti2010restricted,
rudelson2012reconstruction}.
The RE condition was shown to be necessary for any polynomial-time
algorithm returning a sparse vector and achieving fast rates for prediction error
\cite{zhang2014lower}.

\begin{figure}[h]
  \centering
  \begin{subfigure}{.33\textwidth}
    \centering
    \includegraphics[width=\linewidth]{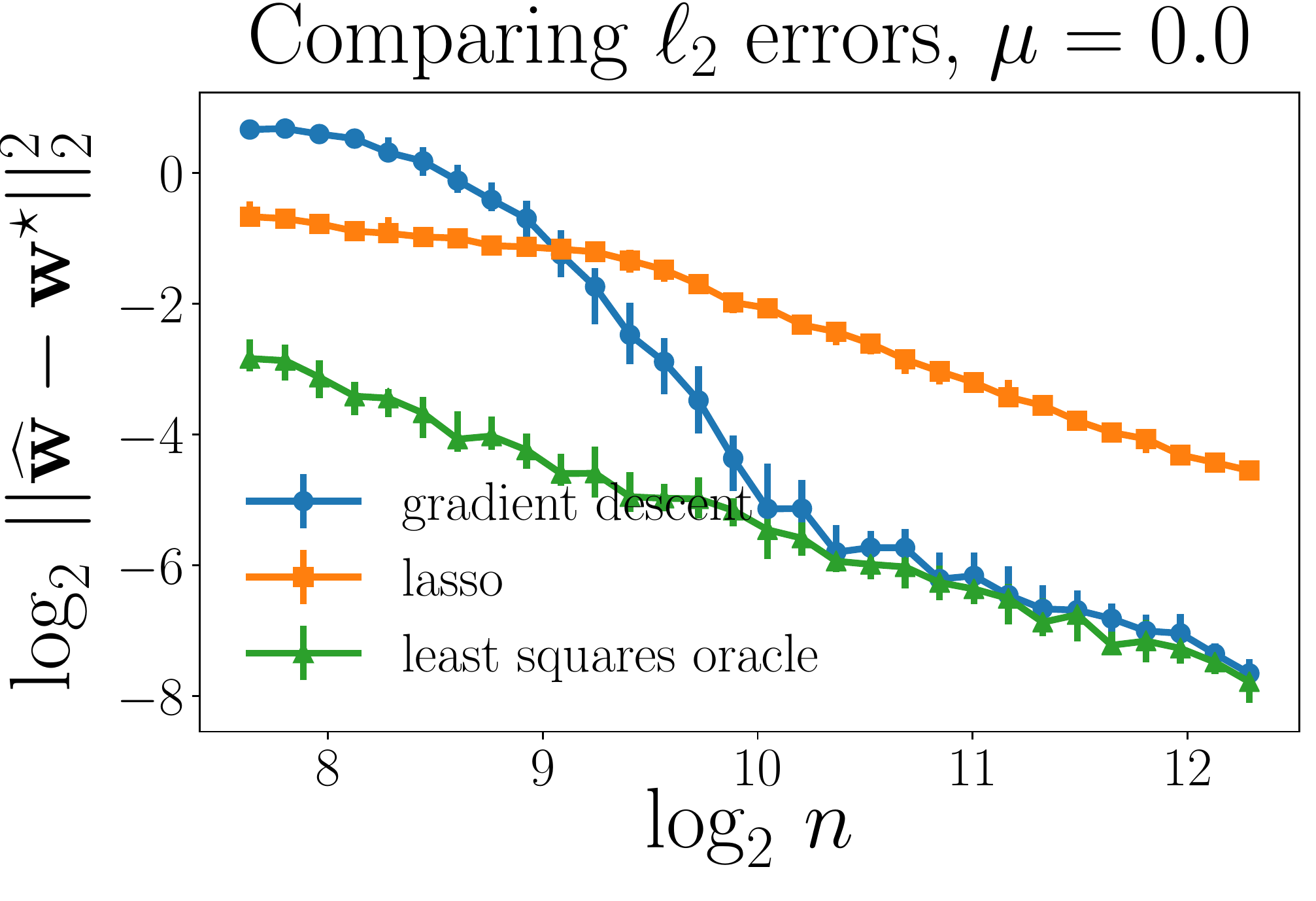}
  \end{subfigure}%
  \begin{subfigure}{.33\textwidth}
    \centering
    \includegraphics[width=\linewidth]{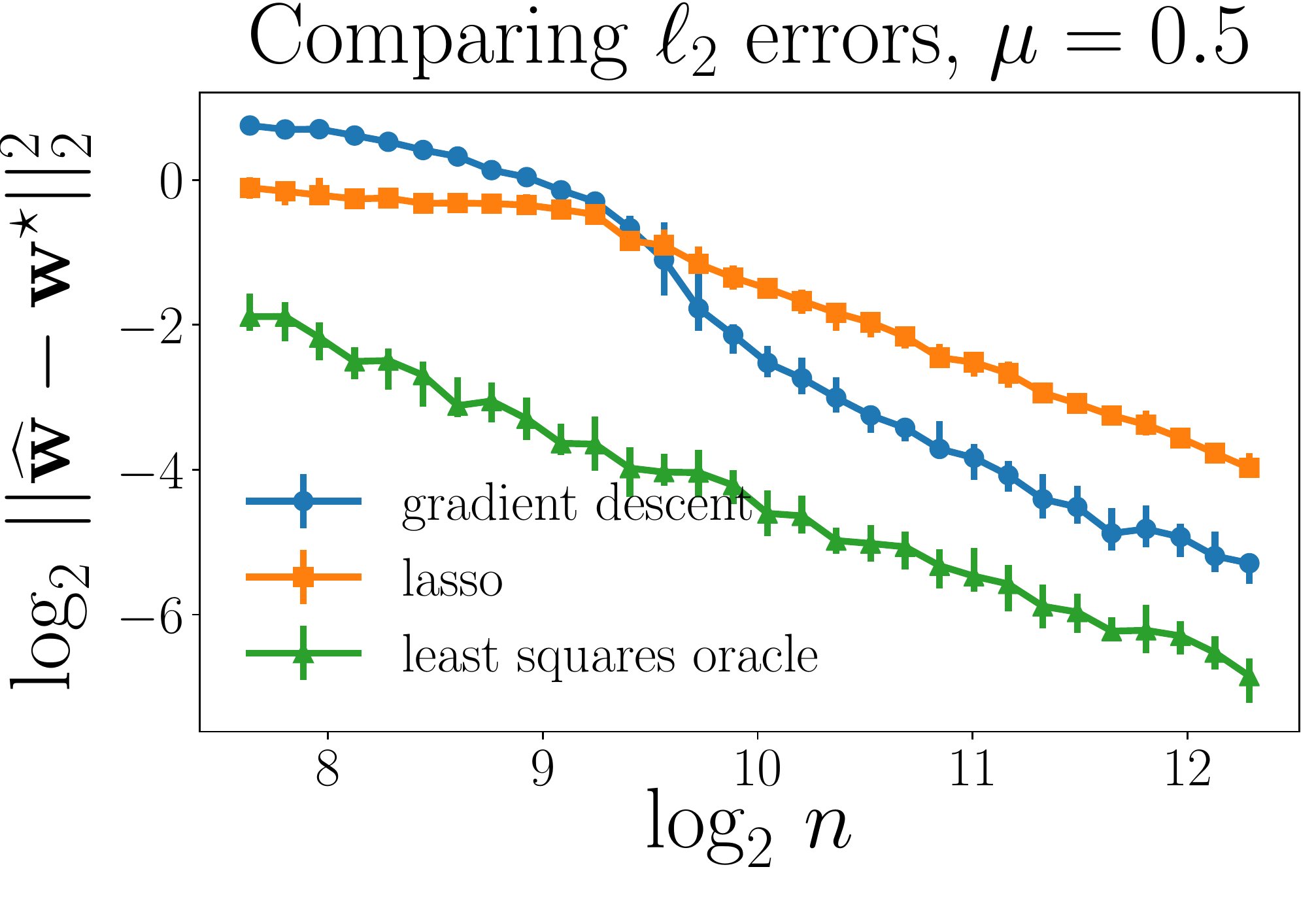}
  \end{subfigure}%
  \caption{Violating the RIP assumption. We consider the same setting as
    in Figure~\ref{fig:phase_transitions} with rows of $\X$ sampled from
    a Gaussian distribution with covariance matrix equal to
    $(1 - \mu)\matrixid + \mu \id{}\id{}^{\mathsf{T}}$.
  }
  \label{fig:violating-rip}
\end{figure}

We sample i.i.d.\ Gaussian ensembles with
covariance matrices equal to $(1 - \mu)\matrixid + \mu \id{}\id{}^{\mathsf{T}}$
for $\mu = 0$ and $0.5$.
For $\mu = 0.5$ the RIP fails but the RE property holds with high probability
\cite[Chapter 7]{wainwright2019high}.
In Figure~\ref{fig:violating-rip} we show empirically that our method achieves the
fast rates and eventually outperforms the lasso even when we violate the
RIP assumption.

%% file: files/appendix.tex
\appendix
\section*{Appendix}
\label{section:appendix}

\setcounter{lemma}{0}
\renewcommand{\thelemma}{\Alph{section}.\arabic{lemma}}

The appendix is organized as follows.

In Appendix~\ref{appendix:main-proofs} we introduce the key ideas and intuition
behind the proof of
Theorem~\ref{thm:constant-step-sizes}.

In Appendix~\ref{appendix:multiplicative-update-sequences} we go deeper into
technical details and prove the main propositions used to prove
Theorem~\ref{thm:constant-step-sizes}.

In Appendix~\ref{appendix:missing-proofs} we prove the lemmas stated in
appendix~\ref{appendix:main-proofs}.

In Appendix~\ref{appendix:selecting-the-step-size} we prove
Theorem~\ref{thm:selecting-the-step-size}.

In Appendix~\ref{appendix:theorem-minimax-rates-exponential-convergence-proof}
we prove
Theorem~\ref{thm:main-theorem-noisy-minimax-rates-exponential-convergence}.

In Appendix~\ref{appendix:gradient-descent-updates} we derive the gradient
descent updates used by our parametrization.

In Appendix~\ref{appendix:comparing-with-colt-paper} we compare our assumptions
with the ones made in \cite{li2018algorithmic}.

In Appendix~\ref{appendix:comparing-with-hadamard-product-paper} we compare our
main result with a recent arXiv preprint \cite{zhao2019implicit}, where
Hadamard product reparametrization was used to induce sparsity implicitly.

In Appendix~\ref{appendix:table-of-notation} we provide a table of notation.

\subfile{appendix/main_proofs.tex}

\subfile{appendix/multiplicative_updates.tex}

\subfile{appendix/missing_proofs.tex}

\subfile{appendix/eta_initialization.tex}

\subfile{appendix/increasing_step_sizes.tex}

\subfile{appendix/gradient_descent_updates.tex}

\subfile{appendix/comparing_with_colt_assumptions.tex}

\subfile{appendix/comparing_with_the_hadamard_product_paper.tex}

\subfile{appendix/table_of_notation.tex}

%% file: appendix/main_proofs.tex
\section{Proof of Theorem~\ref{thm:constant-step-sizes}}
\label{appendix:main-proofs}

This section is dedicated to providing a high level proof for Theorem
\ref{thm:constant-step-sizes}.
In Section~\ref{appendix:main-proofs:set-up-and-intuition}
we set up the notation and explain how we decompose our iterates into
signal and error sequences.
In Section~\ref{appendix:main-proofs:key-propositions} we state and discuss the
implications of the two key propositions allowing to prove
our theorem.
In Section~\ref{appendix:main-proofs:technical-lemmas} we state some technical
lemmas used in the proofs of the main theorem and its corollaries.
In Section~\ref{appendix:main-proofs:proof-of-thm-constant-step-size}
we prove Theorem~\ref{thm:constant-step-sizes}.
Finally in Section~\ref{appendix:main-proofs:proofs-of-corollaries} we prove the
corollaries.

\subfile{appendix/main_proofs/set_up.tex}

\subfile{appendix/main_proofs/key_propositions.tex}

\subfile{appendix/main_proofs/technical_lemmas.tex}

\subfile{appendix/main_proofs/thm_constant_step_size.tex}

\subfile{appendix/main_proofs/corollaries.tex}

%% file: appendix/main_proofs/set_up.tex
\subsection{Set Up and Intuition}
\label{appendix:main-proofs:set-up-and-intuition}

Let
$\vec{w}_{t}
\coloneqq
\vec{w}_{t}^{+} - \vec{w}_{t}^{-}$
where $\vec{w}_{t}^{+} \coloneqq \vec{u}_{t}
\odot \vec{u}_{t}$ and
$\vec{w}_{t}^{-} \coloneqq \vec{v}_{t} \odot \vec{v}_{t}$.
The gradient descent
updates on $\vec{u}_{t}$ and $\vec{v}_{t}$ read as
(see Appendix~\ref{appendix:gradient-descent-updates} for derivation)
\begin{align*}
  \vec{u}_{t+1}
  &= \vec{u}_{t} \odot \left(
       \id{}
       - 4\eta \left(
           \frac{1}{n}\Xt(\X(\vec{w}_{t} - \wstar) - \vec{\xi})
         \right)
       \right), \\
  \vec{v}_{t+1}
  &= \vec{v}_{t} \odot \left(
       \id{}
       + 4\eta \left(
           \frac{1}{n}\Xt(\X(\vec{w}_{t} - \wstar) - \vec{\xi})
         \right)
       \right).
\end{align*}

Let $S^{+}$ denote the coordinates of $\wstar$ such that $w^{\star}_{i} > 0$
and let $S^{-}$ denote the coordinates of $\wstar$ such that
$w^{\star}_{i} < 0$.
So $S = S^{+} \cup S^{-}$ and $S^{+} \cap S^{-} = \emptyset$.
Then define the following sequences
\begin{align}
  \begin{split}
    \label{eq:error-decompositions}
    \vec{s}_{t} &\coloneqq
      \id{S^{+}} \odot \vec{w}_{t}^{+}
      - \id{S^{-}} \odot \vec{w}_{t}^{-}, \\
    \vec{e}_{t} &\coloneqq
      \id{S^{c}} \odot \vec{w}_{t}
      + \id{S^{-}} \odot \vec{w}_{t}^{+}
      - \id{S^{+}} \odot \vec{w}_{t}^{-}, \\
    \vec{b}_{t} &\coloneqq
          \frac{1}{n}\XtX \vec{e}_{t}
          - \frac{1}{n}\Xt\vec{\xi}, \\
    \vec{p}_{t} &\coloneqq \left( \frac{1}{n}\XtX - \matrixid \right)
          \left(\vec{s}_{t} - \wstar \right).
  \end{split}
\end{align}

Having defined the sequences above we can now let $\alpha^{2}$ be the
initialization size and rewrite the updates on
$\vec{w}_{t}$, $\vec{w}_{t}^{+}$ and $\vec{w}_{t}^{-}$
in a more succinct way
\begin{align}
  \begin{split}
    \label{eq:updates-equation-using-b-p-notation}
    \vec{w}_{0}^{+} &= \vec{w}_{0}^{-} = \alpha^{2}, \\
    \vec{w}_{t}
    &= \vec{w}_{t}^{+} - \vec{w}_{t}^{-} \\
    \vec{w}_{t+1}^{+}
    &= \vec{w}_{t}^{+} \odot \left(
         \id{}
         - 4\eta \left(
             \vec{s}_{t} - \wstar + \vec{p}_{t} + \vec{b}_{t}
           \right)
         \right)^{2}, \\
    \vec{w}_{t+1}^{-}
    &= \vec{w}_{t}^{-} \odot \left(
         \id{}
         + 4\eta \left(
             \vec{s}_{t} - \wstar + \vec{p}_{t} + \vec{b}_{t}
           \right)
         \right)^{2}.
  \end{split}
\end{align}

We will now explain the roles played by each sequence defined
in equation~\eqref{eq:error-decompositions}.
\begin{enumerate}
  \item The sequence $(\vec{s}_{t})_{t \geq 0}$ represents the signal that we have
    fit by iteration $t$. In the noiseless setting,
    $\vec{s}_{t}$ would converge to $\wstar$.
    We remark that $\vec{w}_{t}^{+}$ is responsible for fitting the positive
    components of $\wstar$ while $\vec{w}_{t}^{-}$ is responsible for fitting the
    negative components of $\wstar$. If we had the knowledge of $S^{+}$ and
    $S^{-}$ before starting our algorithm, we would set $\vec{w}_{0}$ to
    $\vec{s}_{0}$.
  \item The sequence $(\vec{e}_{t})_{t \geq 0}$ represents the error sequence.
    It has three components:
    $\id{S^{c}} \odot \vec{w}_{t}$,
    $\id{S^{-}} \odot \vec{w}_{t}^{+}$ and
    $\id{S^{+}} \odot \vec{w}_{t}^{-}$
    which represent the errors of our estimator arising due to not having the
    knowledge of $S^{c}$, $S^{+}$ and $S^{-}$ respectively.
    For example, if we knew that $\wstar \succcurlyeq 0$
    we could instead use the parametrization
    $\vec{w}_{0} = \vec{u}_{0} \odot \vec{u}_{0} = \vec{w}_{0}^{+}$
    while if we knew that $\wstar \preccurlyeq 0$ then we would use the
    parametrization
    $\vec{w}_{0} = - \vec{v}_{0} \odot \vec{v}_{0} = -\vec{w}_{0}^{-}$.

    A key property of our main results is that we stop running gradient descent
    before $\norm{\vec{e}_{t}}_{\infty}$ exceeds some function of
    initialization size. This allows us to recover the coordinates
    from the true
    support $S$ that are sufficiently above the noise level
    while keeping the coordinates outside the true support
    arbitrarily close to $0$.
   \item We will think of the sequence $(\vec{b}_{t})_{t \geq 0}$ as a sequence
     of bounded perturbations to our gradient descent updates. These
     perturbations come from two different sources. The first one is the
     term $\noisevec$ which arises due to the noise on the
     labels.
     Hence this part of error is never greater than
     $\maxnoise$ and is hence bounded
     with high probability in the case of subGaussian noise.
     The second source of error is $\frac{1}{n} \XtX \vec{e}_{t}$ and it
     comes from the error sequence $(\vec{e}_{t})_{t \geq 0}$ being non-zero.
     Even though this term is in principle can be unbounded, as remarked in the
     second point above, we will always stop running gradient descent while
     $\norm{\vec{e}_{t}}_{\infty}$ remains close enough to $0$.
     In particular, this allows to treat
     $\frac{1}{n} \XtX \vec{e}_{t}$ as a bounded error term .

   \item We will refer to the final error sequence $(\vec{p}_{t})_{t \geq 0}$ as
     a sequence of errors proportional to convergence distance.
     An intuitive explanation of the restricted isometry property is that
     $\frac{1}{n} \XtX \approx \matrixid$ for sparse vectors.
     The extent to which this approximation is exact is controlled by the
     RIP parameter $\delta$. Hence the sequence $(\vec{p}_{t})_{t \geq 0}$ represents
     the error arising due to $\frac{1}{\sqrt{n}}\X$ not being an exact isometry
     for sparse vectors in a sense that $\delta \neq 0$.
     If we require that $\delta \leq \gamma / \sqrt{k}$ for some
     $\gamma > 0$ then as we shall see in
     section~\ref{appendix:main-proofs:technical-lemmas}
     we can upper bound $\norm{\vec{p}_{t}}_{\infty}$ as
     $$\norm{\vec{p}_{t}}_{\infty} \leq \delta \norm{\vec{s}_{t} - \wstar}_{2}
     \leq \gamma \norm{\vec{s}_{t} - \wstar}_{\infty}.$$
     Since this is the only worst-case control we have on $(\vec{p}_{t})_{t \geq 0}$
     one may immediately see the most challenging part of our analysis.
     For small $t$ we have $\vec{s}_{t} \approx
     0$ and hence in the worst case
     $\norm{\vec{p}_{t}}_{\infty} \approx \gamma \norm{\wstar}_{\infty}$.
     Since $\norm{\wstar}_{\infty}$ can be arbitrarily large, we can hence see
     that while $t$ is small it is possible for some elements of $(\vec{e}_{t})_{t
       \geq 0}$ to grow at a very fast rate, while some of the signal terms
     in the sequence $\vec{s}_{t}$ can actually shrink, for example, if
     $\gamma\norm{\wstar}_{\infty} > \inlineabs{w_{i}^{\star}}$
     for some $i \in S$.
     We address this difficulty in
     Section~\ref{appendix:multiplicative-updates:dealing-with-rip-errors}.
\end{enumerate}

One final thing to discuss regarding our iterates $\vec{w}_{t}$ is how to initialize
$\vec{w}_{0}$. Having the point two above in mind, we will always want
$\norm{\vec{e}_{t}}_{\infty}$ to be as small as possible. Hence we should initialize
the sequences $(\vec{u}_{t})_{t \geq 0}$ and $(\vec{v}_{t})_{t \geq 0}$ as close to
$0$ as possible. Note, however, that due to the
multiplicative nature of gradient descent updates using our parametrization,
we cannot set $\vec{u}_{0} = \vec{v}_{0} = 0$ since this is a saddle point
for our optimization objective function. We will hence
set $\vec{u}_{0} = \vec{v}_{0} = \alpha$ for some small enough
positive real number $\alpha$.

Appendix~\ref{appendix:multiplicative-update-sequences} is dedicated to
understanding the behavior of the updates given in
equation~\eqref{eq:updates-equation-using-b-p-notation}.
In appendix~\ref{appendix:multiplicative-updates:basic-lemmas} we analyze
behavior of $(\vec{w}_{t}^{+})_{t \geq 0}$ assuming that
$\vec{w}_{t}^{-} = 0$, $\vec{p}_{t} = 0$
and $\vec{b}_{t} = 0$. In appendix~\ref{appendix:multiplicative-updates:bounded-errors} we show
how to handle the bounded errors sequence $(\vec{b}_{t})_{t \geq 0}$ and in
appendix~\ref{appendix:multiplicative-updates:dealing-with-rip-errors} we show
how to deal with the errors proportional to convergence distance
$(\vec{p}_{t})_{t \geq 0}$. Finally, in
appendix~\ref{appendix:multiplicative-updates:negative-targets} we show how
to deal with sequences $(\vec{w}_{t}^{+})_{t \geq 0}$ and $(\vec{w}_{t}^{-})_{t \geq 0}$
simultaneously.

%% file: appendix/main_proofs/key_propositions.tex
\subsection{The Key Propositions}
\label{appendix:main-proofs:key-propositions}

In this section we state the key propositions appearing in the proof of
Theorem~\ref{thm:constant-step-sizes} and discuss their implications.

Proposition~\ref{proposition:dealing-with-proportional-errors}
is the core of our proofs.
It allows to ignore the error sequence $(\vec{p}_{t})_{t \geq 0}$
as long as the RIP constant $\delta$ is small enough.
That is, suppose that $\norm{\vec{b}_{t}}_{\infty} \lesssim \zeta$ for some
$\zeta > 0$.
Proposition~\ref{proposition:dealing-with-proportional-errors}
states that if
$\delta \lesssim 1 / \sqrt{k} ( \log \frac{\wmax}{\zeta} \vee 1 )$
then it is possible to fit the signal sequence $(\vec{s}_{t})_{t \geq 0}$
to $\wstar$
up to precision proportional
to $\zeta$ while keeping the error sequence $(\vec{e}_{t})_{t \geq 0}$
arbitrarily small.
See
appendix~\ref{appendix:multiplicative-updates:proof-of-key-proposition}
for proof.

\begin{proposition}
  \label{proposition:dealing-with-proportional-errors}
  Consider the setting of updates given in
  equations~\eqref{eq:error-decompositions} and
  \eqref{eq:updates-equation-using-b-p-notation}.
  Fix any $0 < \zeta \leq \wmax$ and let
  $\gamma = \frac{C_{\gamma}}{
    \ceil{\log_{2} \frac{\wmax}{\zeta}}}$
  where $C_{\gamma}$ is some small enough absolute constant.
  Suppose the error sequences $(\vec{b}_{t})_{t \geq 0}$ and $(\vec{p}_{t})_{t \geq 0}$
  for any $t \geq 0$ satisfy the following:
  \begin{align*}
    \norm{\vec{b}_{t}}_{\infty}
    &\leq
    C_{b}\zeta - \alpha, \\
    \norm{\vec{p}_{t}}_{\infty}
    &\leq
    \gamma \norm{\vec{s}_{t} - \wstar}_{\infty},
  \end{align*}
  where $C_{b}$ is some small enough absolute constant.
  If
  the step size satisfies
  $
    \eta \leq \frac{5}{96\wmax}
  $
  and the initialization satisfies
  $
    \alpha
    \leq
    1
    \wedge \frac{\zeta}{3(\wmax)^{2}}
    \wedge \frac{1}{2} \sqrt{\wmin}
  $
  Then, for some
  $
    T
    = O \left(\frac{1}{\eta\zeta} \log \frac{1}{\alpha}\right)
  $
  and any $0 \leq t \leq T$
  we have
  \begin{align*}
    \norm{\vec{s}_{T} - \wstar}_{\infty}
    &\leq \zeta, \\
    \norm{\vec{e}_{t}}_{\infty}
    &\leq \alpha.
  \end{align*}
\end{proposition}

The proof of Theorem~\ref{thm:constant-step-sizes} in the hard regime
when $\wmin \lesssim \maxnoise \vee \varepsilon$
is then just a simple application of the above theorem with
$\zeta = \frac{2}{C_{b}}(\maxnoise \vee \varepsilon)$ where the absolute
constant $C_{b}$ needs to satisfy the conditions of the above proposition.

On the other hand, if $\wmin \gtrsim \maxnoise \vee \varepsilon$
which happens as soon as we choose small enough $\varepsilon$ and when
we get enough data points $n$, we can apply
Proposition~\ref{proposition:dealing-with-proportional-errors}
with
$\zeta = \frac{1}{5}\wmin$.
Then, after $O(\frac{1}{\eta \wmin} \log \frac{1}{\alpha})$
iterations we can keep $\norm{\vec{e}_{t}}_{\infty}$ below $\alpha$
while $\norm{\vec{s}_{t} - \wstar}_{\infty} \leq \frac{1}{5} \wmin$.
From this point onward, the convergence of the signal sequence
$(\vec{s}_{t})_{t \geq 0}$ does not depend on $\alpha$ anymore
while the error term is smaller than $\alpha$.
We can hence fit the signal sequence to $\wstar$ up to precision
$\norm{\noisevec \odot \id{S}}_{\infty} \vee \varepsilon$
while keeping $\norm{\vec{e}_{t}}_{\infty}$ arbitrarily small.
This idea is formalized in the following proposition.

\begin{proposition}
  \label{proposition:easy-setting-convergence}
  Consider the setting of updates given in
  equations~\eqref{eq:error-decompositions} and
  \eqref{eq:updates-equation-using-b-p-notation}.
  Fix any $\varepsilon > 0$ and
  suppose that the error sequences $(\vec{b}_{t})_{t \geq 0}$ and
  $(\vec{p}_{t})_{t \geq 0}$ for any $t \geq 0$ satisfy
  \begin{align*}
    \norm{\vec{b}_{t} \odot \id{i}}_{\infty}
    &\leq B_{i} \leq \frac{1}{10}\wmin, \\
    \norm{\vec{p}_{t}}_{\infty}
    &\leq
    \frac{1}{20} \norm{\vec{s}_{t} - \wstar}_{\infty}.
  \end{align*}
  Suppose that
  $$
    \norm{\vec{s}_{0} - \wstar}_{\infty}
    \leq \frac{1}{5} \wmin.
  $$
  Let the step size satisfy $\eta \leq \frac{5}{96 \wmax}$. Then for
  all $t \geq 0$
  $$
    \norm{\vec{s}_{t} - \wstar}_{\infty} \leq \frac{1}{5}\wmin
  $$
  and for any
  $t \geq \frac{45}{32 \eta \wmin} \log \frac{\wmin}{\varepsilon}$
  and for any $i \in S$ we have
  $$
    \abs{s_{t,i} - w^{\star}_{i}}
    \lesssim \delta \sqrt{k} \max_{j \in S} B_{j}
    \vee B_{i}
    \vee \varepsilon.
  $$
\end{proposition}

%% file: appendix/main_proofs/technical_lemmas.tex
\subsection{Technical Lemmas}
\label{appendix:main-proofs:technical-lemmas}

In this section we state some technical lemmas which will be used to prove
Theorem~\ref{thm:constant-step-sizes} and its corollaries.
Proofs for all of the lemmas stated in this section can be found in
Appendix~\ref{appendix:missing-proofs}.

We begin with Lemma~\ref{lemma:controlling-errors} which allows to upper-bound
the error sequence $(\vec{e}_{t})_{t \geq 0}$ in terms of sequences
$(\vec{b}_{t})_{t \geq 0}$ and $(\vec{p}_{t})_{t \geq 0}$.

\begin{lemma}
  \label{lemma:controlling-errors}
  Consider the setting of updates given in
  equations\eqref{eq:error-decompositions} and
  \eqref{eq:updates-equation-using-b-p-notation}.
  Suppose that $\norm{\vec{e}_{0}}_{\infty} \leq \frac{1}{4}\wmin$
  and that there exists some $B \in \mathbb{R}$ such that
  for all $t$ we have $\norm{\vec{b}_{t}}_{\infty} + \norm{\vec{p}_{t}}_{\infty} \leq B$.
  Then, if $\eta \leq \frac{1}{12(\wmax + B)}$ for any $t \geq 0$ we have
  $$
    \norm{\vec{e}_{t}}_{\infty}
    \leq
    \norm{\vec{e}_{0}}_{\infty}
    \prod_{i=0}^{t-1}(1 +
    4\eta(\norm{\vec{b}_{i}}_{\infty} + \norm{\vec{p}_{i}}_{\infty}))^{2}.
  $$
\end{lemma}

Once we have an upper-bound on
$\norm{\vec{p}_{t}}_{\infty} + \norm{\vec{b}_{t}}_{\infty}$
we can apply Lemma~\ref{lemma:selective-fitting} to control the size
of $\norm{\vec{e}_{t}}_{\infty}$.
This happens, for example, in the easy setting when $\wmin \gtrsim \maxnoise
\vee \varepsilon$ where after the application of
Proposition~\ref{proposition:dealing-with-proportional-errors} we have
$\norm{\vec{p}_{t}}_{\infty} + \norm{\vec{b}_{t}}_{\infty}
\lesssim \wmin$.

\begin{lemma}
  \label{lemma:selective-fitting}
  Let $(b_{t})_{t \geq 0}$ be a sequence such that for any $t \geq 0$ we have
  $\abs{b_{t}} \leq B$ for some $B > 0$.
  Let the step size $\eta$ satisfy $\eta \leq \frac{1}{8B}$ and
  consider a one-dimensional sequence $(x_{t})_{t \geq 0}$ given by
  \begin{align*}
    0 &< x_{0} < 1, \\
    x_{t+1} &= x_{t}(1 + 4\eta b_{t})^{2}.
  \end{align*}
  Then for any $t \leq \frac{1}{32\eta B} \log \frac{1}{x_{0}^{2}}$ we have
  $$
    x_{t} \leq \sqrt{x_{0}}.
  $$
\end{lemma}

We now introduce the following two lemmas related to the restricted isometry
property. Lemma~\ref{lemma:rip-assumption-pt} allows to control the
$\ell_{\infty}$ norm of the sequence $(\vec{p}_{t})_{t \geq 0}$.
Lemma~\ref{lemma:column-normalization} allows to control the
$\ell_{\infty}$ norm of the term
$\frac{1}{n} \XtX \vec{e}_{t}$ arising in the bounded errors sequence
$(\vec{b}_{t})_{t \geq 0}$.

\begin{lemma}
  \label{lemma:rip-assumption-pt}
  Suppose that $\frac{1}{\sqrt{n}}\X$ is a $n \times d$ matrix satisfying the
  $(k + 1, \delta)$-RIP.
  If $\vec{z} \in \mathbb{R}^{d}$ is a $k$-sparse
  vector then
  $$
    \norm{\left( \frac{1}{n} \XtX - I \right)\vec{z}}_{\infty}
    \leq
    \sqrt{k} \delta \norm{\vec{z}}_{\infty}.
  $$
\end{lemma}

\begin{lemma}
  \label{lemma:column-normalization}
  Suppose that $\frac{1}{\sqrt{n}}\X$ is a $n \times d$ matrix satisfying the
  $(1, \delta)$-RIP with $0 \leq \delta \leq 1$ and let $\vec{X}_{i}$ be the
  $i\th$ column of $\X$.
  Then
  $$
    \max_{i} \norm{\frac{1}{\sqrt{n}} \vec{X}_{i}}_{2} \leq \sqrt{2}
  $$
  and for any vector $\vec{z} \in \mathbb{R}^{d}$ we have
  $$
    \norm{\frac{1}{n} \XtX \vec{z}}_{\infty} \leq 2d\norm{\vec{z}}_{\infty}.
  $$
\end{lemma}


Finally, we introduce a lemma upper-bounding the maximum noise term
\maxnoise when $\vec{\xi}$ is subGaussian with independent entries
and the design matrix $\X$ is treated
as fixed.
\begin{lemma}
  \label{lemma:bounding-max-noise}
  Let $\frac{1}{\sqrt{n}} \X$ be a $n \times d$ matrix such that
  the $\ell_{2}$ norms of its columns are bounded by some absolute constant
  $C$. Let $\xi \in \mathbb{R}^{n}$ be a vector of independent
  $\sigma^{2}$-subGaussian random variables. Then, with probability at
  least $1 - \frac{1}{8d^{3}}$
  $$
  \maxnoise \lesssim \sqrt{\frac{\sigma^{2} \log d}{n}}.
  $$
\end{lemma}

%% file: appendix/main_proofs/thm_constant_step_size.tex
\subsection{Proof of Theorem~\ref{thm:constant-step-sizes}}
\label{appendix:main-proofs:proof-of-thm-constant-step-size}

Let $C_{b}$ and $C_{\gamma}$ be small enough absolute positive constants that
satisfy conditions of
Proposition~\ref{proposition:dealing-with-proportional-errors}.

Let
$$
\zeta
\coloneqq
\frac{1}{5}\wmin
\vee \frac{2}{C_{b}}\maxnoise
\vee \frac{2}{C_{b}} \varepsilon.
$$
and suppose that
$$
  \delta
  \leq
  \frac{C_{\gamma}}{\sqrt{k}
    \left(
    \log_{2}
    \frac{\wmax}{\zeta} + 1
    \right)}.
$$

Setting
$$
\alpha
\leq
1
\wedge \frac{\varepsilon^{2}}{(2d + 1)^{2}}
\wedge \frac{\varepsilon}{\wmax}
\wedge \frac{\zeta}{3(\wmax)^{2}}
\wedge \frac{1}{2}\sqrt{\wmin}
$$
we satisfy pre-conditions of
Proposition~\ref{proposition:dealing-with-proportional-errors}.
Also,
by Lemma~\ref{lemma:column-normalization}
as long as $\norm{\vec{e}_{t}}_{\infty} \leq \sqrt{\alpha}$
we have
$$
\norm{\frac{1}{n}\XtX \vec{e}_{t}}_{\infty} + \alpha
\leq (2d + 1)\sqrt{\alpha}
\leq \varepsilon.
$$
It follows that as long as $\norm{\vec{e}_{t}}_{\infty} \leq \sqrt{\alpha}$
we can upper bound $\norm{\vec{b}_{t}}_{\infty} + \alpha$ as follows:
$$
\norm{\vec{b}_{t}}_{\infty} + \alpha
\leq \maxnoise + \varepsilon
\leq
C_{b}\cdot \frac{2}{C_{b}} \left(\maxnoise \vee \varepsilon\right)
\leq C_{b} \zeta.
$$
By Lemma~\ref{lemma:rip-assumption-pt} we also have
$$
  \norm{\vec{p}_{t}}_{\infty}
  \leq \frac{C_{\gamma}}{\ceil{\log_{2} \frac{\wmax}{\zeta}}}
    \norm{\vec{s}_{t} - \wstar}_{\infty}.
$$
and so both sequences $(\vec{b}_{t})_{t \geq 0}$ and
$(\vec{p}_{t})_{t \geq 0}$ satisfy the assumptions of
Proposition~\ref{proposition:dealing-with-proportional-errors} conditionally
on $\norm{\vec{e}_{t}}_{\infty}$ staying below $\sqrt{\alpha}$.
If $\zeta \geq \wmax$ then the statement of our theorem already holds at
$t = 0$ and we are done. Otherwise, applying
Proposition~\ref{proposition:dealing-with-proportional-errors}
we have after
$$
T = O\left(\frac{1}{\eta \zeta} \log \frac{1}{\alpha} \right)
$$
iterations
\begin{align*}
  \norm{\vec{s}_{T} - \wstar}_{\infty} &\leq \zeta \\
  \norm{\vec{e}_{T}}_{\infty} \leq \alpha.
\end{align*}

If $\frac{1}{5} \wmin \leq \frac{2}{C_{b}} \maxnoise \vee \frac{2}{C_{b}}
\varepsilon$ then we are in what we refer to as the hard regime and we are
done.

On the other hand, suppose that
$\frac{1}{5} \wmin > \frac{2}{C_{b}} \maxnoise \vee \frac{2}{C_{b}}
\varepsilon$
so that we are working in the easy regime and $\zeta = \frac{1}{5}\wmin$.

Conditionally on $\norm{\vec{e}_{t}}_{\infty} \leq \sqrt{\alpha}$,
$\norm{\vec{p}_{t}}_{\infty}$ stays below $C_{\gamma}\cdot \frac{1}{5}\wmin$
by Proposition~\ref{proposition:easy-setting-convergence}.
Hence,
$$\norm{\vec{b}_{t}}_{\infty} + \norm{\vec{p}_{t}}_{\infty}
\leq (C_{b} + C_{\gamma})\cdot\frac{1}{5}\wmin.$$
Applying Lemmas~\ref{lemma:controlling-errors} and
\ref{lemma:selective-fitting}
we can maintain that $\norm{\vec{e}_{t}}_{\infty} \leq \sqrt{\alpha}$
for at least another
$\frac{5}{16(C_{b} + C_{\gamma})\eta \wmin} \log \frac{1}{\alpha}$
iterations after an application of
Proposition~\ref{proposition:dealing-with-proportional-errors}.
Crucially, with a small enough $\alpha$ we can maintain the above property for
as long as we want and in our case here we need $\alpha \leq \varepsilon /
\wmax$.

Choosing small enough $C_{b}$ and $C_{\gamma}$ so that
$C_{b} + C_{\gamma} \leq \frac{2}{9}$ and
$C_{\gamma} \leq \frac{1}{20}$
and applying Proposition~\ref{proposition:dealing-with-proportional-errors}
we have after
$$
  T'
  \coloneqq
  T + \frac{45}{32\eta\wmin} \log \frac{\wmin}{\varepsilon}
  \leq
  T + \frac{5}{16(C_{b} + C_{\gamma})\eta \wmin} \log \frac{1}{\alpha}
$$
iterations
$$
  \norm{\vec{e}_{T'}}_{\infty} \leq \sqrt{\alpha}
$$
and for any $i \in S$
$$
  \abs{s_{T',i} - w^{\star}_{i}}
  \lesssim
  \sqrt{k}\delta \norm{\noisevec \odot \id{S}}_{\infty}
  \vee
  \abs{\left(\noisevec\right)_{i}}
  \vee
  \varepsilon.
$$

Finally, noting that for all $t \leq T'$ we have
$$\abs{w_{t,i} - w^{\star}}
\leq \abs{s_{t,i} - w^{\star}_{i}} + \abs{e_{t,i}}
\leq \abs{s_{t,i} - w^{\star}_{i}} + \sqrt{\alpha}
\leq \abs{s_{t,i} - w^{\star}_{i}} + \varepsilon
$$
our result follows.

%% file: appendix/main_proofs/corollaries.tex
\subsection{Proofs of Corollaries}
\label{appendix:main-proofs:proofs-of-corollaries}

\begin{proof}[Proof of Corollary~\ref{corollary:noiseless}]
Since $\xi = 0$ the bound in Theorem~\ref{thm:constant-step-sizes}
directly reduces to
$$
  \norm{\vec{w}_{t} - \wstar}_{2}^{2}
  \lesssim \sum_{i \in S} \varepsilon^{2}
    + \sum_{i \notin S} \alpha
  \leq k \varepsilon^{2} + (d-k)\frac{\varepsilon^{2}}{(2d + 1)^{2}}
  \lesssim k \varepsilon^{2}.
$$
\end{proof}

\begin{proof}[Proof of Corollary~\ref{corollary:minimax}]
  By Lemma~\ref{lemma:column-normalization} and
  the proof of Lemma~\ref{lemma:bounding-max-noise} with probability at least
  $1 - 1/(8d^{3})$ we have
  $$
    \maxnoise \leq 4\frac{\sqrt{2\sigma^{2}\log(2d)}}{\sqrt{n}}.
  $$
  Hence, letting $\varepsilon = 4\frac{\sqrt{2\sigma^{2}\log(2d)}}{\sqrt{n}}$,
  Theorem~\ref{thm:constant-step-sizes} implies with probability
  at least $1 - 1/(8d^{3})$
  $$
  \norm{\vec{w}_{t} - \wstar}_{2}^{2}
  \lesssim \sum_{i \in S} \varepsilon^{2}
    + \sum_{i \notin S} \alpha
  \leq k \varepsilon^{2} + (d-k)\frac{\varepsilon^{2}}{(2d + 1)^{2}}
  \lesssim \frac{k\sigma^{2} \log d}{n}.
  $$
\end{proof}

\begin{proof}[Proof of Corollary~\ref{corollary:adaptivity}]
  We use the same argument as in proof of Corollary~\ref{corollary:adaptivity}
  with the term $\inlinemaxnoise$ replaced with
  $\sqrt{k}\delta\inlinenorm{\Xt \vec{\xi} \odot \id{S}}_{\infty} / n$.
  Since $\sqrt{k}\delta \lesssim 1$
  an identical result holds with $d$ replaced with $k$.
\end{proof}

%% file: appendix/multiplicative_updates.tex
\section{Understanding Multiplicative Update Sequences}
\label{appendix:multiplicative-update-sequences}

In this section of the appendix, we provide technical lemmas to understand
the behavior of multiplicative updates sequences. We then prove
Propositions~\ref{proposition:dealing-with-proportional-errors} and
\ref{proposition:easy-setting-convergence}.

\subfile{appendix/multiplicative_updates/basic_lemmas.tex}

\subfile{appendix/multiplicative_updates/bounded_errors.tex}

\subfile{appendix/multiplicative_updates/proportional_errors.tex}

\subfile{appendix/multiplicative_updates/negative_targets.tex}

\subfile{appendix/multiplicative_updates/proof_of_key_proposition.tex}

\subfile{appendix/multiplicative_updates/proof_of_easy_setting_proposition.tex}

%% file: appendix/multiplicative_updates/basic_lemmas.tex
\subsection{Basic Lemmas}
\label{appendix:multiplicative-updates:basic-lemmas}

In this section we analyze one-dimensional sequences with positive
target corresponding to gradient descent updates without any perturbations.
That is,
this section corresponds to parametrization $\vec{w}_{t} = \vec{u}_{t} \odot
\vec{u}_{t}$
and gradient descent updates under assumption that
$\frac{1}{n} \XtX = \matrixid$ and ignoring the error sequences $(\vec{b}_{t})_{t \geq 0}$
and $(\vec{p}_{t})_{t \geq 0}$ given in equation~\eqref{eq:error-decompositions}
completely. We will hence look at one-dimensional sequences of the form
\begin{align}
  \begin{split}
    \label{eq:one-dimensional-sequence}
    0 < x_{0} &= \alpha^{2} < x^{\star} \\
    x_{t+1} &= x_{t}(1 - 4\eta(x_{t} - x^{\star}))^{2}.
  \end{split}
\end{align}
Recall the definition of gradient descent updates given in
equations~\eqref{eq:error-decompositions} and
\eqref{eq:updates-equation-using-b-p-notation} and let $\vec{v}_{t} = 0$ for all $t$.
Ignoring the effects of the sequence $(\vec{p}_{t})_{t \geq 0}$
and the term $\frac{1}{n} \XtX
\vec{e}_{t}$ one can immediately see that $\norm{\id{S^{c}} \odot
  \vec{w}_{t}}_{\infty}$ grows at most as fast as
the sequence $(x_{t})_{t \geq 0}$ given in
equation~\eqref{eq:one-dimensional-sequence} with $x^{\star} = \maxnoise$.
Surely, for any $i \in S$ such that $0 < w_{i}^{\star} < \maxnoise$
we cannot fit the $i-th$ component of $w^{\star}$ without fitting any of the
noise variables $\id{S^{c}} \odot w_{t}$. On the other hand,
for any $i \in S$ such that $w^{\star}_{i} \gg \maxnoise$
can fit the sequence $(x_{t})_{t \geq 0}$ with $x^{\star} = w^{\star}_{i}$
while keeping all of the noise variables arbitrarily small, as we shall see
in this section.

\begin{figure}
  \centering
  \begin{subfigure}{.5\textwidth}
    \centering
    \includegraphics[width=\linewidth]{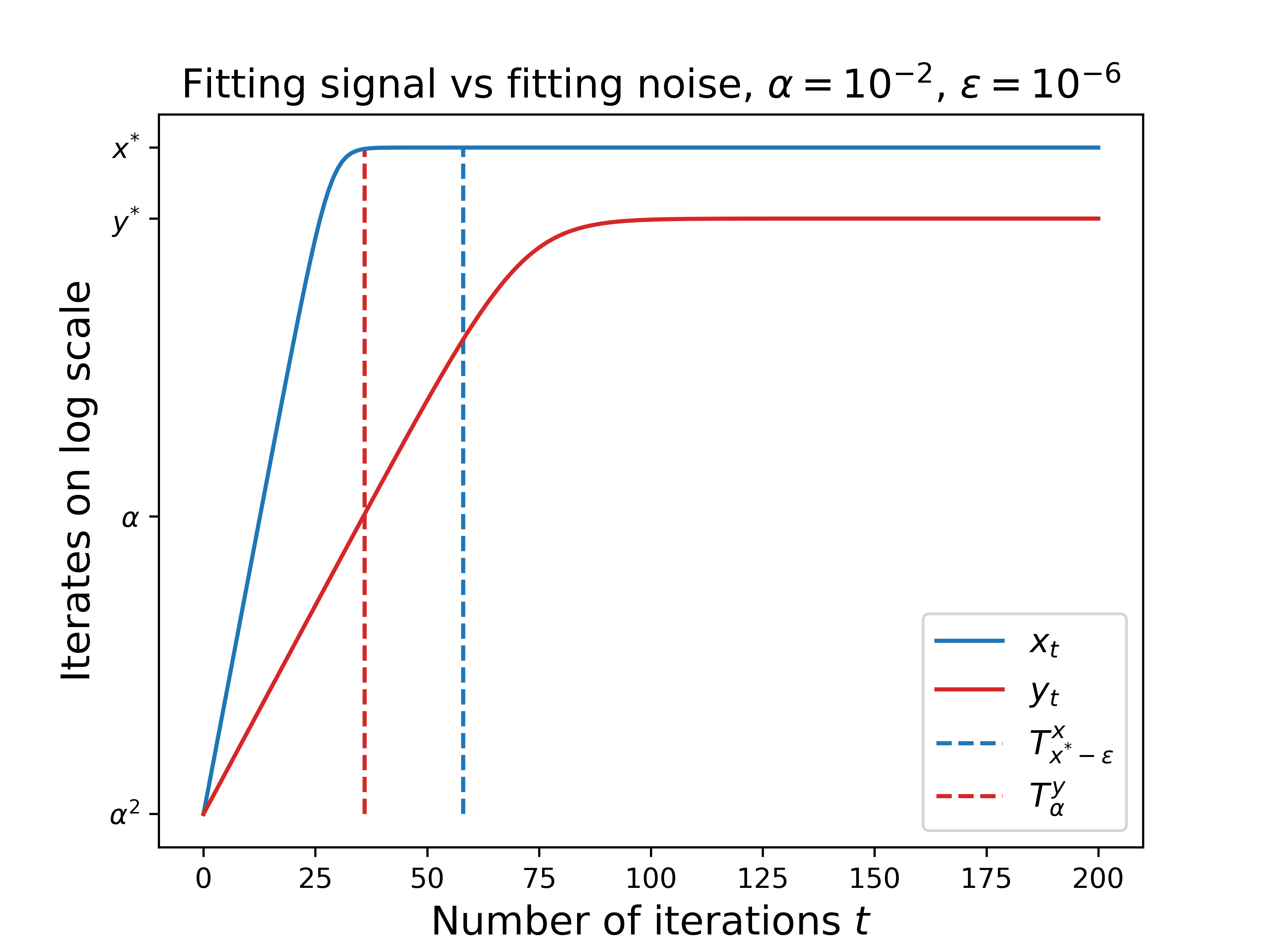}
  \end{subfigure}%
  \begin{subfigure}{.5\textwidth}
    \centering
    \includegraphics[width=\linewidth]{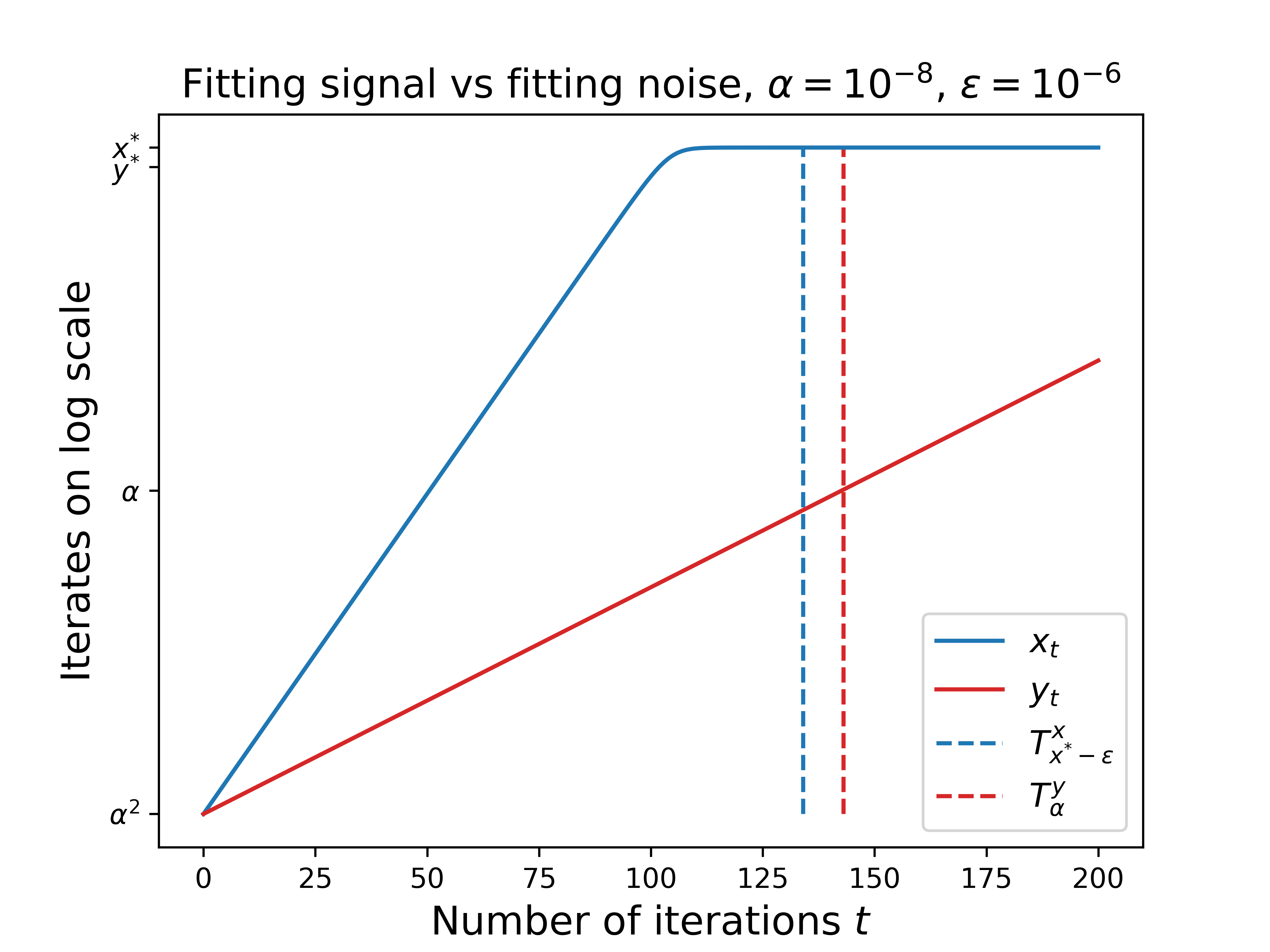}
  \end{subfigure}
  \caption{The blue and red lines represent the signal sequence $(x_{t})_{t
    \geq 0}$ and the noise sequence $(y_{t})_{t \geq 0}$ plotted on $\log$
    scale. The vertical blue and red dashed lines show the hitting times
    $T^{x}_{x^{\star} - \varepsilon}$ and $T^{y}_{\alpha}$ so that we want the
    blue vertical line to appear on the left side of the red vertical line.
    Both plots use the same values of $x^{\star}$, $y^{\star}$ and $\varepsilon$.
    However, the plot on the left is plotted with $\alpha =
    10^{-2}$ and the plot on the right is plotted with
    $\alpha = 10^{-8}$. This shows the effect of decreasing initialization
    size;
    both vertical lines are pushed to the right, but the red vertical line
    is pushed at a faster pace.}
  \label{fig:fitting-signal-vs-noise}
\end{figure}

We can hence formulate a precise question that we answer in this
section. Consider two sequences $(x_{t})_{t \geq 0}$ and $(y_{t})_{t \geq 0}$
with updates as in equation~\eqref{eq:one-dimensional-sequence} with targets
$x^{\star}$ and $y^{\star}$ respectively.
One should think of the sequence $(y_{t})_{t \geq 0}$ as a sequence fitting
the noise, so that $y^{\star} = \maxnoise$. Let $T^{y}_{\alpha}$ be the smallest
$t \geq 0$ such that $y_{t} \geq \alpha$.
On the other hand, one should think of sequence $(x_{t})_{t \geq 0}$
as a sequence fitting the signal.
Let $T^{x}_{x^{\star} - \varepsilon}$ be the smallest $t$ such that
$x_{t} \geq x^{\star} - \varepsilon$.
Since we want to fit the sequence $(x_{t})_{t \geq 0}$ to $x^{\star}$ within
$\varepsilon$ error before $(y_{t})_{t \geq 0}$ exceeds $\alpha$
we want $T^{x}_{x^{\star} - \varepsilon} \leq T^{y}_{\alpha}$.
This can only hold if the variables $x^{\star}, y^{\star}, \alpha$ and $\varepsilon$
satisfy certain conditions.
For instance, decreasing $\varepsilon$ will increase $T^{x}_{x^{\star} -
  \varepsilon}$ without changing $T^{y}_{\alpha}$.
Also, if $x^{\star} < y^{\star}$
then satisfying
$T^{x}_{x^{\star} - \varepsilon} \leq T^{y}_{\alpha}$ is impossible for
sufficiently small $\varepsilon$.
However, as we shall see in this section, if $x^{\star}$ is sufficiently bigger
than $y^{\star}$ then for any $\varepsilon > 0$ one can choose a small enough
$\alpha$ such that $T^{x}_{x^{\star} - \varepsilon} \leq T^{y}_{\alpha}$.
To see this intuitively, note that if we ignore what happens when $x_{t}$
gets close to $x^{\star}$, the sequence $(x_{t})_{t \geq 0}$
behaves as an exponential function
$t \mapsto \alpha^{2}(1 + 4\eta x^{\star})^{2t}$ while the sequence $y^{\star}$
behaves as $t \mapsto \alpha^{2}(1 + 4\eta y^{\star})^{2t}$.
Since exponential function is very sensitive to its base, we can
make the gap between
$\alpha^{2}(1 + 4\eta x^{\star})^{2t}$ and
$\alpha^{2}(1 + 4\eta y^{\star})^{2t}$
as big as we want by decreasing $\alpha$ and increasing $t$.
This intuition is depicted in Figure~\ref{fig:fitting-signal-vs-noise}.

With the above discussion in mind, in this section we will
quantitatively formalize under what conditions on $x^{\star}, y^{\star}, \alpha$ and
$\varepsilon$ the inequality
$T^{x}_{x^{\star} - \varepsilon} \leq T^{y}_{\alpha}$ hold.
We begin by showing that for small enough step
sizes, multiplicative update sequences given in
equation~\eqref{eq:one-dimensional-sequence} behave monotonically.

\begin{lemma}[Iterates behave monotonically]
  \label{lemma:iterates-behave-monotonically}
  Let $\eta > 0$ be the step size and suppose that updates are given by
  \begin{align*}
    x_{t + 1} &= x_{t}(1 - 4\eta(x_{t} - x^{\star}))^{2}.
  \end{align*}
  Then the following holds
  \begin{enumerate}
    \item If $0 < x_{0} \leq x^{\star}$ and
          $\eta \leq \frac{1}{8 x^{\star}}$
          then for any $t > 0$ we have
          $x_{0} \leq x_{t-1} \leq x_{t} \leq x^{\star}$.
    \item If $x^{\star} \leq x_{0} \leq \frac{3}{2}x^{\star}$ and
          $\eta \leq \frac{1}{12 x^{\star}}$ then for any $t > 0$ we have
          $x^{\star} \leq x_{t} \leq x_{t-1} \leq x_{0}$.
  \end{enumerate}
\end{lemma}

\begin{proof}
  Note that if $x_{0} \leq x_{t} \leq x^{\star}$ then $x_{t} - x^{\star} \leq 0$
  and hence $x_{t+1} \geq x_{t}$. Thus for the first part it is enough to
  show that for all $t \geq 0$ we have $x_{t} \leq x^{\star}$.

  Assume for a contradiction that exists $t$ such that
  \begin{align*}
    x_{0} \leq x_{t} \leq x^{\star}, \\
    x_{t+1} > x^{\star}.
  \end{align*}
  Plugging in the update rule for $x_{t+1}$ we can rewrite the above as
  \begin{align*}
    x_{t}
    &\leq x^{\star} \\
    &< x_{t}(1 - 4\eta(x_{t} - x^{\star}))^{2} \\
    &\leq x_{t} \left(
      1 + \frac{1}{2} - \frac{x_{t}}{2x^{\star}}
    \right)^{2}
  \end{align*}
  Letting $\lambda \coloneqq \frac{x_{t}}{x^{\star}}$ we then have by our
  assumption above $0 < \lambda \leq 1$. The above inequality then gives us
  $$
    \sqrt{\frac{1}{\lambda}}
    < 3/2 - \frac{1}{2} \lambda
  $$
  And hence for $0 < \lambda \leq 1$ we have
  $f(\lambda) \coloneqq \sqrt{\frac{1}{\lambda}} + \frac{1}{2} \lambda < 3/2$.
  Since for $0 < \lambda < 1$ we also have
  $f'(\lambda) = \frac{1}{2}(1 - \frac{1}{\lambda^{3/2}}) < 0$
  and so $f(\lambda) \geq f(1) = 3/2$.
  This gives us the desired contradiction and concludes our proof for the first
  part.

  We will now prove the send part. Similarly to the first part, we just need to
  show that for all $t \geq 0$ we have $x_{t} \geq x^{\star}$. Suppose that
  $\frac{3}{2}x^{\star} \geq x_{t} \geq x^{\star}$ and hence we
  can write $x_{t} = x^{\star}(1 + \gamma)$ for some $\gamma \in [0, \frac{1}{2}]$.
  Then we have
  \begin{align*}
    x_{t+1}
    &=
    (1 + \gamma)x^{\star}(1 - 4\eta \gamma x^{\star})^{2} \\
    &\geq
    (1 + \gamma)x^{\star}(1 - \frac{1}{3}\gamma)^{2}.
  \end{align*}
  One may verify that the polynomial $(1 + \gamma)(1 - \frac{1}{3} \gamma)^{2}$
  is no smaller than one for $0 \leq \gamma \leq \frac{1}{2}$
  which finishes the second part of our proof.
\end{proof}

While the above lemma tells us that for small enough step sizes the iterates
are monotonic and bounded, the following two lemmas tell us that we are
converging to the target exponentially fast. We first look at the behavior
near convergence.

\begin{lemma}[Iterates behaviour near convergence]
  \label{lemma:iterates-behaviour-near-convergence}
  Consider the setting of
  Lemma~\ref{lemma:iterates-behave-monotonically}.
  Let $x^{\star} > 0$ and and suppose that
  $\abs{x_{0} - x^{\star}} \leq \frac{1}{2} x^{\star}$.
  Then the following holds.
  \begin{enumerate}
    \item If $0 < x_{0} \leq x^{\star}$ and $\eta \leq \frac{1}{8x^{\star}}$ then for
      any $t \geq \frac{1}{4 \eta x^{\star}}$ we have
          $$
            0 \leq x^{\star} - x_{t}
               \leq \frac{1}{2}\abs{x_{0} - x^{\star}}.
          $$
    \item If $x^{\star} \leq x_{0} \leq \frac{3}{2} x^{\star}$ and
          $\eta \leq \frac{1}{12x^{\star}}$ then for any
          $t \geq \frac{1}{8 \eta x^{\star}}$ we have
          $$
            0 \leq x_{t} - x^{\star}
              \leq \frac{1}{2}\abs{x_{0} - x^{\star}}.
          $$
  \end{enumerate}
\end{lemma}

\begin{proof}
  Let us write $\abs{x_{0} - x^{\star}} = \gamma x^{\star}$ where $\gamma \in [0, \frac{1}{2}]$.

  For the first part we have $x_{0} = (1 - \gamma)x^{\star}$.
  Note that while
  $x_{t} \leq (1 - \frac{\gamma}{2})x^{\star}$
  we have $x_{t+1} \geq x_{t}(1 + 4\eta\frac{\gamma}{2}x^{\star})$.
  Recall that by the Lemma~\ref{lemma:iterates-behave-monotonically} for all
  $t \geq 0$ we have $x_{t} \leq x^{\star}$.
  Hence to find $t$ such that
  $x^{\star} \geq x_{t} \geq (1 - \frac{\gamma}{2})x^{\star}$
  it is enough to find a big enough $t$ satisfying the following inequality
  \begin{align*}
    x_{0}(1 + 2\eta \gamma x^{\star})^{2t}
    \geq
    \left( 1 - \frac{\gamma}{2} \right)x^{\star}.
  \end{align*}
  Noting that for $x > 0$ ant $t \geq 1$ we have $(1 + x)^{t} \geq 1 + tx$
  we have
  \begin{align*}
    x_{0}(1 + 2\eta \gamma x^{\star})^{2t}
    &\geq
    x_{0}(1 + 4\eta \gamma x^{\star} t)
  \end{align*}
  and hence it is enough to find a big enough $t$ satisfying
  \begin{align*}
    &x_{0}(1 + 4\eta \gamma x^{\star} t)
    \geq \left( 1 - \frac{\gamma}{2} \right)x^{\star} \\
    \iff&
    4\eta\gamma x^{\star}t
    \geq \frac{\left(1 - \frac{\gamma}{2}\right)x^{\star} - x_{0}}
        {x_{0}} \\
    \iff&
    4\eta\gamma x^{\star}t
    \geq \frac{\gamma}
        {2(1 - \gamma)} \\
    \iff&
    t
    \geq \frac{1}{8\eta x^{\star}}
      \frac{1}{(1 - \gamma)}
  \end{align*}
  and since $\gamma \in [0, \frac{1}{2}]$ choosing $t \geq \frac{1}{4\eta
    x^{\star}}$ is enough.

  To deal with the second part, now let us write
  $x_{0} = x^{\star}(1 + \gamma)$. We will use a similar approach
  to the one used in the first part.
  If for some $x_{t}$ we have
  $x_{t} \leq (1 + \frac{\gamma}{2})x^{\star}$ by
  Lemma~\ref{lemma:iterates-behave-monotonically} we would be done.
  If $x_{t} > x^{\star}(1 + \frac{\gamma}{2})$ we have
  $x_{t+1} \leq x_{t}(1 - 4\eta\frac{\gamma}{2}x^{\star})^{2}$.
  This can happen for at most $\frac{1}{8 \eta x^{\star}}$
  iterations, since
  \begin{align*}
    &
    x_{0}(1 - 2 \eta \gamma x^{\star})^{2t}
    \leq x^{\star}(1 + \frac{\gamma}{2}) \\
    \iff&
    2t \log(1 - 2\eta \gamma x^{\star})
    \leq
    \log \frac{x^{\star}(1 + \frac{\gamma}{2})}{x^{0}} \\
    \iff&
    t
    \geq
    \frac{1}{2}
    \frac
    {\log \frac{x^{\star}(1 + \frac{\gamma}{2})}{x^{0}}}
    {\log(1 - 2\eta \gamma x^{\star}) }.
  \end{align*}
  We can deal with the term on the right hand side by noting that
  \begin{align*}
    \frac{1}{2}
    \frac
    {\log \frac{x^{\star}(1 + \frac{\gamma}{2})}{x^{0}}}
    {\log(1 - 2\eta \gamma x^{\star}) }
    &=
    \frac{1}{2}
    \frac
    {\log \frac{1 + \frac{\gamma}{2}}{1 + \gamma}}
    {\log(1 - 2\eta \gamma x^{\star}) } \\
    &\leq
    \frac{1}{2}
    \frac
    {\left( \frac{1 + \frac{\gamma}{2}}{1 + \gamma} - 1 \right)
     /
     \left( \frac{1 + \frac{\gamma}{2}}{1 + \gamma}\right)
    }
    {- 2 \eta \gamma x^{\star}} \\
    &=
    \frac{1}{2}
    \frac
    {-\frac{\gamma}{2}
     /
     \left( 1 + \frac{\gamma}{2} \right)
    }
    {- 2 \eta \gamma x^{\star}} \\
    &\leq \frac{1}{8\eta x^{\star}}.
  \end{align*}
  where in the
  second line we have used $\log x \leq x - 1$ and
  $\log x \geq \frac{x-1}{x}$. Note, however, that in the above inequalities
  both logarithms are negative, which is why the inequality signs are reversed.
\end{proof}

\begin{lemma}[Iterates approach target exponentially fast]
  \label{lemma:iterates-approach-target-exponentially-fast}
  Consider the setting of updates as in
  Lemma~\ref{lemma:iterates-behave-monotonically} and
  fix any $\varepsilon > 0$.
  \begin{enumerate}
    \item If $\varepsilon < \abs{x^{\star} - x_{0}} \leq \frac{1}{2} x^{\star}$ and
      $\eta \leq \frac{1}{12 x^{\star}}$ then for any
      $t \geq
      \frac{3}{8\eta x^{\star}}
      \log \frac{\abs{x^{\star} - x_{0}}}{\varepsilon}$
      we have
      $$
        \abs{x^{\star} - x_{t}} \leq \varepsilon.
      $$
    \item If $0 < x_{0} \leq \frac{1}{2}x^{\star}$ and $\eta \leq \frac{1}{8 x^{\star}}$
      then for any $t \geq
        \frac{3}{8 \eta x^{\star}}
        \log \frac{(x^{\star})^{2}}{4x_{0}\varepsilon}$
      we have
      $$
        x^{\star} - \varepsilon \leq x_{t} \leq x^{\star}.
      $$
  \end{enumerate}
\end{lemma}

\begin{proof}
  \hfill
  \begin{enumerate}
    \item
      To prove the first part we simply need to apply
      Lemma~\ref{lemma:iterates-behaviour-near-convergence}
      $\ceil{\log_{2} \frac{\abs{x^{\star} - x_{0}}}{\varepsilon}}$ times.
      Hence after $$\frac{\log_{2} e}{4\eta x^{\star}}
      \log \frac{\abs{x^{\star} - x_{0}}}{\varepsilon}
      \leq \frac{3}{8\eta x^{\star}}
      \log \frac{\abs{x^{\star} - x_{0}}}{\varepsilon}$$
      iterations we are done.
    \item
      We first need to find a lower-bound on time $t$ which ensures
      that $x_{t} \geq \frac{x^{\star}}{2}$. Note that while $x_{t} < \frac{x^{\star}}{2}$
      we have $x_{t+1} \geq x_{t}(1 + 2\eta x^{\star})^{2}$. Hence it is enough to
      choose a big enough $t$ such that
      \begin{align*}
        &
        x_{0}(1 + 2\eta x^{\star})^{2t}
        \geq
        \frac{x^{\star}}{2} \\
        \iff&
        t
        \geq
        \frac{1}{2} \frac{\log \frac{x^{\star}}{2x_{0}}}{\log(1 + 2\eta x^{\star})}.
      \end{align*}
      We can upper-bound the term on the right by using $\log x \geq \frac{x-1}{x}$
      as follows
      \begin{align*}
        \frac{1}{2} \frac{\log \frac{x^{\star}}{2x^{\star}}}{\log(1 + 2\eta x^{\star})}
        &\leq
        \frac{1}{2}
        \frac{1 + 2 \eta x^{\star}}{2 \eta x_{0}}
        \log \frac{x^{\star}}{2x_{0}} \\
        &\leq
        \frac{5}{16 \eta x^{\star}}
        \log \frac{x^{\star}}{2x_{0}}
      \end{align*}
      and so after
      $t \geq
        \frac{5}{16 \eta x^{\star}}
        \log \frac{x^{\star}}{2x_{0}}$
      we have $x_{t} \geq \frac{x^{\star}}{2}$.

      Now we can apply the first part to finish the proof.
      The total sufficient number of iterations is then
      \begin{align*}
        \frac{5}{16 \eta x^{\star}}
        \log \frac{x^{\star}}{2x_{0}}
        +
        \frac{3}{8 \eta x^{\star}}
        \log \frac{x^{\star}}{2\varepsilon}
        &\leq
        \frac{3}{8 \eta x^{\star}}
        \log \frac{x^{\star}}{2x_{0}}
        +
        \frac{3}{8 \eta x^{\star}}
        \log \frac{x^{\star}}{2\varepsilon} \\
        &=
        \frac{3}{8 \eta x^{\star}}
        \log \frac{(x^{\star})^{2}}{4x_{0}\varepsilon}.
      \end{align*}
  \end{enumerate}
\end{proof}

We are now able to answer the question that we set out at the beginning of this
section. That is, under what conditions on $x^{\star}, y^{\star}, \alpha$ and
$\varepsilon$ does the inequality
$T^{x}_{x^{\star} - \varepsilon} \leq T^{y}_{\alpha}$ hold?
Let $\eta \leq \frac{1}{8 x^{\star}}$ and suppose that $x^{\star} \geq 12y^{\star} > 0$.
Lemmas~\ref{lemma:iterates-behave-monotonically}~and~\
\ref{lemma:iterates-approach-target-exponentially-fast} then tell us,
that for any $\varepsilon > 0$ and any
$$
  t
  \geq
  \frac{12}{32 \eta x^{\star}}
    \log \frac{(x^{\star})^{2}}{\alpha^{2} \varepsilon}
$$
the sequence $x_{t}$ has converged up to precision $\varepsilon$. Hence
\begin{equation}
  \label{eq:fitting-signal-xt}
  T^{x}_{x^{\star} - \varepsilon}
  \leq
  \frac{12}{32 \eta x^{\star}}
    \log \frac{(x^{\star})^{2}}{\alpha^{2} \varepsilon}
\end{equation}
On the other hand, we can now apply Lemma~\ref{lemma:selective-fitting}
to see that for any
\begin{align*}
  t
  \leq
  \frac{12}{32 \eta x^{\star}} \log \frac{1}{\alpha^{4}}
  \leq
  \frac{1}{32 \eta y^{\star}} \log \frac{1}{\alpha^{4}}
\end{align*}
we have $y_{t} \leq \alpha$ and hence
\begin{equation}
  \label{eq:fitting-noise-yt}
  T^{y}_{\alpha} \geq
  \frac{12}{32 \eta x^{\star}} \log \frac{1}{\alpha^{4}}
\end{equation}

We can now see from equations~\eqref{eq:fitting-signal-xt} and
\eqref{eq:fitting-noise-yt} that it is enough to set
$\alpha \leq \frac{\sqrt{\varepsilon}}{x^{\star}}$ so that
$T^{x}_{x^{\star} - \varepsilon} \leq T^{y}_{\alpha}$ is satisfied which
answers our question.

%% file: appendix/multiplicative_updates/bounded_errors.tex
\subsection{Dealing With Bounded Errors}
\label{appendix:multiplicative-updates:bounded-errors}

\begin{figure}
  \centering
  \begin{subfigure}{.5\textwidth}
    \centering
    \includegraphics[width=\linewidth]{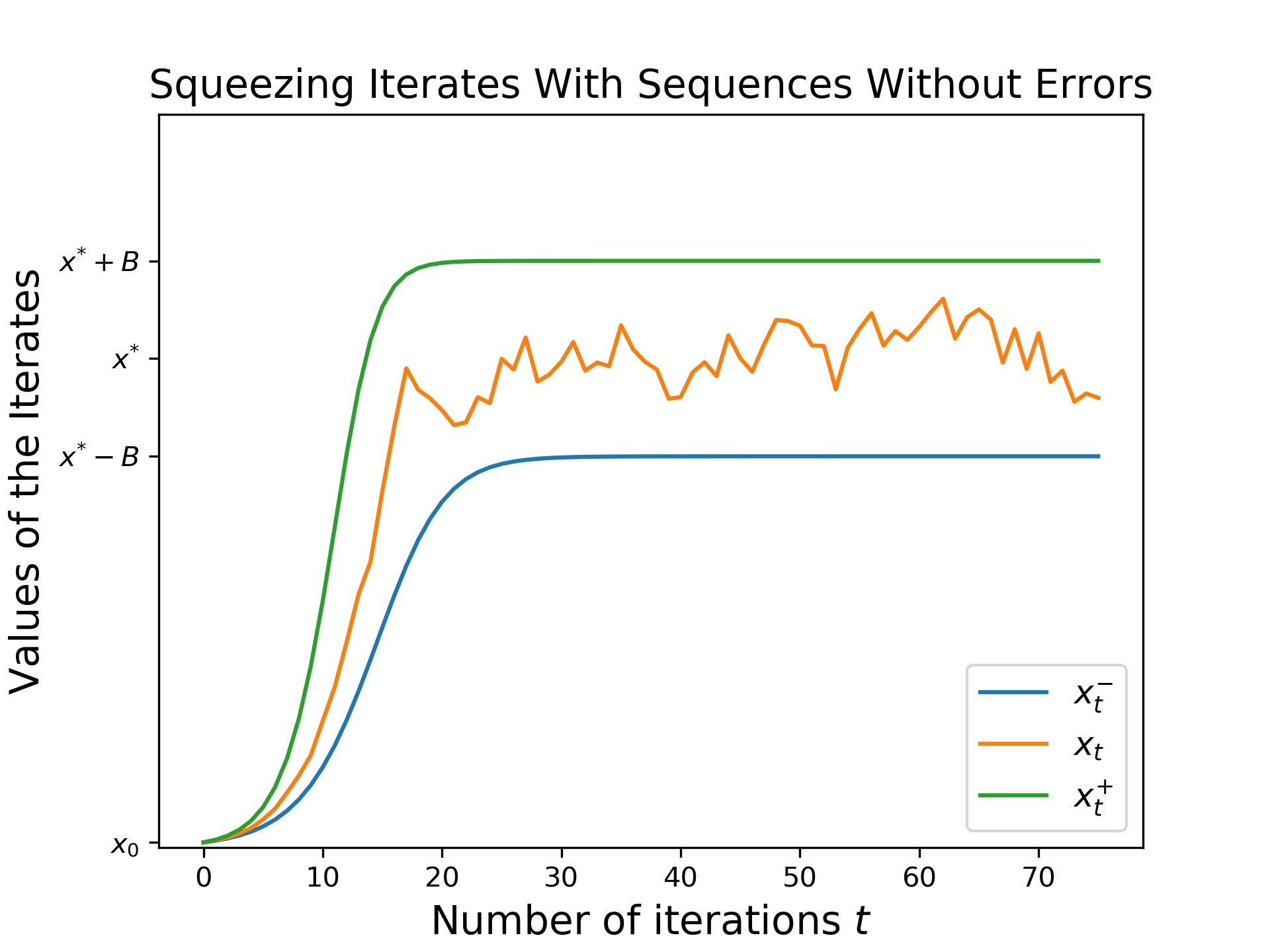}
  \end{subfigure}%
  \begin{subfigure}{.5\textwidth}
    \centering
    \includegraphics[width=\linewidth]{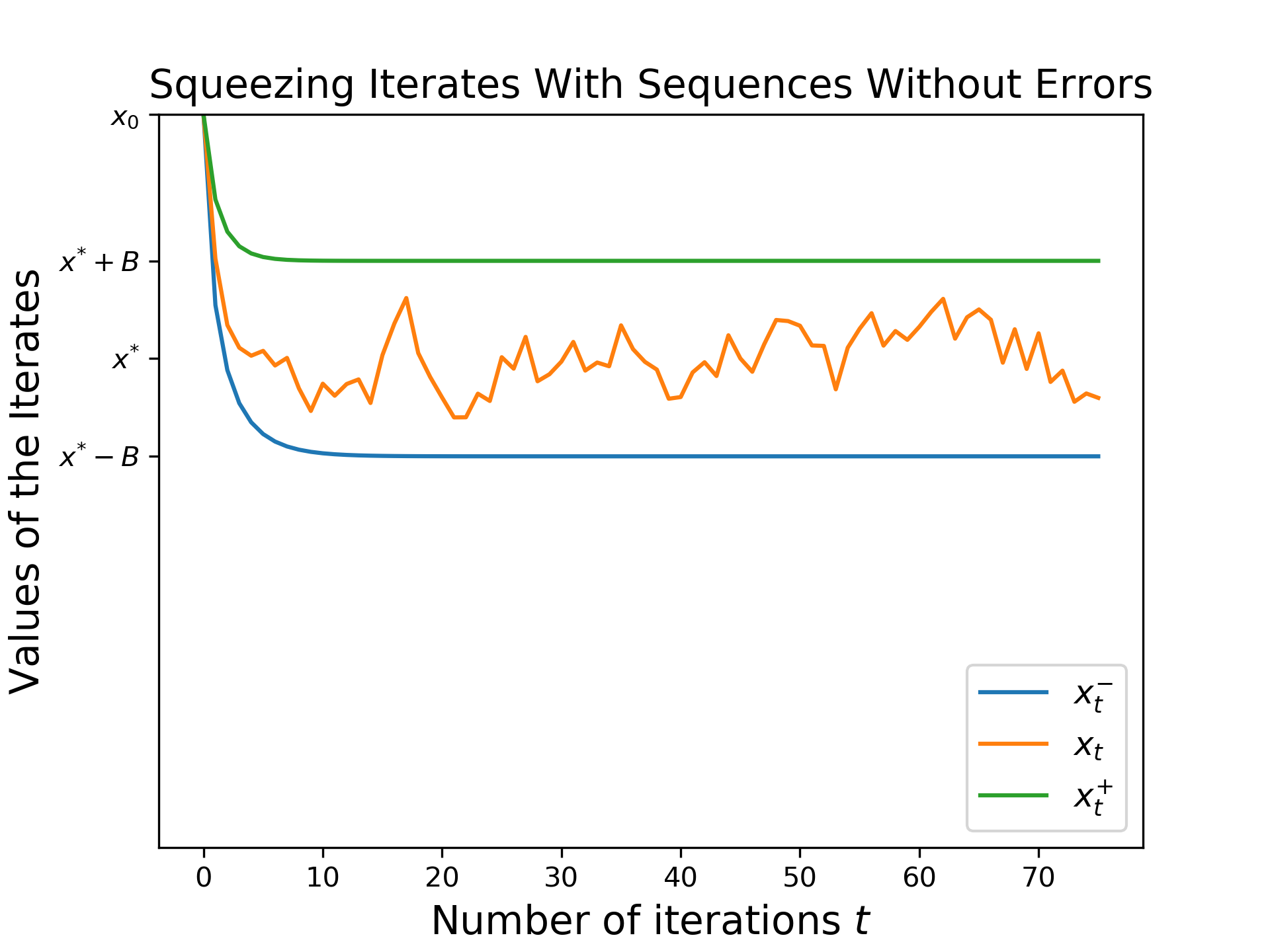}
  \end{subfigure}
  \caption{A graphical illustration of
    Lemmas~\ref{lemma:squeezing-iterates-with-bounded-errors}
    and \ref{lemma:iteratates-with-bounded-errors-monotonic-behaviour}. For a
    given error bound $B$ we have sampled error sequence $(b_{t})_{t \geq 0}$
    from $\text{Uniform}[-B, B]$ distribution. Note that for $B = 0$ the above
    plots illustrate Lemma~\ref{lemma:iterates-behave-monotonically}.}
  \label{fig:squeezed-iterates}
\end{figure}

In Section~\ref{appendix:multiplicative-updates:basic-lemmas} we analyzed one
dimensional multiplicative update sequences and proved that it is possible to
fit large enough signal while fitting a controlled amount of error.
In this section we extend
the setting considered in
Section~\ref{appendix:multiplicative-updates:basic-lemmas} to handle bounded
error sequences $(\vec{b}_{t})_{t \geq 0}$ such that for any $t \geq 0$ we have
$\norm{\vec{b}_{t}}_{\infty} \leq B$ for some $B \in \mathbb{R}$.
That is, we look at one-dimensional multiplicative sequences with positive
target $x^{\star}$ with updates given by
\begin{align}
  \label{eq:iterates-with-bounded-errors}
  x_{t+1} = x_{t}(1 - 4\eta(x_{t} - x^{\star} + b_{t}))^{2}.
\end{align}
Surely, if $B \geq x^{\star}$ one could always set $b_{t} = x^{\star}$ so that
the sequence given with the above updates equation shrinks to $0$ and
convergence to $x^{\star}$ is not possible.
Hence for a given $x^{\star}$ our lemmas below will require
$B$ to be small enough, with a particular choice
$B \leq
\frac{1}{5}x^{\star}$. For a given $B$
one can only expect the sequence $(x_{t})_{t \geq 0}$ to converge to $x^{\star}$
up to precision $B$. To see that, consider two extreme scenarios,
one where for all $t \geq 0$ we have $b_{t} = B$ and another with
$b_{t} = -B$. This gives rise the following two sequences with updates
given by
\begin{align}
  \begin{split}
    \label{eq:x_t_plus_and_x_t_minus_definitions}
    x_{t+1}^{-} &= x_{t}^{-}(1 - 4\eta(x_{t}^{-} - (x^{\star} - B)))^{2}, \\
    x_{t+1}^{+} &= x_{t}^{+}(1 - 4\eta(x_{t}^{-} - (x^{\star} + B)))^{2}.
  \end{split}
\end{align}
We can think of sequences
$(x_{t}^{-})_{t \geq 0}$
and
$(x_{t}^{+})_{t \geq 0}$
as sequences with no errors and targets $x^{\star} - B$ and $x^{\star} + B$
respectively.
We already understand the behavior of such sequences with the lemmas
derived in Section~\ref{appendix:multiplicative-updates:basic-lemmas}.
The following lemma is the key result in this section. It tells us that the
sequence $(x_{t})_{t \geq 0}$ is sandwiched between sequences
$(x_{t}^{-})_{t \geq 0}$
and
$(x_{t}^{+})_{t \geq 0}$
for all iterations $t$.
See Figure~\ref{fig:squeezed-iterates} for a graphical illustration.

\begin{lemma}[Squeezing iterates with bounded errors]
  \label{lemma:squeezing-iterates-with-bounded-errors}
  Let $(b_{t})_{t \geq 0}$ be a sequence of errors such that exists some
  $B > 0$ such that for all $t \geq 0$ we have $\abs{b_{t}} \leq B$.
  Consider the sequences
  $(x_{t}^{-})_{t \geq 0}$,
  $(x_{t})_{t \geq 0}$
  and
  $(x_{t}^{+})_{t \geq 0}$
  as defined in equations~\eqref{eq:iterates-with-bounded-errors} and
  \eqref{eq:x_t_plus_and_x_t_minus_definitions}
  with
  $$0 < x_{0}^{-} = x_{0}^{+} = x_{0} \leq x^{\star} + B$$
  If $\eta \leq \frac{1}{16 (x^{\star} + B)}$
  then for all $t \geq 0$
  $$
    0 \leq x_{t}^{-} \leq x_{t} \leq x_{t}^{+} \leq x^{\star} + B.
  $$
\end{lemma}

\begin{proof}
  We will prove the claim by induction. The claim holds trivially for $t = 0$.
  Then if $x_{t}^{+} \geq x_{t}$, denoting $\Delta \coloneqq x_{t}^{+} - x_{t}
  \geq 0$ and $m_{t} \coloneqq 1 - 4\eta(x_{t} - x^{\star} + b_{t})$ we have
  \begin{align*}
    x_{t+1}^{+}
    &= x_{t}^{+}(1 - 4\eta(x_{t}^{+} - x^{\star} - B))^{2} \\
    &= (x_{t} + \Delta)
       (1 - 4\eta(x_{t} - x^{\star} + b_{t}) - 4\eta(\Delta - B - b_{t}))^{2} \\
    &\geq (x_{t} + \Delta)
       (m_{t} - 4\eta \Delta)^{2} \\
    &= (x_{t} + \Delta)
       (m_{t}^{2} - 8\eta \Delta m_{t} + 16 \eta^{2} \Delta^{2}) \\
    &\geq (x_{t} + \Delta)
       (m_{t}^{2} - 8\eta \Delta m_{t}) \\
    &=  x_{t+1} + \Delta m_{t}^{2} - x_{t}^{+}8 \eta \Delta m_{t} \\
    &= x_{t+1} + \Delta m_{t} (m_{t} - 8 \eta x_{t}^{+}) \\
    &\geq x_{t+1},
  \end{align*}
  where the last line is true since by
  lemma~\ref{lemma:iterates-behave-monotonically} we have
  $0 < x_{t}^{+} \leq x^{\star} + B$ and so using $\eta \leq \frac{1}{16(x^{\star} +
  B)}$ we get
  \begin{align*}
    m_{t} - 8\eta x_{t}^{+}
    &\geq m_{t} - \frac{1}{2} \\
    &= \frac{1}{2} - 4\eta(x_{t} - x^{\star} + b_{t}) \\
    &\geq \frac{1}{2} - 4\eta(x^{\star} + B - x^{\star} + b_{t}) \\
    &\geq \frac{1}{2} - 8\eta B \\
    &\geq 0.
  \end{align*}
  Showing that $x_{t+1} \geq x_{t+1}^{-}$ follows a similar argument.

  Finally, as we have already pointed out $x_{t}^{+} \leq x^{\star} + B$
  holds for all $t$ by the choice of $\eta$ and
  Lemma~\ref{lemma:iterates-behave-monotonically}.
  By induction and the choice of the step size we then also have for all
  $t \geq 0$
  \begin{align*}
    x_{t+1}^{-}
    &= x_{t}^{-}(1 - 4\eta(x_{t}^{-} - x^{\star} + B))^{2} \\
    &\geq x_{t}^{-}(1 - 8\eta B)^{2} \\
    &\geq 0,
  \end{align*}
  which completes our proof.
\end{proof}

Using the above lemma we can show analogous results for iterates with bounded
errors to the ones shown in
Lemmas~\ref{lemma:iterates-behave-monotonically},
\ref{lemma:iterates-behaviour-near-convergence} and
\ref{lemma:iterates-approach-target-exponentially-fast}.

We will first prove a counterpart to
Lemma~\ref{lemma:iterates-behave-monotonically},
which is a crucial result in proving
Proposition~\ref{proposition:dealing-with-proportional-errors}. As illustrated
in Figure~\ref{fig:squeezed-iterates}, monotonicity
will hold while $\abs{x_{t} - x^{\star}} > B$. On the other hand, once
$x_{t}$ hits the $B$-tube around $x^{\star}$ it will always stay inside the tube.
This is formalized in the next lemma.

\begin{lemma}[Iterates with bounded errors monotonic behaviour]
  \label{lemma:iteratates-with-bounded-errors-monotonic-behaviour}
  Consider the setting of
  Lemma~\ref{lemma:squeezing-iterates-with-bounded-errors}
  with $B \leq \frac{1}{5} x^{\star}$, $\eta \leq \frac{5}{96x^{\star}}$
  and $0 < x_{0} \leq \frac{6}{5}x^{\star}$.
  Then the following holds:
  \begin{enumerate}
    \item If $\abs{x_{t} - x^{\star}} > B$ then
          $\abs{x_{t+1} - x^{\star}} < \abs{x_{t} - x^{\star}}$.
    \item If $\abs{x_{t} - x^{\star}} \leq B$
          then $\abs{x_{t+1} - x^{\star}} \leq B$.
  \end{enumerate}
\end{lemma}

\begin{proof}
  First, note that our choice of step size, maximum error $B$ and maximum value for
  $x_{0}$ ensures that we can apply the second part of
  Lemma~\ref{lemma:iterates-behave-monotonically} to the sequence
  $(x_{t}^{-})_{t \geq 0}$ and the first part of
  Lemma~\ref{lemma:iterates-behave-monotonically} to the sequence
  $(x_{t}^{+})_{t \geq 0}$.

  To prove the first part, note that if $0 < x_{t} < x^{\star} - B$
  then $x_{t} < x_{t+1} \leq x_{t+1}^{+} \leq x^{\star} + B$ and the result
  follows. On the other hand, if $x^{\star} + B < x_{t} \leq \frac{6}{5}x^{\star}$
  then applying Lemma~\ref{lemma:squeezing-iterates-with-bounded-errors}
  (with a slight abuse of notation, setting $x_{0} \coloneqq x_{t}$) we get
  $x^{\star} - B \leq x_{t+1}^{-} \leq x_{t+1} < x_{t}$ which finishes the
  proof of the first part.

  The second part is immediate by
  Lemma~\ref{lemma:squeezing-iterates-with-bounded-errors}
  applied again with a slight abuse of notation setting $x_{0} \coloneqq x_{t}$
  and observing that by monotonicity
  Lemma~\ref{lemma:iterates-behave-monotonically} the sequence
  $(x_{t}^{-})_{t \geq 0}$ will monotonically decrease to $x^{\star} - B$ and the
  sequence $(x_{t}^{+})_{t \geq 0}$ will monotonically increase to $x^{\star} + B$.
\end{proof}

\begin{lemma}[Iterates with bounded errors behaviour near convergence]
  \label{lemma:iterates-with-bounded-errors-behaviour-near-convergence}
  Consider the setting of
  Lemma~\ref{lemma:iteratates-with-bounded-errors-monotonic-behaviour}.
  Then the following holds:
  \begin{enumerate}
    \item If $\frac{1}{2} \left(x^{\star} - B\right) \leq x_{0} \leq x^{\star} - 5B$
          then for any $t \geq \frac{5}{8 \eta x^{\star}}$ we have
          $$
            \abs{x^{\star} - x_{t}} \leq \frac{1}{2}\abs{x_{0} - x^{\star}}.
          $$
    \item If $x^{\star} + 4B < x_{0} < \frac{6}{5}x^{\star}$ then for any
          $t \geq \frac{1}{4 \eta x^{\star}}$ we have
          $$
            \abs{x^{\star} - x_{t}} \leq \frac{1}{2}\abs{x_{0} - x^{\star}}.
          $$
  \end{enumerate}
\end{lemma}

\begin{proof}
  Let the sequences
  $(x_{t}^{+})_{t \geq 0}$
  and $(x_{t}^{-})_{t \geq 0}$
  be given as in Lemma~\ref{lemma:squeezing-iterates-with-bounded-errors}.
  For the first part, we apply
  Lemma~\ref{lemma:iterates-behaviour-near-convergence}
  to the sequence $x_{t}^{-}$ twice, to get that for all
  $$
    t \geq \frac{5}{8 \eta x^{\star}}
      \geq 2 \frac{1}{4 \eta (x^{\star} - B)}
  $$
  we have
  \begin{align*}
    0
    &\leq (x^{\star} - B) - x_{t}^{-} \\
    &\leq \frac{1}{4} \abs{x_{0} - (x^{\star} - B)} \\
    &\leq \frac{1}{4} \abs{x_{0} - x^{\star}} + \frac{1}{4}B.
  \end{align*}
  Then, if $x_{t} \leq x^{\star}$ we have by
  Lemma~\ref{lemma:squeezing-iterates-with-bounded-errors}
  and the above inequality
  \begin{align*}
    0
    &\leq x^{\star} - x_{t} \\
    &\leq x^{\star} - x_{t}^{-} \\
    &\leq \frac{1}{4} \abs{x_{0} - x^{\star}} + \frac{5}{4}B \\
    &\leq \frac{1}{2} \abs{x_{0} - x^{\star}}.
  \end{align*}
  If $x_{t} \geq x^{\star}$ then by
  lemma~\ref{lemma:squeezing-iterates-with-bounded-errors} we have
  \begin{align*}
    0 \leq x_{t} - x^{\star} \leq B \leq \frac{1}{5}\abs{x_{0} - x^{\star}},
  \end{align*}
  where the last inequality follows from $x_{0} \leq x^{*} - 5B$.
  This concludes the first part.

  The second part can be shown similarly. We apply
  lemma~\ref{lemma:iterates-behaviour-near-convergence}
  to the sequence $x_{t}^{+}$ twice, to get that for all
  $$
    t \geq 2\frac{1}{8\eta x^{\star}} \geq 2\frac{1}{8 \eta (x^{\star} + B)}
  $$
  we have
  \begin{align*}
    0
    &\leq x_{t}^{+} - (x^{\star} + B) \\
    &\leq \frac{1}{4} \abs{x_{0} - (x^{\star} + B)} \\
    &\leq \frac{1}{4} \abs{x_{0} - x^{\star}} + \frac{1}{4}B.
  \end{align*}
  Then again, if $x_{t} \geq x^{\star}$ then
  \begin{align*}
    0
    &\leq x_{t} - x^{\star} \\
    &\leq x_{t}^{+} - x^{\star} \\
    &\leq \frac{1}{4} \abs{x_{0} - x^{\star}} + \frac{5}{4}B \\
    &\leq \frac{1}{2} \abs{x_{0} - x^{\star}}
  \end{align*}
  and if $x_{t} \leq x^{\star}$ then by
  lemma~\ref{lemma:squeezing-iterates-with-bounded-errors}
  we have
  \begin{align*}
    0 \leq x^{\star} - x_{t} \leq B \leq \frac{1}{4}\abs{x_{0} - x^{\star}}
  \end{align*}
  which finishes our proof.
\end{proof}

\begin{lemma}[Iterates with bounded errors approach target exponentially fast]
  \label{lemma:iterates-with-bounded-errors-approach-target-exponentially-fast}
  Consider the setting of
  Lemma~\ref{lemma:iteratates-with-bounded-errors-monotonic-behaviour} and
  fix any $\varepsilon > 0$. Then the following holds:
  \begin{enumerate}
    \item If $B + \varepsilon < \abs{x^{\star} - x_{0}} \leq \frac{1}{5}x^{\star}$
      then for any
      $t \geq \frac{15}{32 \eta x^{\star}}
       \log \frac{\abs{x^{\star} - x_{0}}}{\varepsilon}$ iterations we have
      $\abs{x^{\star} - x_{t}} \leq B + \varepsilon$.
    \item If $0 < x_{0} \leq x^{\star} - B - \varepsilon$ then for any
      $t \geq
        \frac{15}{32 \eta x^{\star}}
        \log \frac{(x^{\star})^{2}}{x_{0} \varepsilon}$
       we have
       $x^{\star} - B - \varepsilon \leq x_{t} \leq x^{\star} + B$.
  \end{enumerate}
\end{lemma}

\begin{proof}
  \hfill
  \begin{enumerate}
    \item If $x_{0} > x^{\star} + B$ then by
      Lemmas~\ref{lemma:squeezing-iterates-with-bounded-errors} and
      \ref{lemma:iteratates-with-bounded-errors-monotonic-behaviour}
      we only need show that $(x_{t}^{+})_{t \geq 0}$
      hits $x^{\star} + B + \varepsilon$
      within the desired number of iterations.
      By the first part of
      Lemma~\ref{lemma:iterates-approach-target-exponentially-fast}
      applied to the sequence $(x_{t}^{+})_{t \geq 0}$ we see that
      $\frac{3}{8\eta(x^{\star} + B)}
        \log \frac{\abs{x_{0} - (x^{\star} +
        B)}}{\varepsilon}
        \leq \frac{15}{32 \eta x^{\star}}
       \log \frac{\abs{x^{\star} - x_{0}}}{\varepsilon}$ iterations enough.

      Similarly, if $x_{0} < x^{\star} - B$
      by the first part of
      Lemma~\ref{lemma:iterates-approach-target-exponentially-fast}
      applied to the sequence $(x_{t}^{-})_{t \geq 0}$ we see that
      $\frac{3}{8 \eta (x^{\star} - B)}
        \log \frac{\abs{x_{0} - (x^{\star} - B)}}{\varepsilon}
        \leq \frac{15}{32 \eta x^{\star}}
       \log \frac{\abs{x^{\star} - x_{0}}}{\varepsilon}$ iterations enough.

    \item
      The upper-bound is immediate from
      lemma~\ref{lemma:squeezing-iterates-with-bounded-errors}.
      To get the lower-bound we simply apply the second part of
      lemma~\ref{lemma:iterates-approach-target-exponentially-fast} to the
      sequence $(x_{t}^{-})_{t \geq 0}$ given in
      lemma~\ref{lemma:squeezing-iterates-with-bounded-errors} to get that
      for any
      \begin{align*}
        t
        \geq \frac{3}{8 \eta \frac{4}{5} x^{\star}}
             \log \frac{(x^{\star})^{2}}{x_{0} \varepsilon}
        \geq \frac{3}{8 \eta (x^{\star} - B)}
             \log \frac{(x^{\star} - B)^{2}}{x_{0} \varepsilon}
      \end{align*}
      we have
      $x^{\star} - B - \varepsilon \leq x_{t}^{-} \leq x_{t}$
      which is what we wanted to show.
  \end{enumerate}
\end{proof}

%% file: appendix/multiplicative_updates/proportional_errors.tex
\subsection{Dealing With Errors Proportional to Convergence Distance}
\label{appendix:multiplicative-updates:dealing-with-rip-errors}

In this section we derive lemmas helping to deal with errors proportional to
convergence distance, that is, the error sequence $(\vec{p}_{t})_{t \geq 0}$
given in equation~\eqref{eq:error-decompositions} in
Appendix~\ref{appendix:main-proofs:set-up-and-intuition}.
Note that we cannot simply upper-bound $\norm{\vec{b}_{t}}_{\infty} +
\norm{\vec{p}_{t}}_{\infty}$ by some large number independent of $t$
and treat both errors together as a bounded error
sequence since $\norm{\vec{p}_{0}}_{\infty}$ can be much larger than some of the
coordinates of $\wstar$.
On the other hand, by
Sections~\ref{appendix:multiplicative-updates:basic-lemmas} and
\ref{appendix:multiplicative-updates:bounded-errors} we expect
$\norm{\vec{s}_{t} - \vec{w}^{\star}}_{\infty}$ to decay exponentially fast
and hence the
error $\norm{\vec{p}_{t}}_{\infty}$ should also decay exponentially fast.

Let $m$ and $T_{0}, \dots, T_{m-1}$ be some integers and suppose that we run
gradient descent for $\sum_{i=0}^{m-1} T_{i}$ iterations.
Suppose that for each time interval
$\sum_{i=0}^{j-1} T_{i} \leq t \leq \sum_{i=0}^{j} T_{i}$ we can upper-bound
$\norm{\vec{b}_{t}}_{\infty} + \norm{\vec{p}_{t}}_{\infty}$ by $2^{-j}B$
for some $B \in \mathbb{R}$.
The following lemma then shows how to control errors of such type and it is,
in fact, the reason why in the main theorems a logarithmic term appears
in the upper-bounds for the RIP parameter $\delta$.
We once again restrict ourselves to one-dimensional sequences.

\begin{lemma}[Halving errors over doubling time intervals]
  \label{lemma:halving-errors-doubling-intervals}
  Let $T > 0$ be some fixed positive real number,
  $T_{i} \coloneqq 2^{i} T$
  and $\bar{T_{i}} \coloneqq \sum_{j=0}^{i} T_{j}$.
  Further, suppose $(p_{t})_{t \geq 0}$ is a sequence of
  real numbers and let $B \in \mathbb{R}$.
  Suppose that for every integer $i \geq 0$ and for any
  $\bar{T}_{i-1} \leq t < \bar{T}_{i}$
  we have $\abs{p_{t}} \leq 2^{-i}B$.
  Then, for any integer $i \geq 0$ and $\eta \leq \frac{1}{4B}$
  $$
    \prod_{i = 0}^{\bar{T}_{i}-1} (1 + 4\eta p_{t})^{2}
    \leq
    (1 + 4\eta 2^{-i}B)^{2(i + 1) T_{i}}.
  $$
\end{lemma}

\begin{proof}
  Note that for $x, y \geq 0$ we have $(1 + x + y) \leq (1 + x)(1 + y)$ and
  in particular, for any integers $i \geq j \geq 0$
  \begin{align*}
    1 + 4\eta 2^{-j}B
    \leq
    (1 + 4\eta 2^{-j-1}B)^{2}
    \leq
    \dots
    \leq
    (1 + 4\eta 2^{-i}B)^{2^{i-j}}.
  \end{align*}
  It follows that
  \begin{align*}
    \prod_{t = 0}^{\bar{T}_{i} - 1} (1 + 4\eta p_{t})^{2}
    &\leq \prod_{j=0}^{i} (1 + 4\eta 2^{-j}B)^{2T_{j}} \\
    &\leq \prod_{j=0}^{i} (1 + 4\eta 2^{-i}B)^{2^{i-j}2T_{j}} \\
    &= (1 + 4\eta 2^{-i}B)^{2(i+1)T_{i}}.
  \end{align*}
\end{proof}

Sometimes $\norm{\vec{p}_{t}}_{\infty}$ can be much larger
than some coordinates of the true parameter vector $\vec{w}^{\star}$.
For example, if $\wmax \gg \wmin$ then $\norm{\vec{p}_{0}}_{\infty}$
can be much larger than $\wmin$.
In Section~\ref{appendix:multiplicative-updates:bounded-errors} we have shown
how to deal with bounded errors that are much smaller than target.
We now show how to deal with errors much larger than the target.

\begin{lemma}[Handling large errors]
  \label{lemma:handling-large-errors}
  Let $(b_{t})_{t \geq 0}$ be a sequence of errors such that for some $B \in
  \mathbb{R}$ and all $t \geq 0$ we have $\abs{b_{t}} \leq B$.
  Consider a sequence defined as
  \begin{align*}
    x^{\star} + 2B &\leq x_{0} \leq x^{\star} + 4B, \\
    x_{t+1} &= x_{t}(1 - 4\eta(x_{t} - x^{\star} + b_{t}))^{2}.
  \end{align*}
  Then, for $\eta \leq \frac{1}{20B}$ and any $t \geq \frac{1}{10 \eta B}$
  we have
  $$
    0 \leq x_{t} \leq x^{\star} + 2B.
  $$
\end{lemma}

\begin{proof}
  Note that if $x_{t} \geq x^{\star} + 2B$ then
  \begin{align*}
    x_{t+1}
    &= x_{t}(1 - 4\eta(x_{t} - x^{\star} + b_{t}))^{2} \\
    &\leq x_{t}(1 - 4\eta B)^{2}.
  \end{align*}
  Hence to find $t$ such that $x_{t} \leq x^{\star} + 2B$
  it is enough to satisfy the following inequality
  \begin{align*}
    &(x^{\star} + 4B)(1 - 4\eta B)^{2t} \leq x^{\star} + 2B \\
    \iff&
    t \geq
      \frac{1}{2}
      \frac{1}{\log (1 - 4\eta B)}
      \log \frac{x^{\star} + 2B}{x^{\star} + 4B}
  \end{align*}
  Since for any $x \in (0, 1)$ we have $\log (1 - x) \leq -x$
  hence $\log(1 - 4\eta B) \leq - 4 \eta B$.
  Also, since $\frac{x^{\star} + 2B}{x^{\star} + 4B} \geq \frac{1}{2}$
  we have $\log \frac{x^{\star} + 2B}{x^{\star} + 4B} \geq \log \frac{1}{2} \geq
  -\frac{7}{10}$. Hence
  $$
      \frac{1}{2}
      \frac{1}{\log (1 - 4\eta B)}
      \log \frac{x^{\star} + 2B}{x^{\star} + 4B}
      \leq
      \frac{1}{2}
      \cdot \frac{1}{-4\eta B}
      \cdot \frac{-7}{10}.
  $$
  Setting $t \geq \frac{1}{10 \eta B}$ is hence enough.
  To ensure non-negativity of the iterates, note that
  $$
    \abs{4\eta(x_{t} - x^{\star} + b_{t})}
    \leq 20 \eta B
  $$
  and hence setting $\eta \leq \frac{1}{20 B}$ is enough.
\end{proof}

The final challenge caused by the error sequence $(\vec{p}_{t})_{t \geq 0}$ is that
some of the signal components $\id{S} \odot \vec{w}_{t}$ can actually shrink
initially instead of approaching the target.
Hence for all $t \geq 0$ we need to control the maximum shrinkage by bounding
the following term from below
\begin{equation}
  \label{eq:signal-shrinkage-term}
  \alpha^{2} \prod_{i = 0}^{t-1} (1 - 4\eta(\norm{\vec{b}_{t}}_{\infty} +
  \norm{\vec{p}_{t}}_{\infty}))^{2}.
\end{equation}
Recall that we are handling maximum growth of the error sequence
$(\vec{e}_{t})_{t \geq 0}$ by Lemma~\ref{lemma:controlling-errors} which requires
upper-bounding the term
\begin{equation}
  \label{eq:error-controlling-term}
  \alpha^{2} \prod_{i = 0}^{t-1} (1 + 4\eta(\norm{\vec{b}_{t}}_{\infty} +
  \norm{\vec{p}_{t}}_{\infty}))^{2}.
\end{equation}
If the term in equation~\eqref{eq:error-controlling-term} is not too large, then
we can prove that the term in equation~\eqref{eq:signal-shrinkage-term} cannot
be too small. This idea is exploited in the following lemma.

\begin{lemma}[Handling signal shrinkage]
  \label{lemma:handling-signal-shrinkage}
  Consider a sequence
  \begin{align*}
    x_{0} &= \alpha^{2}, \\
    x_{t+1} &= x_{t}(1 - 4\eta(x^{\star} + b_{t} + p_{t}))^{2}
  \end{align*}
  where $x^{\star} > 0$ and exists some $B > 0$ such that for all $t \geq 0$
  we have $\abs{b_{t}} + \abs{p_{t}} \leq B$.
  If $\eta \leq \frac{1}{8B}$ and
  $$
    \prod_{i = 0}^{t-1}
    (1 + 8\eta(\abs{b_{t}} + \abs{p_{t}}))^{2}
    \leq \frac{1}{\alpha}
  $$
  then
  $$
    \prod_{i = 0}^{t-1}
    (1 - 4\eta(\abs{b_{t}} + \abs{p_{t}}))^{2}
    \geq \alpha.
  $$
\end{lemma}

\begin{proof}
  By the choice of step size $\eta$ we always have
  $0 \leq 4\eta(\abs{b_{t}} + \abs{p_{t}}) \leq
  \frac{1}{2}$.
  Since for $x \in [0, \frac{1}{2}]$ we have
  $(1 + 2x)(1 - x) = 1 + x - 2x^{2} \geq 1$
  it follows that
  $$
    \prod_{i = 0}^{t-1}
    (1 + 8\eta(\abs{b_{t}} + \abs{p_{t}}))^{2}
    \prod_{i = 0}^{t-1}
    (1 - 4\eta(\abs{b_{t}} + \abs{p_{t}}))^{2}
    \geq 1
  $$
  and we are done.
\end{proof}

%% file: appendix/multiplicative_updates/negative_targets.tex
\subsection{Dealing With Negative Targets}
\label{appendix:multiplicative-updates:negative-targets}

So far we have only dealt with sequences
converging to some positive target, i.e., the parametrization
$\vec{w}_{t} = \vec{u}_{t} \odot \vec{u}_{t}$.
In this section
we show that handling parametrization $\vec{w}_{t} = \vec{u}_{t} \odot
\vec{u}_{t} - \vec{v}_{t} \odot
\vec{v}_{t}$ can be done by noting that for any coordinate $i$, at least one of
$u_{t,i}$ or $v_{t,i}$ has to be close to its initialization value.
Intuitively, this observation will allow us to treat parametrization
$\vec{w}_{t} = \vec{u}_{t} \odot \vec{u}_{t} - \vec{v}_{t} \odot \vec{v}_{t}$
as if it was
$\vec{w}_{t} \approx \vec{u}_{t} \odot \vec{u}_{t}$ and all coordinates of the target
$\wstar$ are replaced by its absolute values.

Consider two sequences given by
\begin{align*}
  0 < x^{+}_{0} &= \alpha^{2} \leq x^{\star}_{+},\quad
  x^{+}_{t+1} = x^{+}_{t}(1 - 4\eta(x^{+}_{t} - x^{\star}_{+} + b_{t}))^{2} \\
  0 < x^{-}_{0} &= \alpha^{2} \leq -x^{\star}_{-}, \quad
  x_{t+1}^{-} = x^{-}_{t}(1 + 4\eta(-x^{-}_{t} - x_{-}^{\star} + b_{t}))^{2}
\end{align*}
where $(b_{t})_{t \geq 0}$ is some sequence of errors and the targets satisfy
$x_{+}^{\star} > 0$ and $x_{-}^{\star} < 0$. We already know how to deal with the
sequence $(x^{+}_{t})_{t \geq 0}$.
Note that we can rewrite the updates for the sequence
$(x^{-}_{t})_{t \geq 0}$ as follows
$$
  x^{-}_{t+1} = x^{-}_{t}(1 - 4\eta(x^{-}_{t} - \abs{x_{-}^{\star}} - b_{t}))^{2}.
$$
and we know how to deal with sequences of this form. In particular,
$(x^{-}_{t})_{t \geq 0}$ will converge to $\abs{x_{-}^{\star}}$ with error
at most $B$ equal to
some bound on maximum error and hence the sequence
$(-x^{-}_{t})_{t \geq 0}$ will converge to a $B$-tube around $x^{\star}$.
Hence, our theory developed for sequences with
positive targets directly apply for sequences with negative targets of the
form given above.

The following lemma is the key result allowing to treat
$\vec{w}_{t} = \vec{u}_{t} \odot \vec{u}_{t} - \vec{v}_{t} \odot \vec{v}_{t}$
almost as if it was
$\vec{w}_{t} \approx \vec{u}_{t} \odot \vec{u}_{t}$ as discussed at the
beginning of this section.

\begin{lemma}[Handling positive and negative sequences simultaneously]
  \label{lemma:handling-negative-and-positive-target-simultaneously-helper}
  Let $x_{t} = x_{t}^{+} - x^{-}_{t}$ and $x^{\star} \in \mathbb{R}$ be the target such
  that $\abs{x^{\star}} > 0$.
  Suppose the sequences $(x^{+}_{t})_{t \geq 0}$ and
  $(x^{-}_{t})_{t \geq 0}$ evolve
  as follows
  \begin{align*}
    0 < x^{+}_{0} = \alpha^{2} \leq \frac{1}{4}\abs{x^{\star}},\quad
    x^{+}_{t+1} = x^{+}_{t}(1 - 4\eta(x_{t} - x^{\star} + b_{t}))^{2} \\
    0 < x^{-}_{0} = \alpha^{2} \leq \frac{1}{4}\abs{x^{\star}}, \quad
    x^{-}_{t+1} = x^{-}_{t}(1 + 4\eta(x_{t} - x^{\star} + b_{t}))^{2}.
  \end{align*}
  and that there exists $B > 0$ such that $\abs{b_{t}} \leq B$ and
  $\eta \leq \frac{1}{12(x^{\star} + B)}$.
  Then the following holds:
  \begin{enumerate}
    \item For any $t \geq 0$ we have
      $0 \leq x^{+}_{t} \wedge x^{-}_{t} \leq \alpha^{2}$.
    \item For any $t \geq 0$ we have
      \begin{itemize}
        \item If $x^{\star} > 0$ then $x^{-}_{t} \leq \alpha^{2} \prod_{i = 0}^{t-1}(1
          + 4\eta \abs{b_{t}})$.
        \item If $x^{\star} < 0$ then $x^{+}_{t} \leq \alpha^{2} \prod_{i = 0}^{t-1}(1
          + 4\eta \abs{b_{t}})$.
      \end{itemize}
  \end{enumerate}
\end{lemma}

\begin{proof}
  The choice of our step size ensures that
  $\abs{4\eta(x_{t} - x^{\star} + b_{t})} \leq \frac{1}{2}$.
  For any $0 \leq a \leq \frac{1}{2}$ we have
  $0 \leq (1 - a)(1 + a) = 1 - a^{2} \leq 1$.
  In particular, this yields for any $t \geq 0$
  $$
    x^{+}_{t}x^{-}_{t}
    =
    \alpha^{4} \prod_{i = 0}^{t-1}
    (1 - 4\eta(x_{t} - x^{\star} + b_{t}))^{2}
    (1 + 4\eta(x_{t} - x^{\star} + b_{t}))^{2}
    \leq \alpha^{4}
  $$
  which concludes the first part.

  To prove the second part assume $x^{\star} > 0$ and fix any $t \geq 0$.
  Let $0 \leq s \leq t$ be the largest $s$ such that $x^{+}_{s} > x^{\star}$.
  If no such $s$ exists we are done immediately.
  If $s = t$ then by the first part we have $x^{-}_{t} \leq \alpha^{2}$
  and we are done.

  If $s < t$ then we have by the first part
  and by the assumption $\alpha^{2} \leq \frac{1}{4}\abs{x^{\star}}$,\,
  $x^{-}_{s} \leq
  \frac{\alpha^{4}}{x^{+}_{s}} \leq \frac{1}{4} \alpha^{2}$.
  Further, by the choice of step size $\eta$ we have
  $x^{+}_{s} \leq 4x^{\star}$.
  It then follows that
  $$
    (1 + 4\eta(x_{s} - x^{\star} + b_{t}))^{2} \leq 4
  $$
  and hence
  \begin{align*}
    x^{-}_{t}
    &= x^{-}_{s}
      \prod_{i=s}^{t-1}
      (1 + 4\eta(x^{+}_{i} - x^{-}_{i} - x^{\star} + b_{i}))^{2} \\
    &\leq \frac{1}{4}\alpha^{2}
      (1 + 4\eta(x_{s} - x^{\star} + b_{t}))^{2}
      \prod_{i=s+1}^{t-1}
      (1 + 4\eta(x^{+}_{i} - x^{-}_{i} - x^{\star} + b_{i}))^{2} \\
    &\leq \alpha^{2}
      \prod_{i=s+1}^{t-1}
      (1 + 4\eta\abs{b_{t}}))^{2}.
  \end{align*}
  This completes our proof for the case $x^{\star} > 0$.
  For $x^{\star} < 0$ we are done by symmetry.
\end{proof}

%% file: appendix/multiplicative_updates/proof_of_key_proposition.tex
\subsection{Proof of
Proposition~\ref{proposition:dealing-with-proportional-errors}}
\label{appendix:multiplicative-updates:proof-of-key-proposition}

In this section we will prove
Proposition~\ref{proposition:dealing-with-proportional-errors}.
We remind our readers, that the goal of this proposition is showing that
the error sequence $(\vec{p}_{t})_{t \geq 0}$ can be essentially ignored if the
RIP constant $\delta$ is small enough.

Recall that the error arising due to the bounded error sequence
$(\vec{b}_{t})_{t \geq 0}$
is irreducible
as discussed in Section~\ref{appendix:multiplicative-updates:bounded-errors}.
More formally, we will show that if for some
$0 \leq \zeta \leq \wmax$
we have $\norm{\vec{b}_{t}}_{\infty} \lesssim \zeta$
and if
$\norm{\vec{p}_{t}}_{\infty} \lesssim \frac{1}{\log_{2}
  \frac{\wmax}{\zeta}}\norm{s_{t} - \wstar}_{\infty}$
then after $t = O\left(\frac{1}{\eta \zeta} \log \frac{1}{\alpha} \right)$
iterations we have $\norm{\vec{s}_{t} - \wstar}_{\infty} \leq \zeta$.
In particular, up to absolute multiplicative constants we perform as good as
if the error sequence $(\vec{p}_{t})_{t \geq 0}$ was equal to $0$.

The proof idea is simple, but the details can complicated.
We will first prove a counterpart to
Proposition~\ref{proposition:dealing-with-proportional-errors}
which will correspond to parametrization $\vec{w}_{t} = \vec{u}_{t} \odot
\vec{u}_{t}$, that is,
we will only try to fit the positive coordinates of $\wstar$.
We will later use
Lemma~\ref{lemma:handling-negative-and-positive-target-simultaneously-helper}
to extend our result to the general case.
We now list the key ideas appearing in the proof below.
\begin{enumerate}
  \item Initially we have $\norm{\vec{w}_{0} - \wstar}_{\infty} \leq \wmax$. We will
    prove our claim by induction, reducing the above distance by half during
    each induction hypothesis. We will hence need to apply
    $m \coloneqq \ceil{\log_{2} \frac{\wmax}{\zeta}}$ induction steps which
    we will enumerated from $0$ to $m-1$.
  \item At the beginning of the $i\th$ induction step we will have
    $\norm{\vec{w}_{t} - \wstar}_{\infty} \leq 2^{-i}\wmax$.
    Choosing small enough absolute constants for upper-bounds on
    error sequences
    $(\vec{b}_{t})_{t \geq 0}$ and $(\vec{p}_{t})_{t \geq 0}$ we can show that
    $$
      \norm{\vec{b}_{t}}_{\infty} + \norm{\vec{p}_{t}}_{\infty}
      \leq \frac{1}{40} 2^{-i} \wmax \eqqcolon B_{i}.
    $$
    In particular, during the $i\th$ induction step we treat both types of
    errors simultaneously as a bounded error sequence with bound $B_{i}$.
    Since at each induction step $\norm{\vec{w}_{t} - \wstar}_{\infty}$
    decreases by half, the error bound $B_{i}$ also halves.
    This puts us in position to apply
    Lemma~\ref{lemma:halving-errors-doubling-intervals} which plays a key
    role in the proof below.
  \item One technical difficulty is that in
    Section~\ref{appendix:multiplicative-updates:bounded-errors}
    all lemmas require that iterates never exceed the target by more than
    a factor $\frac{6}{5}$. We cannot ensure that since initially our errors
    can be much larger than some of the true parameter $\wstar$ coordinates.
    We instead use Lemma~\ref{lemma:handling-large-errors} to show that for
    any coordinate $j$ we have $w_{t, j} \leq w_{j}^{\star} + 4B_{i}$
    during $i\th$ induction step.
    Then for any $j$ such that $w_{j}^{\star} \geq 20B_{i}$ we can apply the
    results from
    Section~\ref{appendix:multiplicative-updates:bounded-errors}.
    On the other hand, if $w_{j}^{\star} \leq 20B_{i} = \frac{1}{2} 2^{-i}
    \wmax$ then we already have $\abs{w_{t, j} - w^{\star}_{j}} \leq 2^{-i-1}\wmax$
    and the above bound does not change during the $i\th$ induction step.
  \item During the $i\th$ induction step, if $\abs{w_{t,j} - w^{\star}_{j}}
    > 2^{-i-1}\wmax$ then $w_{j}^{\star} \geq 20B_{i}$ and we can apply
    Lemma~\ref{lemma:iteratates-with-bounded-errors-monotonic-behaviour}
    which says that all such coordinates will monotonically approach
    $B$-tube around $w^{\star}_{j}$.
    Lemma~\ref{lemma:iterates-with-bounded-errors-approach-target-exponentially-fast}
    then tells us how many iterations need to be taken for our iterates to
    get close enough to this $B$-tube so that
    $\abs{w_{t,j} - w^{\star}_{j}} \leq 2^{-i-1}\wmax$.
  \item Finally, we control the total accumulation of errors
    $\prod_{i=0}^{t-1} (1 + 4\eta(\norm{\vec{b}_{i}}_{\infty}
    + \norm{\vec{p}_{i}}_{\infty}))^{2}$ using
    Lemma~\ref{lemma:halving-errors-doubling-intervals}
    and ensure that for any $w_{j}^{\star} \geq 0$ the iterates never get
    below $\alpha^{3}$ by applying Lemma~\ref{lemma:handling-signal-shrinkage}.
\end{enumerate}

\begin{lemma}[Dealing with errors proportional to convergence distance]
  \label{lemma:dealing-with-errors-proportional-to-convergence-distance}
  Fix any $0 < \zeta \leq \wmax$ and let
  $\gamma = \frac{C_{\gamma}}{
    \lceil
      \log_{2} \frac{\wmax}{\zeta}
    \rceil }$
  where $C_{\gamma}$ is some small enough absolute constant.
  Let $\wstar \in \mathbb{R}^{k}$ be a target vector which is now allowed to
  have negative components. Denote by
  $\vec{w}^{\star}_{+}$ the positive part of $\wstar$, that is,
  $(w^{\star}_{+})_{i} = \mathbb{1}_{\{w^{\star}_{i} \geq 0\}} w^{\star}_{i}$.
  Let $(\vec{b}_{t})_{t \geq 0}$ and $(\vec{p}_{t})_{t \geq 0}$ sequences of errors such
  that for all $t \geq 0$ we have $\norm{\vec{b}_{t}}_{\infty} \leq C_{b} \zeta$
  for some small enough absolute constant $C_{b}$
  and $\norm{\vec{p}_{t}}_{\infty} \leq \gamma
  \norm{\vec{w}_{t} - \vec{w}^{\star}_{+}}_{\infty}$. Let the updates be given by
  $$
    w_{0, j} = \alpha^{2}, \quad
    w_{t+1, j} =
      w_{t, j}(1 - 4\eta(w_{t, j} - w^{\star}_{j} + b_{t, j} + p_{t, j}))^{2}.
  $$
  If
  the step size satisfies
  $
    \eta \leq \frac{5}{96\wmax}
  $
  and the initialization satisfies
  $
    \alpha \leq \frac{\zeta}{3(\wmax)^{2}}
        \wedge \sqrt{\wmin}
        \wedge 1
  $
  then for
  $
    t
    = O\left(\frac{1}{\eta \zeta} \log \frac{1}{\alpha}\right)
  $
  we have
  \begin{align*}
    \norm{\vec{w}_{t} - \vec{w}^{\star}_{+}}_{\infty}
    &\leq \zeta \\
    \alpha^{2}\prod_{i=0}^{t-1}(1 + 4\eta(
      \norm{\vec{b}_{t}}_{\infty} + \norm{\vec{p}_{t}}_{\infty}))^{2}
    &\leq \alpha.
  \end{align*}
\end{lemma}

\begin{proof}
  Let $T \coloneqq \frac{1}{\eta \wmax}
          \log \frac{1}
                    {\alpha^{4}}$
  and for any integer $i \geq -1$ let $T_{i} \coloneqq 2^{i} T$ and
  $\bar{T}_{i} \coloneqq \sum_{j=0}^{i} T_{j}$.
  We also let $\bar{T}_{-1} = 0$. Let $B_{i} \coloneqq \frac{1}{40}2^{-i}\wmax$.
  Let $m = \ceil{\log_{2} \frac{\wmax}{\zeta}}$
  so that $\gamma = \frac{C_{\gamma}}{m}$.
  We will prove our claim by induction on $i = 0, 1, \dots, m-1$.

  \underline{\textit{Induction hypothesis for $i \in \{0, \dots, m\}$ }} \\
  \begin{enumerate}
    \item For any $j < i$ and $\bar{T}_{j-1} \leq t < \bar{T}_{j}$
      we have $\norm{\vec{w}_{t} - \vec{w}^{\star}_{+}}_{\infty} \leq 2^{-j} \wmax$.
      In particular, this induction hypothesis says that we halve the
      convergence distance during each induction step.
    \item We have $\norm{\vec{w}_{\bar{T}_{i-1}} - \vec{w}^{\star}_{+}}_{\infty} \leq 2^{-i}
      \wmax$. This hypothesis controls the convergence distance at the
      beginning of the $i\th$ induction step.
    \item
      For any $j$
      we have $\alpha^{3} \leq w_{\bar{T}_{i-1}, j} \leq w_{j}^{\star} + 4B_{i}$.
  \end{enumerate}
  \underline{\textit{Base case}} \\
  For $i = 0$ all conditions hold since for all $j$ we have
  $0 \leq \alpha^{2} = w_{0, j} < w^{\star}_{j}$.

  \underline{\textit{Induction step}} \\
  Assume that the induction hypothesis holds for some $0 \leq i < m$.
  We will show that it holds for $i + 1$.
  \begin{enumerate}
    \item
    We want to show that for all
    $t \in \{0, \dots, T_{i} - 1\}$
    $\norm{\vec{w}_{\bar{T}_{i-1} + t} - \vec{w}^{\star}_{+}}_{\infty}$
    remains upper-bounded by $2^{-i}\wmax$.

    Note that
    $2^{-i}\wmax \geq 2^{-m}\wmax \geq
    \frac{1}{2}\zeta$ and hence requiring
    $C_{\gamma} + 2C_{b} \leq \frac{1}{40}$ we have
    \begin{align*}
      \norm{\vec{b}_{\bar{T}_{i-1}}}_{\infty}
      + \norm{\vec{p}_{\bar{T}_{i-1}}}_{\infty}
      &\leq C_{b}\zeta + \gamma 2^{-i}\wmax \\
      &\leq (C_{\gamma} + 2C_{b})2^{-i}\wmax \\
      &\leq \frac{1}{40}2^{-i}\wmax \\
      &= B_{i}.
    \end{align*}

    For any $j$ such that $w_{j}^{\star} \geq 20B_{i}$ the third induction
    hypothesis $w_{\bar{T}_{i-1}, j} \leq w_{j}^{\star} + 4B_{i}$ ensures that
    $w_{\bar{T}_{i-1}, j} \leq \frac{6}{5} w_{j}^{\star}$. Hence, it satisfies
    the pre-conditions of
    Lemma~\ref{lemma:iteratates-with-bounded-errors-monotonic-behaviour} and
    as long as
    $$
      \norm{\vec{w}_{\bar{T}_{i-1} + t} - \vec{w}^{\star}_{+}}_{\infty} \leq 2^{-i}\wmax
    $$
    any such $j$ will monotonically approach the $\frac{1}{40} B_{i}$-tube
    around $w_{j}^{\star}$ maintaining $\abs{w_{t} - w_{j}^{\star}} \leq 2^{-i}\wmax$.

    On the other hand, for any $j$ such that $w_{j}^{\star} \leq 20 B_{i}$
    $w_{t,j}$ will stay in $(0, w_{j}^{\star} + 4B_{i}]$ maintaining
    $\abs{w_{t} - w_{j}^{\star}} \leq 20B_{i} \leq 2^{-i}\wmax$ as required.

    By induction on $t$, we then have for any $t \geq 0$
    $$
      \norm{\vec{w}_{\bar{T}_{i-1} + t} - \vec{w}^{\star}_{+}}_{\infty}
      \leq
      2^{-i}\wmax
    $$
    which is what we wanted to show.
    \item
    To prove the second part of the induction hypothesis, we need to show that
    after $T_{i}$ iterations the maximum convergence distance
    $\norm{\vec{w}_{\bar{T}_{i}} - \vec{w}^{\star}_{+}}_{\infty}$
    decreases at
    least by half.

    Take any $j$ such that $w^{\star}_{j} \geq 0$ and
    $\abs{w_{\bar{T}_{i-1}, j}^{\star} - w^{\star}_{j}} \leq 2^{-i-1}\wmax = 20B_{i}$.
    Then by a similar argument used in to prove the first induction hypothesis
    for any $t \geq 0$ we have
    $\abs{w_{\bar{T}_{i-1} + t, j}^{\star} - w^{\star}_{j}} \leq 2^{-i-1}\wmax$ and
    hence such coordinates can be ignored.

    Now take any $j$ such that $w^{\star}_{j} \geq 0$ and
    $\abs{w_{\bar{T}_{i-1}, j}^{\star} - w^{\star}_{j}} > 2^{-i-1}\wmax$.
    Then, since $20B_{i} = 2^{-i-1}\wmax$ and since by the third induction
    hypothesis $w_{\bar{T}_{i-1}, j} \leq w^{\star}_{j} + 4B_{i}$ it follows that
    $0 \leq w_{\bar{T}_{i-1}, j} < w^{\star}_{j} - 20B_{i}$.
    Applying the second part of
    Lemma~\ref{lemma:iterates-with-bounded-errors-approach-target-exponentially-fast}
    with $\varepsilon = 19B_{i}$ and noting that
    $$
      19B_{i} = \frac{19}{40}2^{-i}\wmax \geq \frac{19}{40}2^{-m+1}\wmax
      \geq \frac{19}{40}\zeta \geq \frac{1}{3} \zeta
    $$
    we have for any
    \begin{align*}
      t
      &\geq T_{i} \\
      &\geq
        2^{i} \frac{1}{\eta \wmax}
        \log \frac{3(\wmax)^{2}}{\alpha^{3} \zeta} \\
      &\geq
        \frac{15}{32 \eta w_{j}^{\star}}
        \log \frac{(w_{j}^{\star})^{2}}{w_{\bar{T}_{i-1}, j} \cdot 19 B_{i}}
    \end{align*}
    iterations the following holds
    $$
      \abs{w_{\bar{T}_{i-1} + t, j} - w^{\star}_{j}} \leq
      20B_{i} \leq 2^{-i-1}\wmax
    $$
    which completes our proof.
    \item The upper bound follows immediately from
      Lemma~\ref{lemma:handling-large-errors} which tells that after
      \begin{align*}
        t
        \geq T_{i}
        \geq 2^{i} \frac{4}{\eta \wmax}
        = \frac{1}{10 \eta B_{i}}.
      \end{align*}
      iterations for any $j$ we have
      $w_{\bar{T}_{i-1} + t,j} \leq w^{\star}_{j} + 2B_{i} = w^{\star}_{j} + 4B_{i+1}$.

      To prove the lower-bound, first note that
      \begin{align}
        \notag
        &\prod_{i=0}^{\bar{T}_{i} - 1}
        (1 + 8\eta(\norm{\vec{b}_{i}}_{\infty} + \norm{\vec{p}_{i}}_{\infty}))^{2} \\
        &\leq
        \label{line:split-errors}
        \prod_{i=0}^{\bar{T}_{i} - 1}
        (1 + 8\eta C_{b}\zeta)^{2}
        (1 + 4\eta \norm{\vec{p}_{i}}_{\infty}))^{4} \\
        &\leq
        \label{line:deal-with-decaying-errors}
        (1 + 8\eta C_{b}\zeta)^{4 T_{i}}
        \left(
          1 + 4\eta \cdot \frac{C_{\gamma}}{m}2^{-i}\wmax
        \right)^{4(i+1) T_{i}} \\
        &\leq
        \label{line:Ti-to-Tm}
        (1 + 8\eta C_{b}\zeta)^{4 T_{m-1}}
        \left(
          1 + 4\eta \cdot \frac{C_{\gamma}}{m}2^{-m+1}\wmax
        \right)^{4m T_{m-1}} \\
        &\leq
        \label{line:taking-out-m}
        \left(
          1 + 4\eta \cdot \frac{1}{m} 2C_{b}\zeta
        \right)^{4 m T_{m-1}}
        \left(
          1 + 4\eta \cdot \frac{C_{\gamma}}{m}2^{-m+1}\wmax
        \right)^{4m T_{m-1}} \\
        &\leq
        \label{line:collecting-errors}
        \left(
          1 + 4\eta \cdot \frac{C_{\gamma}}{m} 2^{-m+1}\wmax
        \right)^{8m T_{m-1}} \\
        &\leq
        \label{line:selective-fitting-lemma-application}
        \frac{1}{\alpha}
      \end{align}
    where line~\ref{line:split-errors} follows by noting that for any
    $x, y \geq 0$ we have $(1 + x + y) \leq (1 + x)(1 + y)$.
    Line~\ref{line:deal-with-decaying-errors} follows by applying
    Lemma~\ref{lemma:halving-errors-doubling-intervals} and noting that
    $\bar{T}_{i} \leq 2T_{i}$.
    Line~\ref{line:Ti-to-Tm} follows by noting that $i \leq m-1$.
    Line~\ref{line:taking-out-m} follows by applying $(1 + mx) \leq (1 + x)^{m}$
    for $x \geq 0$ and $m \geq 1$.
    Line~\ref{line:collecting-errors} follows by noting that
    $\zeta \leq 2^{-m+1}\wmax$ and assuming that $2C_{b} \leq C_{\gamma}$.
    Line~\ref{line:selective-fitting-lemma-application} follows by applying
    Lemma~\ref{lemma:selective-fitting} which in particular says that
    $$
          \left(
            1 + 4\eta \cdot \frac{C_{\gamma}}{m} 2^{-m+1}\wmax
          \right)^{2t}
          \leq \frac{1}{\alpha}
    $$
    for any $t \leq \frac{m2^{m-1}}{32 \eta \wmax C_{\gamma}} \log
    \frac{1}{\alpha^{4}}$. Setting $C_{\gamma} = \frac{1}{128}$ yields the
    desired result.

    The lower-bound is then proved immediately by
    Lemma~\ref{lemma:handling-signal-shrinkage}.
  \end{enumerate}

  By above, the induction hypothesis holds for $i = m$.
  We can still repeat the argument for the first step of induction hypothesis
  to show that for any $t \geq \bar{T}_{m-1}$
  $$
    \norm{\vec{w}_{t} - \vec{w}^{\star}_{+}}_{\infty} \leq 2^{-m}\wmax \leq
    \zeta.
  $$
  Also, the proof for the third induction hypothesis with $i = m$ shows
  that for any $t \leq \bar{T}_{m-1}$ we have
  $$
    \alpha^{2}\prod_{i=0}^{t-1}(1 + 4\eta(
      \norm{\vec{b}_{t}}_{\infty} + \norm{\vec{p}_{t}}_{\infty}))^{2}
    \leq \alpha.
  $$
  To simplify the presentation, note that $\frac{\wmax}{\zeta} \leq 2^{m} <
  \frac{2\wmax}{\zeta}$ and hence we will write
  $$
    \bar{T}_{m-1}
    = (2^{m} - 1) \frac{1}{\eta \wmax} \log \frac{1}{\alpha^{4}}
    = O \left( \frac{1}{\eta \zeta} \log \frac{1}{\alpha} \right).
  $$

  Finally, regarding the absolute constants we have required in our proofs
  above that $C_{\gamma} + 2C_{b} \leq \frac{1}{40}$,
  $C_{b} \leq \frac{1}{2} C_{\gamma}$
  and $C_{\gamma} \leq \frac{1}{128}$. Hence, for example, absolute constants
  $C_{b} = \frac{1}{256}$ and
  $C_{\gamma} = \frac{1}{128}$ satisfy the requirements of this lemma.
\end{proof}

Extending the above lemma to the general setting considered in
Proposition~\ref{proposition:dealing-with-proportional-errors} can now be
done by a simple application of
Lemma~\ref{lemma:handling-negative-and-positive-target-simultaneously-helper}
as follows.

\begin{proof}[Proof of
  Proposition~\ref{proposition:dealing-with-proportional-errors}]

  Lemma~\ref{lemma:handling-negative-and-positive-target-simultaneously-helper}
  allows us to reduce this proof to
  lemma~\ref{lemma:dealing-with-errors-proportional-to-convergence-distance}
  directly.
  In particular, using notation from
  Lemma~\ref{lemma:dealing-with-errors-proportional-to-convergence-distance}
  and using
  Lemma~\ref{lemma:handling-negative-and-positive-target-simultaneously-helper}
  we maintain that for all $t \leq \bar{T}_{m-1}$
  \begin{align*}
    w^{\star}_{j} > 0 &\implies 0 \leq w^{-}_{t} \leq \alpha \\
    w^{\star}_{j} < 0 &\implies 0 \leq w^{+}_{t} \leq \alpha.
  \end{align*}
  Consequently, for $w^{\star}_{j} > 0$ we can ignore sequence
  $(w_{t,j}^{-})_{t \geq 0}$ by
  treating it as a part of bounded error $b_{t}$. The same holds for
  sequence $(w_{t,j}^{+})_{t \geq 0}$ when $w^{\star}_{j} < 0$.
  Then, for $w_{j}^{\star} > 0$ the sequence $(w_{t,j}^{+})$ evolves as follows
  $$
    w_{t+1,j}^{+}
    = w_{t,j}^{+}
      (1 - 4\eta(w_{t,j}^{+} - w^{\star}_{j}
        + (b_{t,j} - w^{-}_{t,j}) + p_{t, j}))^{2}
  $$
  which falls directly into the setting of
  lemma~\ref{lemma:dealing-with-errors-proportional-to-convergence-distance}.
  Similarly, if $w^{\star}_{j} < 0$ then
  \begin{align*}
    w_{t+1,j}^{-}
    &= w_{t,j}^{-}
      (1 + 4\eta(-w_{t,j}^{-} - w^{\star}_{j}
        + (b_{t,j} + w^{+}_{t,j}) + p_{t, j}))^{2} \\
    &= w_{t,j}^{-}
      (1 - 4\eta(w_{t,j}^{-} - \abs{w^{\star}_{j}}
        + (-b_{t,j} - w^{+}_{t,j}) - p_{t, j}))^{2}
  \end{align*}
  and hence this sequence also falls into the setting of
  lemma~\ref{lemma:dealing-with-errors-proportional-to-convergence-distance}.

  Finally, $\norm{\vec{e}_{t}}_{\infty} \leq \alpha$
  follows by Lemma~\ref{lemma:controlling-errors} and we are done.
\end{proof}

%% file: appendix/multiplicative_updates/proof_of_easy_setting_proposition.tex
\subsection{Proof of
Proposition~\ref{proposition:easy-setting-convergence}}
\label{appendix:multiplicative-updates:proof-of-easy-setting-proposition}

We split the proof of Proposition~\ref{proposition:easy-setting-convergence}
in two phases. First, using
Lemma~\ref{lemma:halving-errors-for-theorem-noiseless-recovery}
we show that $\norm{\vec{s}_{t} - \wstar}_{\infty}$ converges to $0$
with error
$\norm{\vec{b}_{t} \odot \id{S}}_{\infty}$ up to some absolute multiplicative constant.
From this point onward, we can apply
Lemma~\ref{lemma:iterates-with-bounded-errors-approach-target-exponentially-fast}
to handle convergence to each individual sequence $i$ on the true support $S$
up to the error
$\norm{\vec{b}_{t} \odot \id{i}}_{\infty}
\vee \sqrt{k}\delta\norm{\vec{b}_{t} \odot \id{S}}_{\infty}$.
This is exactly what allows us to approach an oracle-like performance with
the $\ell_{2}$ parameter estimation error depending on
$\log k$ instead of $\log d$ in the case of sub-Gaussian noise.

\begin{lemma}
  \label{lemma:halving-errors-for-theorem-noiseless-recovery}
  Consider the setting of updates given in
  equations~\eqref{eq:error-decompositions} and
  \eqref{eq:updates-equation-using-b-p-notation}.
  Fix any $\varepsilon > 0$ and
  suppose that the error sequences $(\vec{b}_{t})_{t \geq 0}$ and
  $(\vec{p}_{t})_{t \geq 0}$ satisfy the following for any $t \geq 0$:
  \begin{align*}
    \norm{\vec{b}_{t} \odot \id{S}}_{\infty}
    &\leq
    B, \\
    \norm{\vec{p}_{t}}_{\infty}
    &\leq
    \frac{1}{20} \norm{\vec{s}_{t} - \wstar}_{\infty}.
  \end{align*}
  Suppose that
  $$
    20B
    < \norm{\vec{s}_{0} - \wstar}_{\infty}
    \leq \frac{1}{5} \wmin.
  $$
  Then for $\eta \leq \frac{5}{96 \wmax}$ and any
  $t \geq \frac{5}{8\eta \wmin}$ we have
  $$
    \norm{\vec{s}_{t} - \wstar}_{\infty}
    \leq \frac{1}{2} \norm{\vec{s}_{0} - \wstar}_{\infty}.
  $$
\end{lemma}

\begin{proof}
Note that
$\norm{\vec{b}_{0}}_{\infty}
 + \norm{\vec{p}_{0}}_{\infty}
 \leq
 \frac{1}{10}
 \norm{\vec{s}_{0} - \wstar}_{\infty}$.
By Lemma~\ref{lemma:iteratates-with-bounded-errors-monotonic-behaviour}
for any $t \geq 0$ we have
$\norm{\vec{b}_{t}}_{\infty} + \norm{\vec{p}_{t}}_{\infty}
 \leq \frac{1}{10}\norm{\vec{s}_{0} - \wstar}_{\infty}$.
Hence, for any $i$ such that
$\abs{s_{0,i} - w^{\star}_{i}}
\leq
\frac{1}{2}\norm{\vec{s}_{0} - \wstar}_{\infty}$
Lemma~\ref{lemma:iteratates-with-bounded-errors-monotonic-behaviour}
guarantees that for any $t \geq 0$
we have
$\abs{s_{t,i} - w^{\star}_{i}}
\leq
\frac{1}{2}\norm{\vec{s}_{0} - \wstar}_{\infty}$
On the other hand, for any $i$ such that
$\abs{s_{0,i} - w^{\star}_{i}}
>
\frac{1}{2}\norm{\vec{s}_{0} - \wstar}_{\infty}$
by Lemma~\ref{lemma:iterates-with-bounded-errors-behaviour-near-convergence}
we have
$\abs{s_{t,i} - w^{\star}_{i}}
\leq
\frac{1}{2}\norm{\vec{s}_{0} - \wstar}_{\infty}$
for any $t \geq \frac{5}{8\eta\wmin}$
which is what we wanted to prove.
\end{proof}

\begin{proof}[Proof of Proposition~\ref{proposition:easy-setting-convergence}]
\hfill

Let $B \coloneqq \max_{j\in S}B_{j}$.
To see that $\norm{\vec{s}_{t} - \wstar}_{\infty}$
never exceeds $\frac{1}{5}\wmin$ we use the $B$-tube argument developed in
Section~\ref{appendix:multiplicative-updates:bounded-errors} and formalized in
Lemma~\ref{lemma:iteratates-with-bounded-errors-monotonic-behaviour}.

We begin by applying the Lemma~\ref{lemma:halving-errors-for-theorem-noiseless-recovery}
for $\log_{2} \frac{\wmin}{5(B \vee \varepsilon)}$ times.
Now we have
$\norm{\vec{s}_{t} - \wstar}_{\infty} < 20(B\vee \varepsilon)$
and so $\norm{\vec{p}_{t}}_{\infty} < \delta \sqrt{k} \cdot 20(B \vee \varepsilon)$
Hence, for any $i \in S$ we have
$$
  \norm{\vec{b}_{t} \odot \id{i}}_{\infty}
  + \norm{\vec{p}_{t}}_{\infty}
  \leq
  B_{i} + \sqrt{k}\delta20(B \vee \varepsilon).
$$

Hence for each coordinate $i \in S$ we can apply the first part of
Lemma~\ref{lemma:iterates-with-bounded-errors-approach-target-exponentially-fast}
so that after another
$t = \frac{15}{32\eta \wmin} \log \frac{\wmin}{5 \varepsilon}$
iterations we are done.

Hence the total number of required iterations is
at most $t \leq \frac{45}{32 \eta \wmin} \log \frac{\wmin}{\varepsilon}$.

\end{proof}

%% file: appendix/missing_proofs.tex
\section{Missing Proofs from
  Section~\ref{appendix:main-proofs:technical-lemmas}}
\label{appendix:missing-proofs}

This section provides proofs for the technical lemmas stated in
section~\ref{appendix:main-proofs:technical-lemmas}.

\subfile{appendix/missing_proofs/proof_of_error_controlling_lemma.tex}

\subfile{appendix/missing_proofs/proof_of_selective_fitting_lemma.tex}

\subfile{appendix/missing_proofs/proof_of_halving_errors_lemma.tex}
\subfile{appendix/missing_proofs/proof_of_rip_assumption_pt_lemma.tex}

\subfile{appendix/missing_proofs/proof_of_column_normalization_lemma.tex}

\subfile{appendix/missing_proofs/proof_of_bounding_max_noise_lemma.tex}

%% file: appendix/missing_proofs/proof_of_error_controlling_lemma.tex
\subsection{Proof of
  Lemma~\ref{lemma:controlling-errors}}
\label{appendix:missing-proofs:proof-of-lemma-on-controlling-errors}
  Looking at the updates given by
  equation~\ref{eq:updates-equation-using-b-p-notation}
  in appendix~\ref{appendix:main-proofs:set-up-and-intuition} we have
  \begin{align}
    \id{S^{c}} \odot \vec{e}_{t+1}
    &= \id{S^{c}} \odot
       \vec{w}_{t} \odot (\id{} - 4\eta(\vec{s}_{t} - \wstar + \vec{b}_{t} + \vec{p}_{t}))^{2} \\
    &= \id{S^{c}} \odot
       \vec{e}_{t} \odot
       \left( \id{S^{c}} - \id{S^{c}} \odot
          4\eta(\vec{s}_{t} - \wstar + \vec{b}_{t} + \vec{p}_{t})
       \right)^{2} \\
    &= \id{S^{c}} \odot
       \vec{e}_{t} \odot
       \left( \id{} - 4\eta(\vec{b}_{t} + \vec{p}_{t})
       \right)^{2}
  \end{align}
  and hence
  \begin{align*}
    \norm{\id{S^{c}} \odot \vec{e}_{t+1}}_{\infty}
    &\leq
    \norm{\id{S^{c}} \odot \vec{e}_{t}}_{\infty}
    (1 + 4\eta(\norm{\vec{b}_{t}}_{\infty} + \norm{\vec{p}_{t}}_{\infty}))^{2}
  \end{align*}
  which completes the proof for $\id{S^{c}} \odot \vec{e}_{t}$.

  On the other hand,
  Lemma~\ref{lemma:handling-negative-and-positive-target-simultaneously-helper}
  deals with $\id{S} \odot \vec{e}_{t}$ immediately and we are done.
\hfill\qedsymbol

%% file: appendix/missing_proofs/proof_of_selective_fitting_lemma.tex
\subsection{Proof of
Lemma~\ref{lemma:selective-fitting}}
\label{section:appendix:missing-proofs:proof-of-selective-fitting-lemma}

Note that
$$
  1 + 4\eta b_{t} \leq 1 + 4\eta B
$$
and hence
\begin{align*}
  x_{t}
  \leq x_{0}(1 + 4 \eta B)^{2t}.
\end{align*}
To ensure that $x_{t} \leq \sqrt{x_{0}}$ it is enough to ensure that the
right hand side of the above expression is not greater than $\sqrt{x_{0}}$.
This is satisfied by all $t$ such that
$$
  t
  \leq
  \frac{1}{2}
  \frac
  {\log \frac{1}{\sqrt{x_{0}}}}
  {\log \left( 1 + 4\eta B \right) }
$$
Now by using $\log x \leq x - 1$ we have
\begin{align*}
  \frac{1}{2}
  \frac
  {\log \frac{1}{\sqrt{x_{0}}}}
  {\log \left( 1 + 4 \eta B \right) }
  &\geq
  \frac{1}{2}
  \frac
  {\log \frac{1}{\sqrt{x_{0}}}}
  {4 \eta B} \\
  &=
  \frac{1}{32 \eta B}
  \log \frac{1}{x_{0}^{2}}
\end{align*}
which concludes our proof.
\hfill\qedsymbol

%% file: appendix/missing_proofs/proof_of_rip_assumption_pt_lemma.tex
\subsection{Proof of
  Lemma~\ref{lemma:rip-assumption-pt}}
\label{appendix:missing-proofs:proof-rip-assumption-pt-lemma}

For any index set $S$ of size $k + 1$ let
$\matrix{X}_{S}$ be the
$n \times (k + 1)$ sub-matrix of $\X$ containing columns indexed by $S$.
Let $\lambda_{\max}\left(\frac{1}{n} \matrix{X}_{S}^{\mathsf{T}}\matrix{X}_{S}\right)$
and $\lambda_{\min}\left(\frac{1}{n} \matrix{X}_{S}^{\mathsf{T}}\matrix{X}_{S}\right)$
denote the maximum and minimum eigenvalues of
$\left(\frac{1}{n} \matrix{X}_{S}^{\mathsf{T}}\matrix{X}_{S}\right)$ respectively.
It is then a standard consequence of the $(k + 1, \delta)$-RIP that
$$
 1 - \delta
 \leq
 \lambda_{\min}\left(\frac{1}{n} \matrix{X}_{S}^{\mathsf{T}}\matrix{X}_{S}\right)
 \leq
 \lambda_{\max}\left(\frac{1}{n} \matrix{X}_{S}^{\mathsf{T}}\matrix{X}_{S}\right)
 \leq
 1 + \delta.
$$
Let $\vec{z} \in \mathbb{R}^{d}$ be any $k$-sparse vector.
Then, for any $i \in \{1, \dots, d\}$ the joint support of $\id{i}$
and $\vec{z}$ is of size at most $k + 1$. We denote the joint support by $S$
and
we will also denote by $\vec{z}_{S}$ and $(\id{i})_{S}$
the restrictions of
$\vec{z}$ and $\id{i}$ on their support, i.e., vectors in
$\mathbb{R}^{k+1}$.
Letting $\norm{\cdot}$ be the spectral norm, we have
\begin{align*}
  \abs{\left(\frac{1}{n}\XtX \vec{z}\right)_{i} - \vec{z}_{i}}
  &=
  \abs{
    \left\langle \frac{1}{n} \XtX \vec{z}, \id{i} \right\rangle
    -
    \left\langle \vec{z}, \id{i} \right\rangle
  } \\
  &=
  \abs{
    \left\langle \frac{1}{\sqrt{n}}\X\vec{z},
                 \frac{1}{\sqrt{n}}\X\id{i} \right\rangle
    -
    \left\langle \vec{z}, \id{i} \right\rangle
  } \\
  &=
  \abs{\left\langle
        \frac{1}{\sqrt{n}}\matrix{X}_{S}\vec{z}_{S},
        \frac{1}{\sqrt{n}}\matrix{X}_{S}(\id{i})_{S}
       \right\rangle
     - \left\langle \vec{z}_{S}, (\id{i})_{S} \right\rangle} \\
  &=
  \abs{\left\langle
       \left(
           \frac{1}{n}\matrix{X}_{S}^{\mathsf{T}}\matrix{X}_{S} - \matrixid
       \right)\vec{z}_{S},
       (\id{i})_{S}
       \right\rangle} \\
  &\leq
    \norm{
    \frac{1}{n}\matrix{X}_{S}^{\mathsf{T}}\matrix{X}_{S} - \matrixid}
    \norm{\vec{z}}_{2}
    \norm{\id{i}}_{2} \\
  &\leq
    \delta
    \norm{\vec{z}}_{2}
\end{align*}
where the penultimate line follows by the Cauchy-Schwarz inequality and the
last line follows by the $(k+1, \delta)$-RIP.
Since $i$ was arbitrary it hence follows that
\begin{align*}
  \norm{\left( \frac{1}{n} \XtX - \matrixid \right)\vec{z}}_{\infty}
  \leq
  \delta \norm{\vec{z}}_{2}
  \leq
  \delta \sqrt{k} \norm{\vec{z}}_{\infty}.
\end{align*}
\hfill\qedsymbol

%% file: appendix/missing_proofs/proof_of_column_normalization_lemma.tex
\subsection{Proof of
  Lemma~\ref{lemma:column-normalization}}
\label{appendix:missing-proofs:proof-of-coumn-normalization-lemma}

For any $i \in \{1, \dots, d\}$
we can write $\vec{X}_{i} = \X \id{i}$. The result is then immediate by
the $(k+1, \delta)$-RIP since
$$
  \norm{\frac{1}{\sqrt{n}}\X \id{i}}_{2}^{2}
  \leq (1 + \delta)\norm{\id{i}}_{2}^{2}
  \leq 2.
$$
By the Cauchy-Schwarz inequality we then have, for any $i,j \in \{1, \dots,
  d\}$,
$$
  \abs{\left(\frac{1}{n}\XtX\right)_{i,j}}
  \leq
  \norm{\frac{1}{\sqrt{n}}\vec{X}_{i}}_{2}
  \norm{\frac{1}{\sqrt{n}}\vec{X}_{j}}_{2}
  \leq
  2
$$
and for any $\vec{z} \in \mathbb{R}^{d}$ it follows that
$$
  \norm{\frac{1}{n}\XtX \vec{z}}_{\infty} \leq 2d\norm{\vec{z}}_{\infty}.
$$
\hfill\qedsymbol

%% file: appendix/missing_proofs/proof_of_bounding_max_noise_lemma.tex
\subsection{Proof of
  Lemma~\ref{lemma:bounding-max-noise}}
\label{appendix:missing-proofs:proof-of-bounding-max-noise-lemma}

Since for any column $\vec{X}_{i}$ of the matrix $\X$ we have
$\norm{X_{i}}_{2} / \sqrt{n} \leq C$ and since the vector
$\vec{\xi}$ consists of independent $\sigma^{2}$-subGaussian random variables,
the random variable $\frac{1}{\sqrt{n}}\left(\Xt \vec{\xi}\right)_{i}$
is $C^{2}\sigma^{2}$-subGaussian.

It is then a standard result that
for any $\varepsilon > 0$
$$
  \pr{\norm{\frac{1}{\sqrt{n}}\Xt\vec{\xi}}_{\infty} > \varepsilon}
  \leq 2de^{-\frac{\varepsilon^{2}}{2C^{2}\sigma^{2}}}.
$$
Setting $\varepsilon = 2\sqrt{2C^{2}\sigma^{2}\log (2d)}$
we have with probability at least $1 - \frac{1}{8d^{3}}$ we have
\begin{align*}
  \norm{\frac{1}{\sqrt{n}}\Xt\vec{\xi}}_{\infty}
  &\leq
  4\sqrt{C^{2}\sigma^{2}\log (2d)} \\
  &\lesssim
  \sqrt{\sigma^{2} \log d}.
\end{align*}
\hfill\qedsymbol

%% file: appendix/eta_initialization.tex
\section{Proof of Theorem~\ref{thm:selecting-the-step-size}}
\label{appendix:selecting-the-step-size}

Recall the updates equations for our model parameters given in
equations~\eqref{eq:error-decompositions} and
\eqref{eq:updates-equation-using-b-p-notation} as defined in
Appendix~\ref{appendix:main-proofs:set-up-and-intuition}.

Since $\vec{w}_{0} = 0$ we can rewrite the first update written on $\vec{u}$
and $\vec{v}$ as
\begin{align}
  \begin{split}
    \label{eq:step-one-updates}
    \vec{u}_{1}
    &= \vec{u}_{0} \odot \left(
         \id{}
         - 4\eta \left(
             -\wstar
             + \left(\matrixid - \frac{1}{n} \XtX\right) \wstar - \frac{1}{n} \Xt \xi)
           \right)
         \right), \\
    \vec{v}_{1}
    &= \vec{v}_{0} \odot \left(
         \id{}
         + 4\eta \left(
             -\wstar
             + \left(\matrixid - \frac{1}{n} \XtX\right) \wstar - \frac{1}{n} \Xt \xi)
           \right)
         \right).
  \end{split}
\end{align}

By Lemma~\ref{lemma:rip-assumption-pt} we have
$\norm{\left(\matrixid - \frac{1}{n} \XtX \right) \wstar}_{\infty}
\leq \frac{1}{20} \wmax$.
The term $\frac{1}{n} \Xt \xi$ can be simply bounded by $\maxnoise$.
If $\wmax \geq 5\maxnoise$ (note that otherwise returning a $0$ vector is
minimax-optimal) then
$$
  \frac{1}{20}\wmax + \maxnoise \leq \frac{1}{4} \wmax.
$$

We can hence bound the below term appearing in
equation~\eqref{eq:step-one-updates} as follows:
$$
  \frac{3}{4} \wmax
  \leq
  \norm{-\wstar
  + \left(\matrixid - \frac{1}{n} \XtX\right) \wstar - \frac{1}{n} \Xt \xi)}_{\infty}
  \leq
  \frac{5}{4} \wmax
$$
The main idea here is that we can recover the above factor by computing one
gradient descent iteration and hence we can recover $\wmax$ up to some
multiplicative constants.

In fact, with $0 < \eta \leq \frac{1}{5 \wmax}$ so that the multiplicative
factors are non-negative, the above inequality implies that
$$
 1 + 3\eta \wmax \leq f_{\max} \leq 1 + 5\eta \wmax
$$
and so
$$
  \wmax
  \leq
  \frac{f_{\max} - 1}{3\eta}
  \leq
  \frac{5}{3}\wmax
$$
which is what we wanted to show.

Note that after an application of this theorem we can now reset the step size
to
$$
  \frac{3 \eta}{20 \left(f_{\max} - 1 \right)}.
$$
This new step size satisfies the conditions of
Theorems~\ref{thm:constant-step-sizes}
and \ref{thm:main-theorem-noisy-minimax-rates-exponential-convergence}
while being at most two times smaller than required.

%% file: appendix/increasing_step_sizes.tex
\section{Proof of
Theorem~\ref{thm:main-theorem-noisy-minimax-rates-exponential-convergence}}
\label{appendix:theorem-minimax-rates-exponential-convergence-proof}

For proving
Theorem~\ref{thm:main-theorem-noisy-minimax-rates-exponential-convergence}
we first prove
Propositions~\ref{proposition:dealing-with-proportional-errors-with-increasing-step-sizes}
and \ref{proposition:easy-setting-convergence-with-increasing-step-sizes}
which correspond to
Propositions~\ref{proposition:dealing-with-proportional-errors}
and
\ref{proposition:easy-setting-convergence}
but allows for different step sizes along each dimension.
We present the proof of
Proposition~\ref{proposition:dealing-with-proportional-errors-with-increasing-step-sizes}
in
Section~\ref{appendix:increasing-step-sizes:key-proposition}.

\begin{proposition}
  \label{proposition:dealing-with-proportional-errors-with-increasing-step-sizes}
  Consider the setting of
  Proposition~\ref{proposition:dealing-with-proportional-errors} and run
  Algorithm~\ref{alg:gd-increasing-steps} with
  $\tau = 640$.

  Then, for some early stopping time
  $
    T
    = O \left(\log {\frac{\wmax}{\zeta}} \log \frac{1}{\alpha}\right)
  $
  and any $0 \leq t \leq T$
  we have
  \begin{align*}
    \norm{\vec{s}_{T} - \vec{w}^{\star}}_{\infty}
    &\leq \zeta, \\
    \norm{\vec{e}_{t}}_{\infty}
    &\leq \alpha.
  \end{align*}
  Further, let $\eta_{T, j}$ be the step size for the $j\th$ coodinate at time
  $T$. Then, for all $j$ such that $\inlineabs{w^{\star}_{j}} > \zeta$ we have
  $$
    \frac{1}{16}\cdot\frac{1}{20 \abs{w^{\star}_{j}}}
    \leq \eta_{T, j}
    \leq \frac{1}{20 \abs{w^{\star}_{j}}}.
  $$
\end{proposition}

\begin{proposition}
  \label{proposition:easy-setting-convergence-with-increasing-step-sizes}
  Consider the setting of updates given in
  equations~\eqref{eq:error-decompositions} and
  \eqref{eq:updates-equation-using-b-p-notation}.
  Fix any $\varepsilon > 0$ and
  suppose that the error sequences $(\vec{b}_{t})_{t \geq 0}$ and
  $(\vec{p}_{t})_{t \geq 0}$ satisfy for any $t \geq 0$:
  \begin{align*}
    \norm{\vec{b}_{t} \odot \id{i}}_{\infty}
    &\leq B_{i} \leq \frac{1}{10}\wmin, \\
    \norm{\vec{p}_{t}}_{\infty}
    &\leq
    \frac{1}{20} \norm{\vec{s}_{t} - \wstar}_{\infty}.
  \end{align*}
  Suppose that
  $$
    \norm{\vec{s}_{0} - \wstar}_{\infty}
    \leq \frac{1}{5} \wmin.
  $$
  For each $i \in S$ let the step size satisfy
  $\frac{1}{\eta_{i} \abs{w^{\star}_{i}}} \leq 320$. Then for
  all $t \geq 0$
  $$
    \norm{\vec{s}_{t} - \wstar}_{\infty} \leq \frac{1}{5}\wmin
  $$
  and for any
  $t \geq 450 \log \frac{\wmin}{\varepsilon}$ we have
  for any $i \in S$.
  $$
    \abs{s_{t,i} - w^{\star}_{i}}
    \lesssim \delta \sqrt{k} \max_{j \in S} B_{j}
    \vee B_{i}
    \vee \varepsilon
  $$
\end{proposition}

\begin{proof}
  We follow the same strategy as in the proof of
  Proposition~\ref{proposition:easy-setting-convergence}.
  The only difference here is that the worst case convergence time
  $\frac{1}{\eta\wmin}$ is
  replaced by
  $\max_{i\in S} \frac{1}{\eta_{i} \abs{w^{\star}_{i}}} \leq 320$
  and the result follows.
\end{proof}

\begin{proof}[Proof of
  Theorem~\ref{thm:main-theorem-noisy-minimax-rates-exponential-convergence}]

  The proof is identical to the proof of
  Theorem~\ref{thm:constant-step-sizes} with application
  of Proposition~\ref{proposition:dealing-with-proportional-errors} replaced
  with
  Proposition~\ref{proposition:dealing-with-proportional-errors-with-increasing-step-sizes}
  and in the easy setting the application of
  Proposition~\ref{proposition:easy-setting-convergence} replaced with
  an application of
  Proposition~\ref{proposition:easy-setting-convergence-with-increasing-step-sizes}.

  The only difference is that extra care must be taken when applying
  Proposition~\ref{proposition:easy-setting-convergence-with-increasing-step-sizes}.
  First, note that the pre-conditions on step sizes are satisfied by
  Proposition~\ref{proposition:dealing-with-proportional-errors-with-increasing-step-sizes}.
  Second, the number of iterations required by
  Proposition~\ref{proposition:easy-setting-convergence-with-increasing-step-sizes}
  is fewer than step-size doubling intervals, and hence the step sizes will not
  change after the application of
  Proposition~\ref{proposition:dealing-with-proportional-errors-with-increasing-step-sizes}.
  In particular,
  Proposition~\ref{proposition:dealing-with-proportional-errors-with-increasing-step-sizes}
  requires $450 \log \frac{\wmin}{\varepsilon}$ iterations and we double the
  step sizes every $640 \log \frac{1}{\alpha}$ iterations. This finishes our
  proof.
\end{proof}

\subfile{appendix/increasing_step_sizes/key_proposition.tex}

%% file: appendix/increasing_step_sizes/key_proposition.tex
\subsection{Proof of
Proposition~\ref{proposition:dealing-with-proportional-errors-with-increasing-step-sizes}}
\label{appendix:increasing-step-sizes:key-proposition}

Recall the proof of
Proposition~\ref{proposition:dealing-with-proportional-errors} that we have
shown in
Appendix~\ref{appendix:multiplicative-updates:proof-of-key-proposition}.
We have used a constant step size
$\eta \leq \frac{5}{96\wmax}$. With a constant step size this is in fact
unavoidable up to multiplicative constants -- for larger step sizes the
iterates can explode.

Looking at our proof by induction of
Lemma~\ref{lemma:dealing-with-errors-proportional-to-convergence-distance},
the inefficiency of Algorithm~\ref{alg:gd} comes from doubling the number of iterations during each
induction step. This happens because during the $i\th$ induction step
the smallest coordinates of $\wstar$ that we consider are of size
$2^{-i-1}\wmax$. For such coordinates, step size $\eta \leq \frac{5}{96\wmax}$
could be at least $2^{i}$ times bigger and hence the convergence would be
$2^{i}$ times faster.
The lemmas derived in
Appendix~\ref{appendix:multiplicative-updates:bounded-errors} indicate that
fitting signal of such size will require number of iterations proportional
to $\frac{1}{\eta 2^{-i-1} \wmax} = 2^{i+1}\frac{1}{\eta \wmax}$ which is
where the exponential increase in the number of iterations for each induction
step comes from.

We can get rid of this inefficiency if for each coordinate $j$ we use a
different step size, so that for all $j$ such that $\abs{w_{j}^{\star}} \ll \wmax$
we set $\eta_{j} \gg \frac{5}{96 \wmax}$.
In fact, the only constraint we have is that $\eta_{j}$ never exceeds
$\frac{5}{96 \abs{w^{\star}}_{j}}$. To see that we can change the step sizes for
small enough signal in practice, note that after two induction steps in
Proposition~\ref{proposition:dealing-with-proportional-errors} we have
$\norm{\vec{s}_{t} - \vec{w}^{\star}}_{\infty} \leq \frac{1}{4} \wmax$ and
$\norm{\vec{e}_{t}}_{\infty} \leq \alpha$. We can then show, that for each
$j$ such that $\abs{w^{\star}}_{j} > \frac{1}{2} \wmax$ we have
$\abs{w_{t,j}} > \frac{1}{4} \wmax$. On the other hand, if
$\abs{w^{\star}}_{j} \leq \frac{1}{8} \wmax$ then
$w_{t, j} \leq w^{\star}_{j} + 4B_{1} \leq \frac{1}{4}\wmax$, where
$B_{1}$ is given as in
Lemma~\ref{lemma:dealing-with-errors-proportional-to-convergence-distance}.
In particular, after the second induction step one can take all
$j$ such that $\abs{w_{t, j} } \leq \frac{1}{4} \wmax$ and double its
associated step sizes.

We exploit the above idea in the following lemma, which is a counterpart
to Lemma~\ref{lemma:dealing-with-errors-proportional-to-convergence-distance}.
One final thing to note is that we do not really know what $\wmax$ is which
is necessary in the argument sketched above.
However, in Theorem~\ref{thm:selecting-the-step-size}
we showed that
we can compute some $\hat{z}$ such that $\wmax \leq \hat{z} \leq 2 \wmax$
and as we shall see this is enough.

\begin{lemma}[Counterpart to
  Lemma~\ref{lemma:dealing-with-errors-proportional-to-convergence-distance}
  with increasing step sizes]
  \label{lemma:dealing-with-errors-proportional-to-convergence-distance-with-increasing-step-sizes}
  Consider the same setting of
  Lemma~\ref{lemma:dealing-with-errors-proportional-to-convergence-distance}.
  Run Algorithm~\ref{alg:gd-increasing-steps} with $\tau = 640$
  and parametrization $\vec{w}_{t} = \vec{u}_{t} \odot \vec{u}_{t}$.

  Then, for $t = \ceil{640 \log_{2} \frac{\wmax}{\zeta} \log \frac{1}{\alpha}}$
  and any $j$ we have
  \begin{align*}
    \norm{\vec{w}_{t} - \vec{w}^{\star}_{+}}_{\infty}
    &\leq \zeta \\
    \alpha^{2}\prod_{i=0}^{t-1}(1 + 4\eta_{t,j}(
      \norm{\vec{b}_{t}}_{\infty} + \norm{\vec{p}_{t}}_{\infty}))^{2}
    &\leq \alpha.
  \end{align*}
\end{lemma}

\begin{proof}
  Following the notation used in
  Lemma~\ref{lemma:dealing-with-errors-proportional-to-convergence-distance}
  for any integer $i \geq -1$ let $T_{i} \coloneqq T$ and
  $\bar{T}_{i} \coloneqq \sum_{j=0}^{i} T_{j} = (i+1) T$.
  We remark now that we have the same $T$ for each induction step in contrast
  to exponentially increasing number of iterations in
  Lemma~\ref{lemma:dealing-with-errors-proportional-to-convergence-distance}.
  Let $B_{i} \coloneqq \frac{1}{40}2^{-i}\wmax$.
  Let $m = \ceil{\log_{2} \frac{\wmax}{\zeta}}$
  so that $\gamma = \frac{C_{\gamma}}{m}$.
  We will prove our claim by induction on $i = 0, 1, \dots, m-1$.

  \underline{\textit{Induction hypothesis for $i \in \{0, \dots, m \}$}} \\
  \begin{enumerate}
    \item For any $j < i$ and $\bar{T}_{j-1} \leq t < \bar{T}_{j}$
      we have $\norm{\vec{w}_{t} - \vec{w}^{\star}_{+}}_{\infty} \leq 2^{-j} \wmax$.
      In particular, this induction hypothesis says that we halve the
      convergence distance during each induction step.
    \item We have
      $\norm{\vec{w}_{\bar{T}_{i-1}} - \vec{w}^{\star}_{+}}_{\infty} \leq 2^{-i}
      \wmax$. This hypothesis controls the convergence distance at the
      beginning of each induction step.
    \item
      For any $j$ such that $w^{\star}_{j} \leq 20B_{i} = 2^{-i-1}\wmax$
      we have $\alpha^{3} \leq w_{\bar{T}_{i-1}, j} \leq w_{j}^{\star} + 4B_{i}$.
      On the other hand, for any $j$ such that $w^{\star}_{j} \geq 20B_{i}$
      we have $\alpha^{3} \leq w_{\bar{T}_{i-1}, j} \leq \frac{6}{5}w_{j}^{\star}$.
    \item
      Let $l$ be any integer such that $0 \leq l \leq i$.
      Then for any $j$ such that
      $2^{-l-1}\wmax < w^{\star}_{j} \leq 2^{-l}\wmax$
      we have
      $$
      2^{l-3}\eta_{0,j}
      \leq
      \eta_{\bar{T}_{i-1}, j}
      \leq
      2^{l}\eta_{0,j}
      $$
    For any $j$ such that $w^{\star}_{j} \leq 2^{-i-1}$ we have
    $$
      2^{i-2}\eta_{0,j}
      \leq \eta_{\bar{T}_{i-1}, j}
      \leq 2^{(i-1) \vee 0}\eta_{0, j}.
    $$
    In particular, the above conditions ensure that we $\eta_{t, j}$
    never exceeds $\frac{1}{20 w^{\star}_{j}}$ so that the step-size pre-conditions
    of all lemmas derived in previous appendix sections always hold
    during each induction step.
    Further, it ensures
    that once we fit small coordinates, the step size is up to absolute
    constants as big as possible.
  \end{enumerate}
  We remark the that in addition to induction hypotheses used in
  Lemma~\ref{lemma:dealing-with-errors-proportional-to-convergence-distance}
  the fourth induction hypothesis allows to control what happens to the step
  sizes with our doubling step size scheme.
  There is also a small modification to the third induction hypothesis, where
  right now we sometimes allow $w_{t,j} > w^{\star}_{j} + 4B_{i}$ because due to
  increasing step sizes we have to deal iterates larger than target slightly
  differently. In particular, we can only apply
  Lemma~\ref{lemma:handling-large-errors} for coordinates $j$ with sufficiently
  small $w^{\star}_{j}$, because the step sizes of such coordinates will be
  larger which allows for faster convergence.

  \underline{\textit{Base case}} \\
  For $i = 0$ all conditions hold since for all $j$ we have
  $0 \leq \alpha^{2} = w_{0, j} < w^{\star}_{j}$
  and since all $\eta_{0, j} \leq \frac{1}{20 \wmax}$.

  \underline{\textit{Induction step}} \\
  Assume that the induction hypothesis holds for some $0 \leq i < m$.
  We will show that it also holds for $i + 1$.
  \begin{enumerate}
    \item
    The proof is based on monotonic convergence to $B_{i}$ tube argument
    and is identical to the one used in
    Lemma~\ref{lemma:dealing-with-errors-proportional-to-convergence-distance}
    with the same conditions on $C_{b}$ and $C_{\gamma}$.

    \item
    Similarly to the proof of
    Lemma~\ref{lemma:dealing-with-errors-proportional-to-convergence-distance}
    here we only need to handle coordinates $j$ such that
    $w^{\star}_{j} > 20B_{i} = 2^{-i-1}\wmax$
    and $\abs{w_{\bar{T}_{i-1}, j} - w^{\star}_{j}} > 2^{-i-1}\wmax$.

    If $w_{\bar{T}_{i-1}, j} \leq w^{\star}_{j}$
    we apply the second part of
    Lemma~\ref{lemma:iterates-with-bounded-errors-approach-target-exponentially-fast}
    with $\varepsilon = 19B_{i}$ to obtain that for any
    \begin{align*}
      t
      &\geq
        \frac{1}{2} \frac{1}{\eta_{\bar{T}_{i-1}, j} w^{\star}_{j}}
        \log \frac{1}{\alpha^{4}} \\
      &\geq
        \frac{15}{32 \eta w_{j}^{\star}}
        \frac{1}{\eta_{\bar{T}_{i-1}, j} w^{\star}_{j}}
        \log \frac{(w_{j}^{\star})^{2}}{w_{\bar{T}_{i-1}, j} \cdot 19 B_{i}}
    \end{align*}
    iterations the following holds
    $$
      \abs{w_{\bar{T}_{i-1} + t, j} - w^{\star}_{j}} \leq
      20B_{i} \leq 2^{-i-1}\wmax.
    $$
    By the fourth induction hypothesis and by definition of $\eta_{0, j}$
    we have
    $$
      \frac{1}{\eta_{\bar{T}_{i-1}, j} w^{\star}_{j}}
      \leq \frac{8}{\eta_{0,j}\wmax}
      \leq 16 \cdot 20.
    $$
    and hence $T$ iterations are enough.

    If $w_{\bar{T}_{i-1}, j} \geq w^{\star}_{j}$
    by the third induction hypothesis we also have $w_{\bar{T}_{i-1}, j} \leq
    \frac{6}{5}w^{\star}_{j}$ so that the pre-condition of
    Lemma~\ref{lemma:iterates-with-bounded-errors-behaviour-near-convergence}
    apply and we are done, since it requires fewer iterations than considered
    above.

    \item We first deal with the upper-bound. For $j$ such that
      $w^{\star}_{j} \geq 20B_{i}$ we have by the third induction hypothesis
      $w_{\bar{T}_{i-1}, j} \leq \frac{6}{5}w^{\star}_{j}$ and hence by the
      monotonic convergence to $B_{i}$-tube argument given in
      Lemma~\ref{lemma:iteratates-with-bounded-errors-monotonic-behaviour}
      this bound still holds after the $i\th$ induction step.
      For any $j$ such that $w^{\star}_{j} \leq 20B_{i}$ we use
      Lemma~\ref{lemma:handling-large-errors}
      and the fourth induction hypothesis
      $\eta_{\bar{T}_{i-1}, j} \geq 2^{i-3}\eta_{0, j}$ to show that
      after
      \begin{align*}
        T
        \geq \frac{32}{\eta_{0, j} \wmax}
        \geq \frac{2^{i+2}}{\eta_{\bar{T}_{i-1}, j} \wmax}
        = \frac{1}{10 \eta_{\bar{T}_{i-1}, j} B_{i}}.
      \end{align*}
      iterations for any such $j$ we have
      $w_{\bar{T}_{i-1} + t} \leq w^{\star}_{j} + 2B_{i} = w^{\star}_{j} + 4B_{i+1}$.
      Finally, this implies that if $10B_{i} \leq w^{\star}_{j} \leq 20B_{i}$
      then after $T$ iterations $w_{\bar{T}_{i}, j} \leq \frac{6}{5}
      w^{\star}_{j}$.

      To prove the lower-bound, note that during the $i\th$ induction step
      for any $j$ we have $\eta_{j, \bar{T}_{i-1}} \leq 2^{i}\eta_{0, j}$
      since each step size at most doubles after every induction step.
      Hence during the $i\th$ induction step, the accumulation of error can
      be upper-bounded by
      \begin{align*}
        &\prod_{i = \bar{T}_{i-1}}^{\bar{T}_{i} - 1}
        (1 + 4\eta_{\bar{T}_{i-1}, j}(\norm{\vec{b}_{i}}_{\infty} +
        \norm{\vec{p}_{i}}_{\infty}))^{2} \\
        &\leq
        (1 + 4\cdot 2^{i}\eta_{0, j}(\norm{\vec{b}_{i}}_{\infty} +
        \norm{\vec{p}_{i}}_{\infty}))^{2T} \\
        &\leq
        (1 + 4\cdot \eta_{0, j}(\norm{\vec{b}_{i}}_{\infty} +
        \norm{\vec{p}_{i}}_{\infty}))^{2 \cdot 2^{i}T}.
      \end{align*}
      Now since our $2_{i}T$ is simply the same $T_{i}$ as used in
      Lemma~\ref{lemma:dealing-with-errors-proportional-to-convergence-distance}
      rescaled at most $8$ times, the same bounds holds on the
      accumulation of error
      as in Lemma~\ref{lemma:dealing-with-errors-proportional-to-convergence-distance}
      with absolute constants $C_{b}$ and $C_{\gamma}$ rescaled by
      $\frac{1}{8}$ in this lemma. This completes the third induction
      hypothesis step.

    \item
      After the $i\th$ induction step (recall that the induction steps are
      numbered starting from $0$), if $i \geq 1$ our step size scheme
      doubles $\eta_{\bar{T}_{i}, j}$ if $w_{\bar{T}_{i}, j} \leq 2^{-i-2}\hat{z}$.
      Recall that after $i\th$ induction step we have
      $\norm{\vec{w}_{t} - \vec{w}^{\star}_{+}}_{\infty}
      \leq 2^{-i-1}\wmax$.

      For every $j$ such that
      $w^{\star}_{j} > 2^{-i}\wmax$ we have $w_{\bar{T}_{i}, j} >
      2^{-i-1}\wmax \geq 2^{-i-2}\hat{z}$ and hence $\eta_{\bar{T}_{i}, j}$
      will not be affected.

      For every $j$ such that
      $w^{\star}_{j} \leq 2^{-i-3}\wmax$ we have
      $w_{\bar{T}_{i}, j} \leq w^{\star}_{j} + 4B_{i+1} \leq 2^{-i-2}\wmax$
      and for such $j$ the step size will be doubled.

      Hence for any non-negative integer $k$ and any $j$ such that
      $2^{-k-1} \wmax < w^{\star}_{j} \leq 2^{-k}\wmax$ the corresponding step size
      will be doubled after $i\th$ induction step for
      $i = 1, \dots, k-3$ and will not be touched anymore after and including
      the $k + 1\th$ induction step.
      We are only uncertain about what happens for such $j$ after the
      $k-2, k-1$ and $k\th$ induction steps, which is where the factor of $8$
      comes from. This concludes the proof of the fourth induction hypothesis.
  \end{enumerate}

  The result then follows after $mT$ iterations which is what we wanted to
  show.
\end{proof}

Similarly to the proof of
Proposition~\ref{proposition:dealing-with-proportional-errors} we can extend
the above Lemma to a general setting (i.e. parametrization $\vec{w}_{t} \coloneqq
\vec{u}_{t} \odot \vec{u}_{t} - \vec{v}_{t} \odot \vec{v}_{t}$) by using
Lemma~\ref{lemma:handling-negative-and-positive-target-simultaneously-helper}.
The following proposition then corresponds to
Proposition~\ref{proposition:dealing-with-proportional-errors} but allows to
use our increasing step sizes scheme.

\begin{proof}[Proof of
  Proposition~\ref{proposition:dealing-with-proportional-errors-with-increasing-step-sizes}]
  Immediate by
  Lemma~\ref{lemma:handling-negative-and-positive-target-simultaneously-helper}
  by the same argument as used in the proof of
  Proposition~\ref{proposition:dealing-with-proportional-errors}.
\end{proof}

%% file: appendix/gradient_descent_updates.tex
\section{Gradient Descent Updates}
\label{appendix:gradient-descent-updates}

We add the derivation of gradient descent updates for completeness.
Let $\vec{w} = \vec{u} \odot \vec{u} - \vec{v} \odot \vec{v}$ and suppose
$$
  \mathcal{L}(\vec{w}) = \frac{1}{n}\norm{\X\vec{w} - \vec{y}}^{2}_{2}.
$$
We then have for any $i = 1, \dots, d$
\begin{align*}
  \frac{\partial}{\partial u_{i}}
  \mathcal{L}(\vec{w})
  &=
  \frac{1}{n}
  \sum_{j=1}^{n}
  \frac{\partial}{\partial u_{i}}
  (\X \vec{w} - \vec{y})_{j}^{2} \\
  &=
  \frac{1}{n}
  \sum_{j=1}^{n}
  2(\X \vec{w} - \vec{y})_{j} \cdot \frac{\partial}{\partial u_{i}}
  (\X \vec{w} - \vec{y})_{j} \\
  &=
  \frac{1}{n}
  \sum_{j=1}^{n}
  2(\X \vec{w} - \vec{y})_{j} \cdot \frac{\partial}{\partial u_{i}}
  (\X(\vec{u} \odot \vec{u}))_{j} \\
  &=
  \frac{1}{n}
  \sum_{j=1}^{n}
  2(\X \vec{w} - \vec{y})_{j}
  \cdot
  2u_{i} X_{ji} \\
  &=
  4u_{i}
  \frac{1}{n}
  \sum_{j=1}^{n}
  X_{ji}(\X \vec{w} - \vec{y})_{j} \\
  &=
  4u_{i} \frac{1}{n}
  \left(
    \Xt(\X \vec{w} - \vec{y})
  \right)_{i}
\end{align*}
and hence
\begin{align*}
  \nabla_{\vec{u}} \mathcal{L}(\vec{w}) &= \frac{4}{n} \Xt(\X \vec{w} -
  \vec{y}) \odot \vec{u}, \\
  \nabla_{\vec{v}} \mathcal{L}(\vec{w}) &= - \frac{4}{n} \Xt(\X\vec{w} -
  \vec{y}) \odot \vec{v}.
\end{align*}

%% file: appendix/comparing_with_colt_assumptions.tex
\section{Comparing Assumptions to \texorpdfstring{\cite{li2018algorithmic}}{Li
et al. [2018]}}
\label{appendix:comparing-with-colt-paper}

We compare our conditions on $\alpha, \delta$ and $\eta$
to the related work analyzing implicit regularization effects
of gradient descent for noiseless low-rank matrix recovery problem with a
similar parametrization \cite{li2018algorithmic}.

The parameter $\alpha$ plays a similar role in both papers:
$\ell_{2}$ (or reconstruction) error in the noiseless setting is directly
controlled by the size of $\alpha$ as we show in
Corollary~\ref{corollary:noiseless}. In both settings the number of iterations
is affected only by a multiplicative factor of $O(\log 1 / \alpha)$.

The conditions imposed on $\alpha$ and $\eta$ in \cite{li2018algorithmic} are much stronger than required in our work.
Our results do not follow  from the main result of
\cite{li2018algorithmic} by considering a matrix recovery problem for
the ground truth matrix $\operatorname{diag}(\wstar)$.
Letting $\kappa = \wmax/\wmin$
the assumptions of \cite{li2018algorithmic} require
$\delta \lesssim 1/(\kappa^{3} \sqrt{k} \log^{2} d)$
and $\eta \lesssim \delta$ yielding
$\Omega(\kappa / \eta \log 1/\alpha) = \Omega(\kappa^{4} \log^{2} d \sqrt{k}
\log 1/\alpha)$ iteration complexity.
In contrast, our theorem only requires $\delta$ to scale only as $1/\log\kappa$.
We are able to set the step-size using data and do not rely on knowing the unknown quantities $\kappa$ and $k$.

Crucially, when $\wmin \lesssim \inlinemaxnoise$ in the sub-Gaussian noise
setting the assumption $\delta \lesssim 1/(\kappa^{3} \sqrt{k} \log^{2} d)$
implies that for sample size $n$, the RIP parameter $\delta = O(n^{-3/2})$,
which is in general impossible to satisfy, e.g. when the entries of $\X$ are
i.i.d. Gaussian.  Hence moving the dependence on $\kappa$ into a logarithmic
factor as done in our analysis is key for handling the general noisy setting.
For this reason, our proof techniques are necessarily quite different and may
be of independent interest.

%% file: appendix/comparing_with_the_hadamard_product_paper.tex
\section{Comparing Our Results to
\texorpdfstring{\cite{zhao2019implicit}}{Zhao et al. [2019]}}
\label{appendix:comparing-with-hadamard-product-paper}

Instead of using parametrization $w = u \odot u - v \odot v$, the authors of
\cite{zhao2019implicit} consider a closely related Hadamard product
reparametrization $w = u \odot v$ and perform gradient descent updates
on $u$ and $v$ for the least squares objective function with no explicit
regularization.
This work is related to ours in that the ideas of implicit regularization
and sparsity are combined to yield a statistically optimal estimator
for sparse recovery under the RIP assumption.
In this section, we compare this work to ours, pointing out the key
similarities and differences.

To simplify the notation, in all points below we assume that
$\wmin \gtrsim \inlinemaxnoise$ so that the variable $m$ used in
\cite{zhao2019implicit} coincides with $\wmin$ used in this paper.

\paragraph{(Difference) Properly handling noisy setting:}
Let $\kappa \coloneqq \wmax/\wmin$.
The assumption (B) in \cite{zhao2019implicit} requires $\X/\sqrt{n}$
to satisfy $(k + 1, \delta)$-RIP with $\delta \lesssim \frac{1}{\kappa
\sqrt{k}\log (d/\alpha)}$.
On the other hand, for our results to hold it is enough to have
$\delta \lesssim \frac{1}{\sqrt{k} \log \kappa}$. Moving $\kappa$ into a
logarithmic factor is the key difference, which requires a different proof
technique
and also allows to handle the noise properly.
To see why the latter point is true, consider $\wmin \asymp \sigma \sqrt{\log
  d}/\sqrt{n}$.
The assumption (B) in \cite{zhao2019implicit} then requires
$\delta = O(1/(\sqrt{k} \sqrt{n}))$, which is in general impossible to satisfy
with random design matrices, e.g., when entries of $\X$ are i.i.d. Gaussian.
Hence, in contrast to our results,
the results of \cite{zhao2019implicit} cannot recover the smallest possible
signals (i.e., $\wstar$ coordinates of order $\sigma \sqrt{\log d}/{\sqrt{n}}$).

\paragraph{(Difference) Computational optimality:}
In this paper we consider an increasing step size scheme which yields
up to poly-logarithmic factors a computationally optimal algorithm for
sparse recovery under the RIP. On the other hand, only constant step sizes were
considered in \cite{zhao2019implicit}, which does not result in a
computationally optimal algorithm.

Moreover, due to different constraints on step sizes,
the two papers yield different iteration complexities for early stopping times
even in the setting of running gradient descent with constant step sizes.
In \cite[Theorem 3.2]{zhao2019implicit} the required number of iterations is
 $\Omega(\frac{\log (d/\alpha)}{\eta \wmin})
= \Omega(\frac{\kappa}{\wmin}\log^{2} (d/\alpha))$.
If $\wmin \asymp \sigma \sqrt{\log d}/\sqrt{n}$
the required number of iterations is then $\Omega(\frac{n \wmax}{\sigma^{2}}
\log (d/\alpha))$.
On the other hand, in our paper Theorem~\ref{thm:constant-step-sizes}
together with step size tuned by using
Theorem~\ref{thm:selecting-the-step-size},
requires $O(\kappa \log{\alpha^{-1}}) = O(\frac{\sqrt{n}\wmax}{\sigma} \log
\alpha^{-1})$ iterations, yielding an algorithm faster by a factor of
$\sqrt{n}$.

\paragraph{(Difference) Conditions on step size:}
We require $\eta \lesssim 1/\wmax$ while \cite{zhao2019implicit} requires
(Assumption (C)) that $\eta \lesssim
\frac{\wmin}{\wmax}(\log\frac{d}{\alpha})^{-1}$.
The crucial difference is that this step size can be much smaller than
$1/\wmax$ required in our theorems and impacts computational efficiency
as discussed in the computational optimality paragraph above.

Furthermore, a crucial result in our paper is
Theorem~\ref{thm:selecting-the-step-size} which allows us to optimally tune the
step size with an estimate of $\wmax$ that can be computed from the data.
On the other hand, in \cite{zhao2019implicit}
$\eta$ also depends on $\wmin$. It is not clear how to choose such an $\eta$ in
practice and hence it
becomes an additional hyperparameter which needs to be
tuned.

\paragraph{(Difference) Dependence on $\wmax$:}
Our results establish explicit dependence on $\wmax$, while
assumption (A) in \cite{zhao2019implicit} requires $\wmax \lesssim 1$.

\paragraph{(Similarity) Recovering only coordinates above the noise level:}
In both papers, the early stopping procedure stops while
for all $i \in S$ such that $\inlineabs{w^{\star}_{i}} \lesssim
\inlinemaxnoise$ we have $w_{t, i} \approx 0$.
Essentially, such coordinates are treated as if they did not belong to the
true support, since they cannot be recovered as certified by minimax-optimality
bounds.

\paragraph{(Similarity) Statistical optimality:}
Both papers achieve minimax-optimal rates with early stopping and also
prove dimension-independent rates when $\wmin \gtrsim \inlinemaxnoise$.
Our dimension-independent rate (Corollary~\ref{corollary:adaptivity})
has an extra $\log k$ not present in results of \cite{zhao2019implicit}.
We attribute this difference to stronger assumptions imposed on RIP parameter
$\delta$ in \cite{zhao2019implicit}. Indeed, the $\log k$ factor comes from
the $\delta\sqrt{k}\norm{\Xt \vec{\xi}/n \odot \id{S}}_{\infty}$
term in Theorems~\ref{thm:constant-step-sizes} and
\ref{thm:main-theorem-noisy-minimax-rates-exponential-convergence}, which
gets smaller with decreasing $\delta$.

%% file: appendix/table_of_notation.tex
\clearpage
\section{Table of Notation}
\label{appendix:table-of-notation}

We denote vectors with boldface letters and real numbers with normal font.
Hence $\vec{w}$ denotes a vector, while for example,
$w_{i}$ denotes the $i\th$ coordinate of $\vec{w}$.
We let $\X$ be a $n \times d$ design matrix, where $n$ is the number of
observations and $d$ is the number of features.
The true parameter is a $k$-sparse vector denoted by $\wstar$ whose
unknown support is denoted by $S \subseteq \{1, \dots, d\}$.
We let $\wmax = \max_{i \in S} \abs{w^{\star}_{i}}$
and $\wmin = \min_{i \in S} \abs{w^{\star}_{i}}$.
We let $\id{}$ be a vector of ones, and for any index set $A$
we let $\id{A}$ denote a vector equal to $1$ for all coordinates $i \in A$
and equal to $0$ everywhere else.
We denote coordinate-wise product
of vectors by $\odot$ and coordinate-wise inequalities by
$\preccurlyeq$. With a slight abuse of notation we write
$\vec{w}^{2}$ to mean coordinate-wise square of each element for
a vector $\vec{w}$.
Finally, we denote inequalities up to multiplicative absolute constants,
meaning that they do not depend on any parameters of the problem, by
$\lesssim$.

\begin{table}[h]
  \caption{Table of notation}
  \label{table:notation}
  \centering
  \begin{tabular}{ll}
    \toprule
    Symbol            & Description \\
    \midrule
    $n$               & Number of data points  \\
    $d$               & Number of features \\
    $k$               & Sparsity of the true solution \\
    $\wstar$           & Ground truth parameter \\
    \wmax             & $\max_{i \in \{1, \dots, k\} } \abs{w^{\star}_{i}}$ \\
    \wmin             & $\min_{i \in \{1, \dots, k\} } \abs{w^{\star}_{i}}$ \\
    $\kappa$            & $\wmax/\wmin$ \\
    $\kappa^{\text{eff}}$   & $\wmax/(\wmin \vee \varepsilon \vee
                          (\inlinemaxnoise))$ \\
     $\odot$					& Coordinatewise multiplication operator for vectors \\
     $\preccurlyeq$   & A coordinatewise inequality symbol for vectors \\
     $\lesssim$       & An inequality up to some multiplicative absolute
                 constant \\
		$\vec{w}_{t}$           & Gradient descent iterate at time $t$ equal to
												$u_{t} \odot u_{t} + v_{t} \odot v_{t}$ \\
    $\vec{u}_{t}$           & Parametrization of the positive part of $w_{t}$ \\
    $\vec{v}_{t}$           & Parametrization of the negative part of $w_{t}$ \\
    $\alpha$          & Initialization of $u_{0}$ and $v_{0}$ \\
    $\eta$            & The step size for gradient descent updates \\
	  $\vec{w}_{t}^{+}$       & $u_{t} \odot u_{t}$ \\
		$\vec{w}_{t}^{-}$       & $v_{t} \odot v_{t}$ \\
		$S$               & Support of the true parameter $w^{*}$ \\
		$S^{+}$           & Support of positive elements of the true parameter
 												$w^{*}$ \\
		$S^{-}$           & Support of negative elements of the true parameter
 												$w^{*}$ \\
		$\id{A}$          & A vector with coordinates set to $1$ on some index set
                        $A$ and $0$ everywhere else \\
    $\id{i}$          & A short-hand notation for $\id{\{i\}}$ \\
	  $\vec{s}_{t}$           & The signal sequence equal to
												$\id{S^{+}} \odot w_{t}^{+}
                        +\id{S^{-}} \odot w_{t}^{-}$ \\
		$\vec{e}_{t}$           & The error sequence equal to
                        $\id{S^{c}} \odot w_{t}
                        +\id{S^{-}} \odot w_{t}^{+}
                        +\id{S^{+}} \odot w_{t}^{-}$ \\
		$\vec{b}_{t}$           & Represents sequences of bounded errors \\
    $\vec{p}_{t}$           & Represents sequences with errors proportional to
                              the \\
                            & convergence distance
                              $\norm{\vec{s}_{t} - \wstar}_{\infty}$ \\
    \bottomrule
  \end{tabular}
\end{table}

%% file: main.bbl
\begin{thebibliography}{56}
\providecommand{\natexlab}[1]{#1}
\providecommand{\url}[1]{\texttt{#1}}
\expandafter\ifx\csname urlstyle\endcsname\relax
  \providecommand{\doi}[1]{doi: #1}\else
  \providecommand{\doi}{doi: \begingroup \urlstyle{rm}\Url}\fi

\bibitem[Adamczak et~al.(2011)Adamczak, Litvak, Pajor, and
  Tomczak-Jaegermann]{adamczak2011restricted}
Radoslaw Adamczak, Alexander~E Litvak, Alain Pajor, and Nicole
  Tomczak-Jaegermann.
\newblock Restricted isometry property of matrices with independent columns and
  neighborly polytopes by random sampling.
\newblock \emph{Constructive Approximation}, 34\penalty0 (1):\penalty0 61--88,
  2011.

\bibitem[Agarwal et~al.(2010)Agarwal, Negahban, and
  Wainwright]{agarwal2010fast}
Alekh Agarwal, Sahand Negahban, and Martin~J Wainwright.
\newblock Fast global convergence rates of gradient methods for
  high-dimensional statistical recovery.
\newblock In \emph{Advances in Neural Information Processing Systems}, pages
  37--45, 2010.

\bibitem[Ali et~al.(2018)Ali, Kolter, and Tibshirani]{ali2018continuous}
Alnur Ali, J~Zico Kolter, and Ryan~J Tibshirani.
\newblock A continuous-time view of early stopping for least squares
  regression.
\newblock \emph{arXiv preprint arXiv:1810.10082}, 2018.

\bibitem[Bach et~al.(2012)Bach, Jenatton, Mairal, Obozinski,
  et~al.]{bach2012optimization}
Francis Bach, Rodolphe Jenatton, Julien Mairal, Guillaume Obozinski, et~al.
\newblock Optimization with sparsity-inducing penalties.
\newblock \emph{Foundations and Trends{\textregistered} in Machine Learning},
  4\penalty0 (1):\penalty0 1--106, 2012.

\bibitem[Bandeira et~al.(2013)Bandeira, Dobriban, Mixon, and
  Sawin]{bandeira2013certifying}
Afonso~S Bandeira, Edgar Dobriban, Dustin~G Mixon, and William~F Sawin.
\newblock Certifying the restricted isometry property is hard.
\newblock \emph{IEEE transactions on information theory}, 59\penalty0
  (6):\penalty0 3448--3450, 2013.

\bibitem[Baraniuk et~al.(2008)Baraniuk, Davenport, DeVore, and
  Wakin]{baraniuk2008simple}
Richard Baraniuk, Mark Davenport, Ronald DeVore, and Michael Wakin.
\newblock A simple proof of the restricted isometry property for random
  matrices.
\newblock \emph{Constructive Approximation}, 28\penalty0 (3):\penalty0
  253--263, 2008.

\bibitem[Bauer et~al.(2007)Bauer, Pereverzev, and
  Rosasco]{bauer2007regularization}
Frank Bauer, Sergei Pereverzev, and Lorenzo Rosasco.
\newblock On regularization algorithms in learning theory.
\newblock \emph{Journal of complexity}, 23\penalty0 (1):\penalty0 52--72, 2007.

\bibitem[Bertsimas et~al.(2016)Bertsimas, King, and
  Mazumder]{bertsimas2016best}
Dimitris Bertsimas, Angela King, and Rahul Mazumder.
\newblock Best subset selection via a modern optimization lens.
\newblock \emph{The Annals of Statistics}, 44\penalty0 (2):\penalty0 813--852,
  2016.

\bibitem[Bickel et~al.(2009)Bickel, Ritov, Tsybakov,
  et~al.]{bickel2009simultaneous}
Peter~J Bickel, Ya'acov Ritov, Alexandre~B Tsybakov, et~al.
\newblock Simultaneous analysis of {L}asso and {D}antzig selector.
\newblock \emph{The Annals of Statistics}, 37\penalty0 (4):\penalty0
  1705--1732, 2009.

\bibitem[Bredies and Lorenz(2008)]{Bredies2008}
Kristian Bredies and Dirk~A. Lorenz.
\newblock Linear convergence of iterative soft-thresholding.
\newblock \emph{Journal of Fourier Analysis and Applications}, 14\penalty0
  (5):\penalty0 813--837, Dec 2008.
\newblock ISSN 1531-5851.
\newblock \doi{10.1007/s00041-008-9041-1}.
\newblock URL \url{https://doi.org/10.1007/s00041-008-9041-1}.

\bibitem[B{\"u}hlmann and Van De~Geer(2011)]{buhlmann2011statistics}
Peter B{\"u}hlmann and Sara Van De~Geer.
\newblock \emph{Statistics for high-dimensional data: methods, theory and
  applications}.
\newblock Springer Science \& Business Media, 2011.

\bibitem[B{\"u}hlmann and Yu(2003)]{buhlmann2003boosting}
Peter B{\"u}hlmann and Bin Yu.
\newblock Boosting with the $\ell_2$ loss: Regression and classification.
\newblock \emph{Journal of the American Statistical Association}, 98\penalty0
  (462):\penalty0 324--339, 2003.

\bibitem[Candes and Tao(2007)]{candes2007dantzig}
Emmanuel Candes and Terence Tao.
\newblock The {D}antzig selector: Statistical estimation when p is much larger
  than n.
\newblock \emph{The Annals of Statistics}, 35\penalty0 (6):\penalty0
  2313--2351, 2007.

\bibitem[Candes and Tao(2005)]{candes2005decoding}
Emmanuel~J Candes and Terence Tao.
\newblock Decoding by linear programming.
\newblock \emph{IEEE Transactions on Information Theory}, 51\penalty0
  (12):\penalty0 4203--4215, 2005.

\bibitem[Chen et~al.(1998)Chen, Donoho, and Saunders]{chen1998atomic}
Scott~Shaobing Chen, David~L Donoho, and Michael~A Saunders.
\newblock Atomic decomposition by basis pursuit.
\newblock \emph{SIAM Journal on Scientific Computing}, 20\penalty0
  (1):\penalty0 33, 1998.

\bibitem[Cohen et~al.(2009)Cohen, Dahmen, and DeVore]{cohen2009compressed}
Albert Cohen, Wolfgang Dahmen, and Ronald DeVore.
\newblock Compressed sensing and best k-term approximation.
\newblock \emph{Journal of the American mathematical society}, 22\penalty0
  (1):\penalty0 211--231, 2009.

\bibitem[Donoho and Huo(2001)]{donoho2001uncertainty}
David~L Donoho and Xiaoming Huo.
\newblock Uncertainty principles and ideal atomic decomposition.
\newblock \emph{IEEE transactions on information theory}, 47\penalty0
  (7):\penalty0 2845--2862, 2001.

\bibitem[Efron et~al.(2004)Efron, Hastie, Johnstone, and
  Tibshirani]{efron2004least}
Bradley Efron, Trevor Hastie, Iain Johnstone, and Robert Tibshirani.
\newblock Least angle regression.
\newblock \emph{The Annals of Statistics}, 32\penalty0 (2):\penalty0 407--499,
  2004.

\bibitem[Feuer and Nemirovski(2003)]{feuer2003sparse}
Arie Feuer and Arkadi Nemirovski.
\newblock On sparse representation in pairs of bases.
\newblock \emph{IEEE Transactions on Information Theory}, 49\penalty0
  (6):\penalty0 1579--1581, 2003.

\bibitem[Friedman and Popescu(2004)]{friedman2004gradient}
Jerome Friedman and Bogdan~E Popescu.
\newblock Gradient directed regularization.
\newblock \emph{Technical report}, 2004.

\bibitem[Friedman et~al.(2001)Friedman, Hastie, and
  Tibshirani]{friedman2001elements}
Jerome Friedman, Trevor Hastie, and Robert Tibshirani.
\newblock \emph{The elements of statistical learning}, volume~1.
\newblock Springer series in statistics New York, 2001.

\bibitem[Friedman et~al.(2010)Friedman, Hastie, and
  Tibshirani]{friedman2010regularization}
Jerome Friedman, Trevor Hastie, and Rob Tibshirani.
\newblock Regularization paths for generalized linear models via coordinate
  descent.
\newblock \emph{Journal of Statistical Software}, 33\penalty0 (1):\penalty0 1,
  2010.

\bibitem[Gu{\'e}don et~al.(2007)Gu{\'e}don, Mendelson, Pajor, and
  Tomczak-Jaegermann]{guedon2007subspaces}
Olivier Gu{\'e}don, Shahar Mendelson, Alain Pajor, and Nicole
  Tomczak-Jaegermann.
\newblock Subspaces and orthogonal decompositions generated by bounded
  orthogonal systems.
\newblock \emph{Positivity}, 11\penalty0 (2):\penalty0 269--283, 2007.

\bibitem[Gu{\'e}don et~al.(2008)Gu{\'e}don, Mendelson, Pajor,
  Tomczak-Jaegermann, et~al.]{guedon2008majorizing}
Olivier Gu{\'e}don, Shahar Mendelson, Alain Pajor, Nicole Tomczak-Jaegermann,
  et~al.
\newblock Majorizing measures and proportional subsets of bounded orthonormal
  systems.
\newblock \emph{Revista matem{\'a}tica iberoamericana}, 24\penalty0
  (3):\penalty0 1075--1095, 2008.

\bibitem[Gunasekar et~al.(2017)Gunasekar, Woodworth, Bhojanapalli, Neyshabur,
  and Srebro]{gunasekar2017implicit}
Suriya Gunasekar, Blake~E Woodworth, Srinadh Bhojanapalli, Behnam Neyshabur,
  and Nati Srebro.
\newblock Implicit regularization in matrix factorization.
\newblock In \emph{Advances in Neural Information Processing Systems}, pages
  6151--6159, 2017.

\bibitem[Gunasekar et~al.(2018{\natexlab{a}})Gunasekar, Lee, Soudry, and
  Srebro]{gunasekar2018characterizing}
Suriya Gunasekar, Jason Lee, Daniel Soudry, and Nathan Srebro.
\newblock Characterizing implicit bias in terms of optimization geometry.
\newblock In \emph{International Conference on Machine Learning}, pages
  1827--1836, 2018{\natexlab{a}}.

\bibitem[Gunasekar et~al.(2018{\natexlab{b}})Gunasekar, Lee, Soudry, and
  Srebro]{gunasekar2018implicit}
Suriya Gunasekar, Jason~D Lee, Daniel Soudry, and Nati Srebro.
\newblock Implicit bias of gradient descent on linear convolutional networks.
\newblock In \emph{Advances in Neural Information Processing Systems}, pages
  9461--9471, 2018{\natexlab{b}}.

\bibitem[Hale et~al.(2008)Hale, Yin, and Zhang]{hale2008fixed}
Elaine~T Hale, Wotao Yin, and Yin Zhang.
\newblock Fixed-point continuation for $\ell\_1$-minimization: Methodology and
  convergence.
\newblock \emph{SIAM Journal on Optimization}, 19\penalty0 (3):\penalty0
  1107--1130, 2008.

\bibitem[Hastie et~al.(2017)Hastie, Tibshirani, and
  Tibshirani]{hastie2017extended}
Trevor Hastie, Robert Tibshirani, and Ryan~J Tibshirani.
\newblock Extended comparisons of best subset selection, forward stepwise
  selection, and the lasso.
\newblock \emph{arXiv preprint arXiv:1707.08692}, 2017.

\bibitem[Li et~al.(2018)Li, Ma, and Zhang]{li2018algorithmic}
Yuanzhi Li, Tengyu Ma, and Hongyang Zhang.
\newblock Algorithmic regularization in over-parameterized matrix sensing and
  neural networks with quadratic activations.
\newblock In \emph{Conference On Learning Theory}, pages 2--47, 2018.

\bibitem[Meinshausen and Yu(2009)]{meinshausen2009lasso}
Nicolai Meinshausen and Bin Yu.
\newblock Lasso-type recovery of sparse representations for high-dimensional
  data.
\newblock \emph{The Annals of Statistics}, 37\penalty0 (1):\penalty0 246--270,
  2009.

\bibitem[Mendelson et~al.(2008)Mendelson, Pajor, and
  Tomczak-Jaegermann]{mendelson2008uniform}
Shahar Mendelson, Alain Pajor, and Nicole Tomczak-Jaegermann.
\newblock Uniform uncertainty principle for bernoulli and subgaussian
  ensembles.
\newblock \emph{Constructive Approximation}, 28\penalty0 (3):\penalty0
  277--289, 2008.

\bibitem[Negahban et~al.(2012)Negahban, Ravikumar, Wainwright, Yu,
  et~al.]{negahban2012unified}
Sahand~N Negahban, Pradeep Ravikumar, Martin~J Wainwright, Bin Yu, et~al.
\newblock A unified framework for high-dimensional analysis of ${M}$-estimators
  with decomposable regularizers.
\newblock \emph{Statistical Science}, 27\penalty0 (4):\penalty0 538--557, 2012.

\bibitem[Neu and Rosasco(2018)]{neu2018iterate}
Gergely Neu and Lorenzo Rosasco.
\newblock Iterate averaging as regularization for stochastic gradient descent.
\newblock \emph{Proceedings of Machine Learning Research vol}, 75:\penalty0
  1--21, 2018.

\bibitem[Parikh et~al.(2014)Parikh, Boyd, et~al.]{parikh2014proximal}
Neal Parikh, Stephen Boyd, et~al.
\newblock Proximal algorithms.
\newblock \emph{Foundations and Trends{\textregistered} in Optimization},
  1\penalty0 (3):\penalty0 127--239, 2014.

\bibitem[Raskutti et~al.(2010)Raskutti, Wainwright, and
  Yu]{raskutti2010restricted}
Garvesh Raskutti, Martin~J Wainwright, and Bin Yu.
\newblock Restricted eigenvalue properties for correlated gaussian designs.
\newblock \emph{Journal of Machine Learning Research}, 11\penalty0
  (Aug):\penalty0 2241--2259, 2010.

\bibitem[Raskutti et~al.(2011)Raskutti, Wainwright, and
  Yu]{raskutti2011minimax}
Garvesh Raskutti, Martin~J Wainwright, and Bin Yu.
\newblock Minimax rates of estimation for high-dimensional linear regression
  over $\ell_q$-balls.
\newblock \emph{IEEE transactions on information theory}, 57\penalty0
  (10):\penalty0 6976--6994, 2011.

\bibitem[Raskutti et~al.(2014)Raskutti, Wainwright, and Yu]{raskutti2014early}
Garvesh Raskutti, Martin~J Wainwright, and Bin Yu.
\newblock Early stopping and non-parametric regression: an optimal
  data-dependent stopping rule.
\newblock \emph{The Journal of Machine Learning Research}, 15\penalty0
  (1):\penalty0 335--366, 2014.

\bibitem[Romberg(2009)]{romberg2009compressive}
Justin Romberg.
\newblock Compressive sensing by random convolution.
\newblock \emph{SIAM Journal on Imaging Sciences}, 2\penalty0 (4):\penalty0
  1098--1128, 2009.

\bibitem[Rosset et~al.(2004)Rosset, Zhu, and Hastie]{rosset2004boosting}
Saharon Rosset, Ji~Zhu, and Trevor Hastie.
\newblock Boosting as a regularized path to a maximum margin classifier.
\newblock \emph{Journal of Machine Learning Research}, 5\penalty0
  (Aug):\penalty0 941--973, 2004.

\bibitem[Rudelson and Vershynin(2008)]{rudelson2008sparse}
Mark Rudelson and Roman Vershynin.
\newblock On sparse reconstruction from fourier and gaussian measurements.
\newblock \emph{Communications on Pure and Applied Mathematics: A Journal
  Issued by the Courant Institute of Mathematical Sciences}, 61\penalty0
  (8):\penalty0 1025--1045, 2008.

\bibitem[Rudelson and Zhou(2012)]{rudelson2012reconstruction}
Mark Rudelson and Shuheng Zhou.
\newblock Reconstruction from anisotropic random measurements.
\newblock In \emph{Conference on Learning Theory}, pages 10--1, 2012.

\bibitem[Soudry et~al.(2018)Soudry, Hoffer, Nacson, Gunasekar, and
  Srebro]{soudry2018implicit}
Daniel Soudry, Elad Hoffer, Mor~Shpigel Nacson, Suriya Gunasekar, and Nathan
  Srebro.
\newblock The implicit bias of gradient descent on separable data.
\newblock \emph{The Journal of Machine Learning Research}, 19\penalty0
  (1):\penalty0 2822--2878, 2018.

\bibitem[Suggala et~al.(2018)Suggala, Prasad, and
  Ravikumar]{suggala2018connecting}
Arun Suggala, Adarsh Prasad, and Pradeep~K Ravikumar.
\newblock Connecting optimization and regularization paths.
\newblock In \emph{Advances in Neural Information Processing Systems}, pages
  10608--10619, 2018.

\bibitem[Tao et~al.(2016)Tao, Boley, and Zhang]{tao2016local}
Shaozhe Tao, Daniel Boley, and Shuzhong Zhang.
\newblock Local linear convergence of {ISTA} and {FISTA} on the {LASSO}
  problem.
\newblock \emph{SIAM Journal on Optimization}, 26\penalty0 (1):\penalty0
  313--336, 2016.

\bibitem[Tibshirani(1996)]{tibshirani1996regression}
Robert Tibshirani.
\newblock Regression shrinkage and selection via the {L}asso.
\newblock \emph{Journal of the Royal Statistical Society: Series B
  (Methodological)}, 58\penalty0 (1):\penalty0 267--288, 1996.

\bibitem[Tibshirani et~al.(2015)Tibshirani, Wainwright, and
  Hastie]{tibshirani2015statistical}
Robert Tibshirani, Martin Wainwright, and Trevor Hastie.
\newblock \emph{Statistical learning with sparsity: the lasso and
  generalizations}.
\newblock Chapman and Hall/CRC, 2015.

\bibitem[van~de Geer(2007)]{van2007deterministic}
Sara van~de Geer.
\newblock The deterministic {L}asso.
\newblock \emph{Research Report}, 140, 2007.

\bibitem[Van De~Geer et~al.(2009)Van De~Geer, B{\"u}hlmann,
  et~al.]{van2009conditions}
Sara~A Van De~Geer, Peter B{\"u}hlmann, et~al.
\newblock On the conditions used to prove oracle results for the lasso.
\newblock \emph{Electronic Journal of Statistics}, 3:\penalty0 1360--1392,
  2009.

\bibitem[Wainwright(2019)]{wainwright2019high}
Martin~J Wainwright.
\newblock \emph{High-dimensional statistics: A non-asymptotic viewpoint},
  volume~48.
\newblock Cambridge University Press, 2019.

\bibitem[Wei et~al.(2017)Wei, Yang, and Wainwright]{wei2017early}
Yuting Wei, Fanny Yang, and Martin~J Wainwright.
\newblock Early stopping for kernel boosting algorithms: A general analysis
  with localized complexities.
\newblock In \emph{Advances in Neural Information Processing Systems}, pages
  6065--6075, 2017.

\bibitem[Yao et~al.(2007)Yao, Rosasco, and Caponnetto]{yao2007early}
Yuan Yao, Lorenzo Rosasco, and Andrea Caponnetto.
\newblock On early stopping in gradient descent learning.
\newblock \emph{Constructive Approximation}, 26\penalty0 (2):\penalty0
  289--315, 2007.

\bibitem[Zhang et~al.(2016)Zhang, Bengio, Hardt, Recht, and
  Vinyals]{zhang2016understanding}
Chiyuan Zhang, Samy Bengio, Moritz Hardt, Benjamin Recht, and Oriol Vinyals.
\newblock Understanding deep learning requires rethinking generalization.
\newblock \emph{arXiv preprint arXiv:1611.03530}, 2016.

\bibitem[Zhang and Yu(2005)]{zhang2005boosting}
Tong Zhang and Bin Yu.
\newblock Boosting with early stopping: Convergence and consistency.
\newblock \emph{The Annals of Statistics}, 33\penalty0 (4):\penalty0
  1538--1579, 2005.

\bibitem[Zhang et~al.(2014)Zhang, Wainwright, and Jordan]{zhang2014lower}
Yuchen Zhang, Martin~J Wainwright, and Michael~I Jordan.
\newblock Lower bounds on the performance of polynomial-time algorithms for
  sparse linear regression.
\newblock In \emph{Conference on Learning Theory}, pages 921--948, 2014.

\bibitem[Zhao et~al.(2019)Zhao, Yang, and He]{zhao2019implicit}
Peng Zhao, Yun Yang, and Qiao-Chu He.
\newblock Implicit regularization via hadamard product over-parametrization in
  high-dimensional linear regression.
\newblock \emph{arXiv preprint arXiv:1903.09367}, 2019.

\end{thebibliography}
